\documentclass[final,3p,times,twocolumn]{elsarticle}

\usepackage[T1]{fontenc}
\usepackage{amssymb}
\usepackage{amsmath}
\usepackage{adjustbox}
\usepackage{graphicx}
\usepackage{array}
\usepackage{makecell}
\usepackage{rotating}
\usepackage{multirow}
\usepackage{float}
\usepackage{soul}
\usepackage{xcolor}
\usepackage{subcaption}
\usepackage[utf8]{inputenc}
\usepackage{xcolor}

\journal{Transportation Research Part A}

\begin{document}

\begin{frontmatter}

\title{Intelligent CCTV for Urban Design: AI-Based Analysis of Soft Infrastructure at Intersections}

\author[label1]{Vinit Katariya\corref{cor1}}
% \ead{vkatariy@uwyo.edu}
\cortext[cor1]{Corresponding author: vkatariy@uwyo.edu}
\author[label1]{Seungjin Kim}
\author[label3]{Curtis Craig}
\author[label3]{Nichole Morris}
\author[label2]{Hamed Tabkhi}

\affiliation[label1]{
  organization={University of Wyoming},
  city={Laramie},
  state={WY},
  country={USA}
}

\affiliation[label3]{
  organization={University of North Carolina at Charlotte},
  city={Charlotte},
  state={NC},
  country={USA}
}

\affiliation[label2]{
  organization={University of Minnesota},
  city={Minneapolis},
  state={MN},
  country={USA}
}

\begin{abstract}
Artificial intelligence (AI) and computer vision are transforming transportation data collection. This study introduces an AI-enabled analytics framework leveraging existing CCTV infrastructure to evaluate the impact of soft interventions, such as temporary pedestrian refuges and curb extensions, on vehicle speed and safety. Using deep learning and perspective-based speed estimation, we evaluated driver behavior before and after interventions, with repeated post-installation monitoring in Week 1 and Week 2, in Minneapolis. Findings reveal that at unsignalized intersections, mean and 85th-percentile speeds fell by up to 18.75\% and 16.56\%, respectively, while pass-through traffic decreased by as much as 12.2\%. Signalized intersections showed comparable reductions except one location, with mean and 85th-percentile speeds dropping by up to 20.0\% and 17.19\%. These results demonstrate the traffic-calming effectiveness of soft infrastructure and underscore the utility of AI-powered methods for rapid, low-cost, and evidence-based transport policy evaluation.
\end{abstract}  


\end{frontmatter}

\section{Introduction}

Urban intersections are becoming increasingly complex hubs of activity as vehicle traffic, pedestrian flows, and micro-mobility users continue to grow in number and diversity. However, traditional traffic management and safety systems at these junctions have struggled to keep pace with this complexity. Most conventional approaches rely on fixing the urban roadways and infrastructure after reports of safety concerns and crashes, leading to changes that are reactive rather than proactive. Moreover, current infrastructure modifications, such as concrete refuges and dividers, are costly and often require specialized agencies, often leading to months-long delays before implementation. Furthermore, the effectiveness of these interventions must be quickly and correctly evaluated, as they do not always guarantee improved safety for road users. Traditional assessment methods, such as manual speed measurements using speed guns or high-cost speed sensors, further extend this timeline, adding additional months to the evaluation process. This lag highlights the critical need for cost-effective techniques that minimize delays, reduce the financial burden on agencies, and streamline both the implementation and assessment process, making these safety improvements more feasible and effective.

\begin{figure*}[t!]
    \centering
    \includegraphics[width=\textwidth]{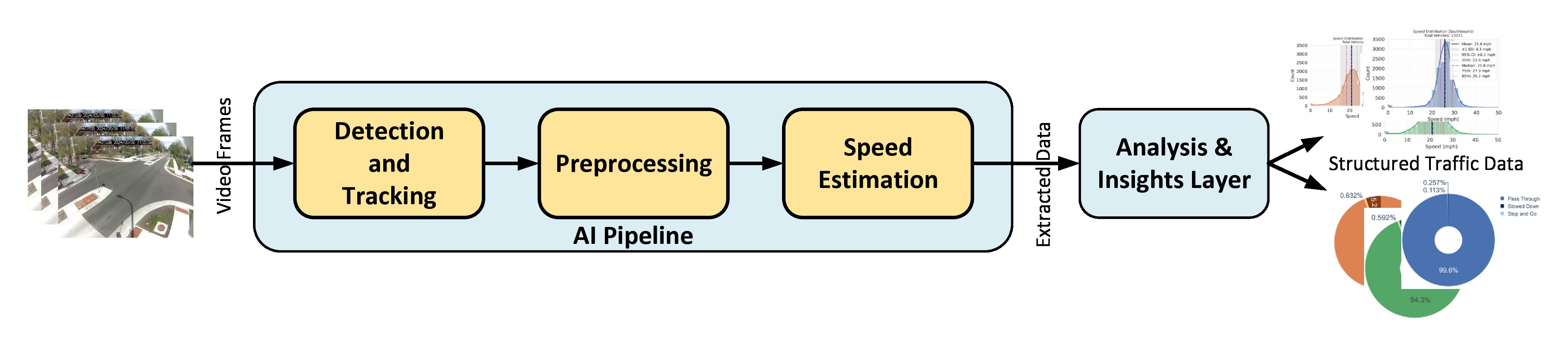}
    \caption{Overview of the proposed AI-driven analytics framework. The system comprises detection \& tracking, preprocessing, speed estimation, and a video analytics module that derives key metrics and insights from the extracted data.}
    \label{fig:framework}
\end{figure*}

Recent advances in artificial intelligence (AI) and computer vision offer a promising avenue to transform passive traffic monitoring into an active, data-driven safety management tool. Many urban areas are already equipped with closed-circuit television (CCTV) cameras and traffic sensors that passively record intersection activity \cite{ali2023bayesian}. 
AI-driven computer vision algorithms can leverage this existing infrastructure to automatically detect and analyze road user behaviors in real time \cite{pourhomayoun2024artificial, katariya2024vegaedge}. By processing live video feeds or recorded footage, these systems can identify hazardous situations—such as vehicles failing to yield to pedestrians or near-miss collisions—that would otherwise go unnoticed in traditional monitoring \cite{zhu2022deep, ali2023bayesian}. 
This capability enables a shift from retrospective analysis to proactive intervention; for instance, if a pattern of high mean speeds (higher than speed limits) is detected repeatedly at certain intersections, city authorities/DoTs can be alerted to implement traffic calming measures or adjust signal timing before an injury occurs. In essence, AI-driven vision systems can convert a city's ubiquitous but underutilized cameras into intelligent agents for pedestrian safety.

One critical use case for AI in traffic safety is evaluating the impact of infrastructure interventions aimed at improving pedestrian safety. Soft (quick-build) interventions, such as bollards, temporary curb extensions, and temporary pedestrian refuge islands, are increasingly deployed to calm traffic and improve crossing safety \cite{gitelman2012characterization, blackburn2018guide}. In this study, we focus on evaluating these soft treatments using CCTV-based AI analytics, while including a small number of permanent/hardened implementations as baseline references for contextual comparison. While these interventions are widely considered effective in reducing vehicle speeds and enhancing driver yielding behavior \cite{zegeer2012pedestrian}, quantifying their impact has remained challenging. Prior studies have relied on crash data collected over long periods or limited observational studies, making it difficult to assess behavioral changes immediately following implementation \cite{ahmad2021effective}.

In this study, we demonstrate the application of AI-based computer vision to bridge this gap by capturing high-resolution, before-and-after behavioral evidence of how soft infrastructure changes influence vehicle movement and pedestrian interactions. As shown in Figure \ref{fig:framework}, by leveraging deep learning combined with transformation techniques for speed estimation, trajectory analysis, and behavioral pattern recognition, we provide a scalable and cost-effective AI-based analytics framework for assessing the effectiveness of roadway interventions on pedestrian safety. Our work not only highlights the potential of AI-driven monitoring in infrastructure evaluation but also underscores its broader role in transforming urban traffic management and proactive safety planning.

We analyze hours of CCTV footage from a busy urban intersection, both prior to and following the implementation of the soft infrastructure interventions. Using object detection and tracking algorithms \cite{Jocher_YOLO_by_Ultralytics_2023,zhang2022bytetrack}, we automatically detect and re-ID the vehicles in the area of interest. This automated analysis yields rich data on key safety metrics of vehicle speeds as they approach and navigate the intersection, and stopping and yielding behavior at crosswalks. 
By comparing these metrics before versus after the installation of bollards, bump-outs, and refuge islands, we can assess changes in driver behavior (e.g., are cars slowing down or stopping more frequently at the crosswalk?) and vehicle speed analysis near the pedestrian crossing. Our AI-driven approach thus provides an objective, quantitative evaluation of infrastructure changes.

In summary, our work brings together advancements in computer vision, data science, urban infrastructure design, and transportation engineering to enhance pedestrian safety at intersections. The key contributions of this paper are as follows:

\begin{itemize}
    \item A novel AI-enabled analytics framework that leverages CCTV infrastructure to gather high-resolution behavioral data, enabling cost-effective and scalable evaluation of soft (quick-build) interventions, with a small set of permanent-treatment baselines for contextual comparison.
    
    \item A before–after field evaluation with repeated post-installation measurements (Week 1 and Week 2) using real-world urban traffic footage, quantifying shifts in key safety metrics such as mean and 85th-percentile speeds, and driver yielding behavior near crosswalks, offering actionable evidence to inform transport planning and policy.
    
    \item A modular AI pipeline for vehicle detection, tracking, and speed estimation using monocular video, designed for rapid deployment by transportation agencies to support data-informed, proactive decision-making in intersection safety and urban design.
\end{itemize}

Overall, this research underscores the relevance of AI-based traffic monitoring solutions in advancing urban safety interventions. By combining technological innovation with thoughtful infrastructure design, the approach presented in this paper aims to contribute to safer, more pedestrian-friendly cities.

\section{Related Works}
Pedestrian safety in urban environments is closely linked to infrastructure design and vehicle speeds, with higher speeds significantly increasing crash severity. Research highlights the need for speed management strategies to mitigate risks. Advancements in AI-powered video analytics now enable real-time assessment of pedestrian-vehicle interactions, leveraging image processing and deep learning. A key challenge in these analytics systems is calculating accurate vehicle speed estimation, where camera-based methods offer a cost-effective alternative to traditional sensor-based approaches. The following sections examine the previous works and the impact of infrastructure and speed, AI-driven video analytics, and vehicle speed estimation techniques.

\subsection{Impact of Infrastructure and Vehicle Speed on Pedestrian Safety}
One critical risk factor highlighted in the traffic safety literature is vehicle speeding. Higher vehicle speeds not only increase the likelihood of collisions but also dramatically worsen outcomes for pedestrians.Vehicle speed directly correlates with injury severity; fatality risks rise from <10\% at 20 mph to >80\% at 40 mph \cite{zegeer2012pedestrian, hauer2009speed,tefft2013impact}. Consequently, speed limit enforcement and automated cameras are critical interventions. Due to such dangers, managing vehicle speeds in urban areas is a top priority for safety. Measures like enforcing lower speed limits and deploying automated speed cameras are widely regarded as effective strategies to protect pedestrians. As Fernández-Llorca et al. \cite{fernandez2021vision} note, the introduction of appropriate speed limits and enforcement via speed cameras has been one of the most impactful interventions for improving road safety. Consequently, there is strong motivation to develop accurate and efficient methods to monitor vehicle speeds in traffic \cite{van1985increasing}, especially using cost-effective technologies that can be deployed at scale in cities.

Research on hard infrastructure interventions such as curb extensions (bumpouts) and pedestrian refuge islands or median refuges suggests their effectiveness in reducing vehicle speeds and improving pedestrian safety. Gitelman et al. \cite{gitelman2012characterization}, in a study conducted in Israel found that curb extensions lead to a reduction in average speeds due to their ability to visually and physically narrow the roadway, prompting driver caution. Similarly, Blackburn et al. \cite{blackburn2018guide} reported that pedestrian refuge islands not only provide safer crossing points but also contribute to speed reductions at uncontrolled (non-signalized) crossings, particularly in urban settings with moderate traffic flow. However, Budzyński et al. \cite{budzynski2021assessment} add depth to this narrative, demonstrating that while infrastructure like refuge islands and roundabouts effectively lowers driver speeds, it does not significantly improve stopping rates at pedestrian crossings. This suggests that speed reduction, though critical, is not the only part of the safety equation, as driver compliance with pedestrian rights-of-way remains a challenge. Their findings highlight the complexity of infrastructure’s effects, indicating that physical changes alone may not fully address all safety concerns. In a simulation study by \cite{bella2015effects}, which examined driver speed and simulated pedestrians, the best performance was observed when zebra crossings were combined with curb extensions supported by \cite{ahmad2021effective}.

However, the effectiveness of soft infrastructure changes such as bollard refuges and curb extensions (bumpouts) has not been studied as extensively. While physical modifications like concrete refuges provide clear visual and structural deterrents, softer interventions rely on perceptual cues to influence driver behavior. Our study builds upon prior research by utilizing AI-driven computer vision techniques to measure vehicle speeds before and after the introduction of soft and hard infrastructure changes. By leveraging existing CCTV camera data, we aim to demonstrate a scalable and cost-effective approach for evaluating roadway safety interventions, extending the applicability of AI in transportation engineering. This approach aims to provide a scalable, cost-effective way to assess the multifaceted impacts of interventions, building on insights from \cite{zegeer2012pedestrian,budzynski2021assessment, bella2015effects} to advance pedestrian safety research.

\subsection{AI-Based Video Analytics for Pedestrian Safety}
In recent years, AI-driven video analytics have been leveraged to enhance pedestrian safety in urban environments. The study by Ali et. al. \cite{ali2023bayesian} leverages Extreme Value Theory (EVT) and Bayesian modeling to estimate real-time pedestrian crash risk at signalized intersections using AI-based analytics. While EVT enables risk estimation from traffic conflicts rather than crash data, it requires large datasets and strong distributional assumptions, highlighting the need for scalable AI-driven approaches in pedestrian safety assessment. Similarly, Hu et al. \cite{xu2019detecting} demonstrated the scalability of detecting crossing events using traffic camera data, which is unexpected for its potential to leverage existing infrastructure without additional costs.

Advanced computer vision systems can automatically detect and track pedestrians and vehicles in traffic camera footage, enabling proactive identification of hazardous situations. Pourhomayoun \cite{pourhomayoun2024artificial} developed an end-to-end system to detect and report near-miss collisions at intersections, using deep learning models to recognize all traffic participants and flag potential pedestrian-vehicle conflicts in real time. Similarly, Zhang et al. \cite{zhu2022deep} applied deep neural networks (e.g., YOLO, Mask R-CNN) to urban intersection videos to analyze dangerous interactions and predict pedestrian conflicts before accidents occur. These works demonstrate that AI-based analysis can serve as an effective tool for surrogate safety assessment—identifying risk events (like close calls and violation of traffic rules) to guide safety improvements without waiting for crashes to happen.

\subsection{Vehicle Speed Estimation Approaches}
Traditional speed detection relies on specialized sensors such as radar and LiDAR \cite{zhang2020vehicle, fernandez2021vision, jeng2013estimating}, which can directly measure vehicle speeds with high accuracy. While effective, these sensor systems are expensive and impractical for blanket coverage of urban road networks. Camera-based speed estimation offers a low-cost alternative, leveraging the ubiquitous presence of traffic cameras. Vision-based methods for speed estimation have been extensively researched over the past decades \cite{elvik1997effects, cypto2022automatic,rasouli2019autonomous}. While early methods relied on feature tracking, perspective transformation (homography) has become the standard for static cameras \cite{luvizon2014vehicle, kocur2020detection}. Kocur and Ftáčnik, for instance, demonstrated that extending 2D bounding boxes of detected vehicles into 3D using a vanishing-point-based calibration allows a single monocular camera to achieve high-precision speed measurements \cite{kocur2020detection}.

In recent years, researchers have also explored deep learning techniques for speed estimation. Instead of relying on manual calibration, these approaches train models to infer speed directly from the video data \cite{perunivcic2023vision}. García-Aguilar et al. \cite{garcia2024real} present a fully learning-based pipeline combining vehicle detection and a CNN model to estimate the speed of surrounding cars in real time for autonomous driving scenarios. While such deep learning approaches are promising, studies indicate that they often struggle to surpass the accuracy of classical geometry-based methods. For example, Rodríguez-Rangel et. al. \cite{rodriguez2022analysis, shaqib2024vehicle} evaluated various algorithms for real-time speed estimation (including a neural network regressor) and found that a simple linear regression model using perspective-corrected distance measurements outperformed more complex machine learning models in accuracy. This outcome highlights the advantage of physics-based calibration, when a camera is properly calibrated, speed calculation becomes a straightforward geometric computation with little ambiguity, whereas purely data-driven models may introduce estimation errors. Thus, in fixed-camera settings, perspective transform techniques remain the gold standard for accurate speed estimation, often yielding better accuracy than end-to-end deep learning methods in practice.

Vision-based methods can face challenges if the camera itself is moving (as in dashcam or drone footage) or if calibration is poor. In those cases, hybrid approaches or additional sensors may be necessary. For instance, on moving platforms where background is not fixed, optical methods become unreliable, and one often resorts to onboard automotive radar or LiDAR to directly measure relative speeds. Researchers have demonstrated that roadside LiDAR units can track vehicles and estimate speeds with very high accuracy, but at a cost too high for ubiquitous deployment. Therefore, for urban deployments, standard video cameras coupled with perspective calibration remain an appealing solution, offering a compelling balance between accuracy and cost-effectiveness.

\section{Methodology}

In this section, we present the methodology employed to evaluate the infrastructure changes for pedestrian safety through AI-driven video analytics. We begin by defining the scope and nature of the changes under investigation, followed by an explanation of our video collection and analytics strategy. Next, we introduce our AI pipeline, outlining the data preprocessing, object detection, and tracking techniques, and speed estimation, concluding with evaluation metrics.

To ensure a robust evaluation of the soft infrastructure changes, video collection was conducted in two distinct phases of before and after the installation of the modifications, in a before–after design with two post-installation observation windows (Week 1 and Week 2). Pre-installation footage was recorded for one week, before the changes, providing a baseline of typical road user behavior (speed distributions and vehicle behavior near pedestrian crossing). Post-installation footage was then gathered continuously over two weeks, with performance evaluated separately for Week 1 and Week 2. This phased approach allows for the observation of both immediate and adjusted responses: \textit{Week 1} captures initial reactions, potentially influenced by the novelty of the changes, while \textit{Week 2} reflects more stable, habituated behavior as road users acclimate to the new infrastructure. By distinguishing these periods, the study mitigates the impact of short-term behavioral anomalies, ensuring that the analysis focuses on the sustained effects of the modifications on pedestrian safety and traffic dynamics.

\subsection{Locations and Infrastructure Changes}

\begin{figure*}[t!]
    \centering
    % Define image width once
    \newcommand{\imgW}{0.18\textwidth} 
    \newcommand{\imgH}{2.5cm} 
    
    \setlength{\tabcolsep}{2pt} 
    \renewcommand{\arraystretch}{0.9}
    
    \begin{tabular}{c >{\centering\arraybackslash}p{\imgW} >{\centering\arraybackslash}p{\imgW} >{\centering\arraybackslash}p{\imgW} >{\centering\arraybackslash}p{\imgW} >{\centering\arraybackslash}p{\imgW}}
        % Block 1: Locations 1-5
        & \textbf{Location 1} & \textbf{Location 2} & \textbf{Location 3} & \textbf{Location 4} & \textbf{Location 5} \\
        \rotatebox{90}{\hspace{7mm}\textbf{Before}} &
        \includegraphics[width=\imgW, height=\imgH, trim=50 50 50 50, clip]{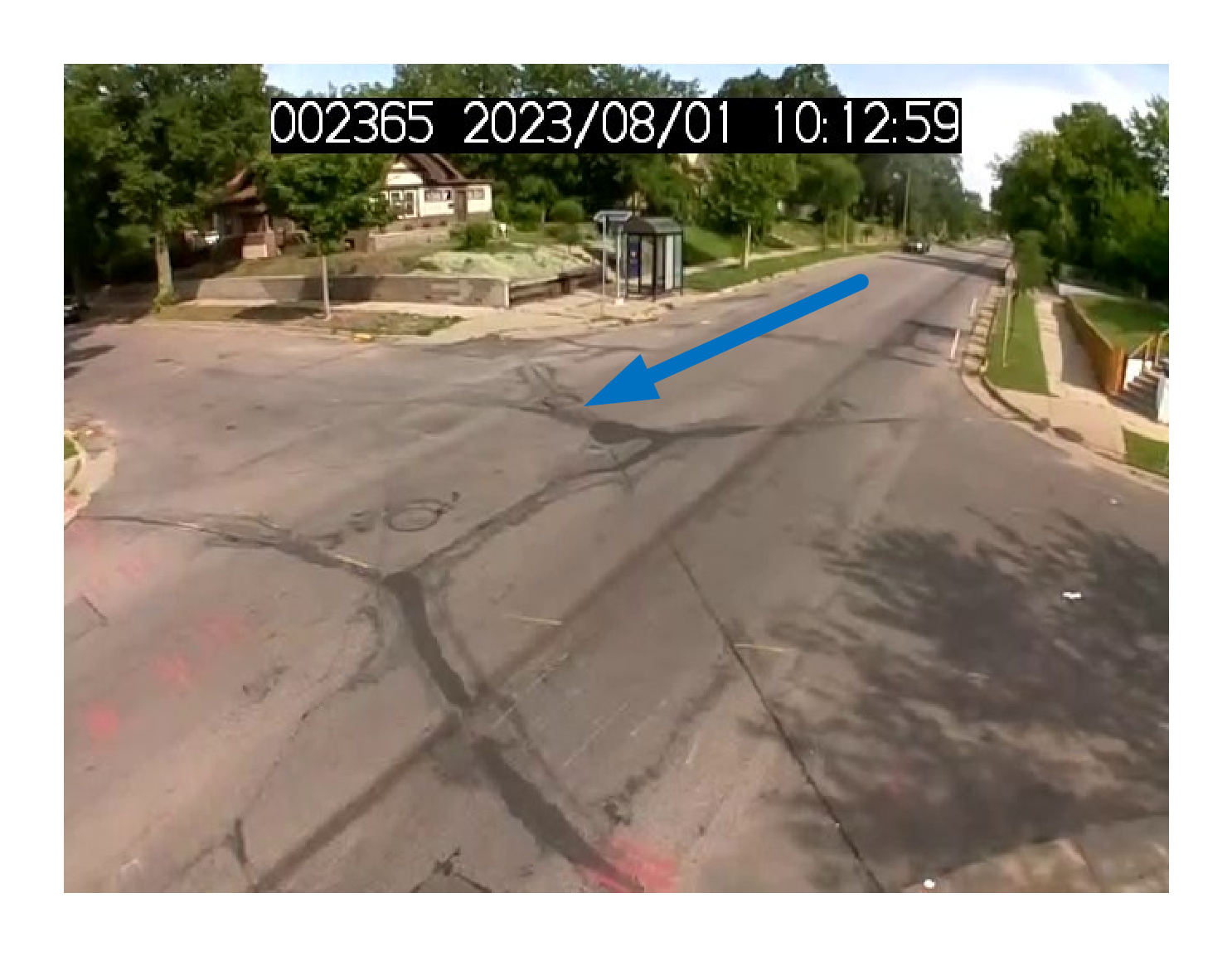} &
        \includegraphics[width=\imgW, height=\imgH, trim=50 50 50 50, clip]{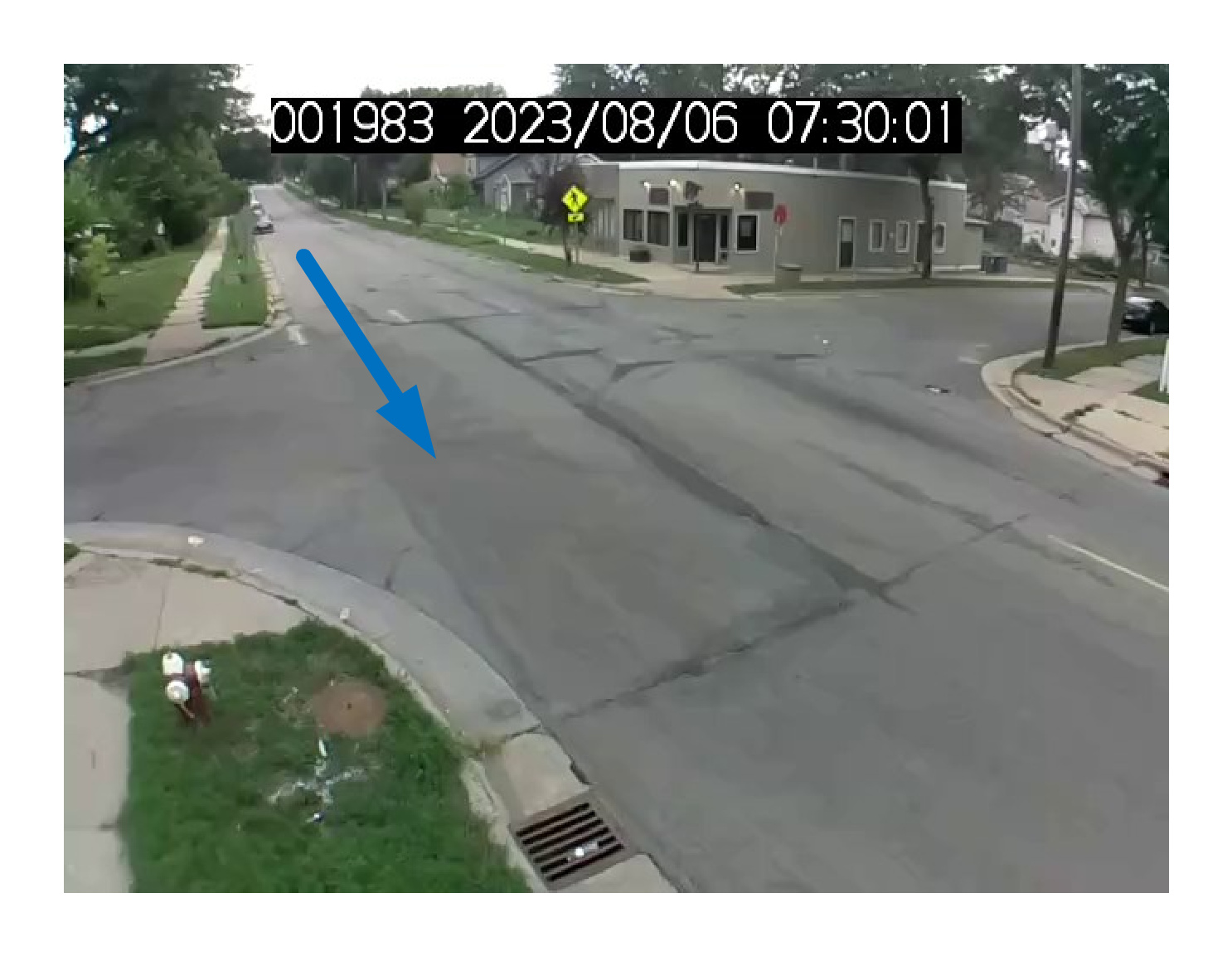} &
        \includegraphics[width=\imgW, height=\imgH, trim=50 50 50 50, clip]{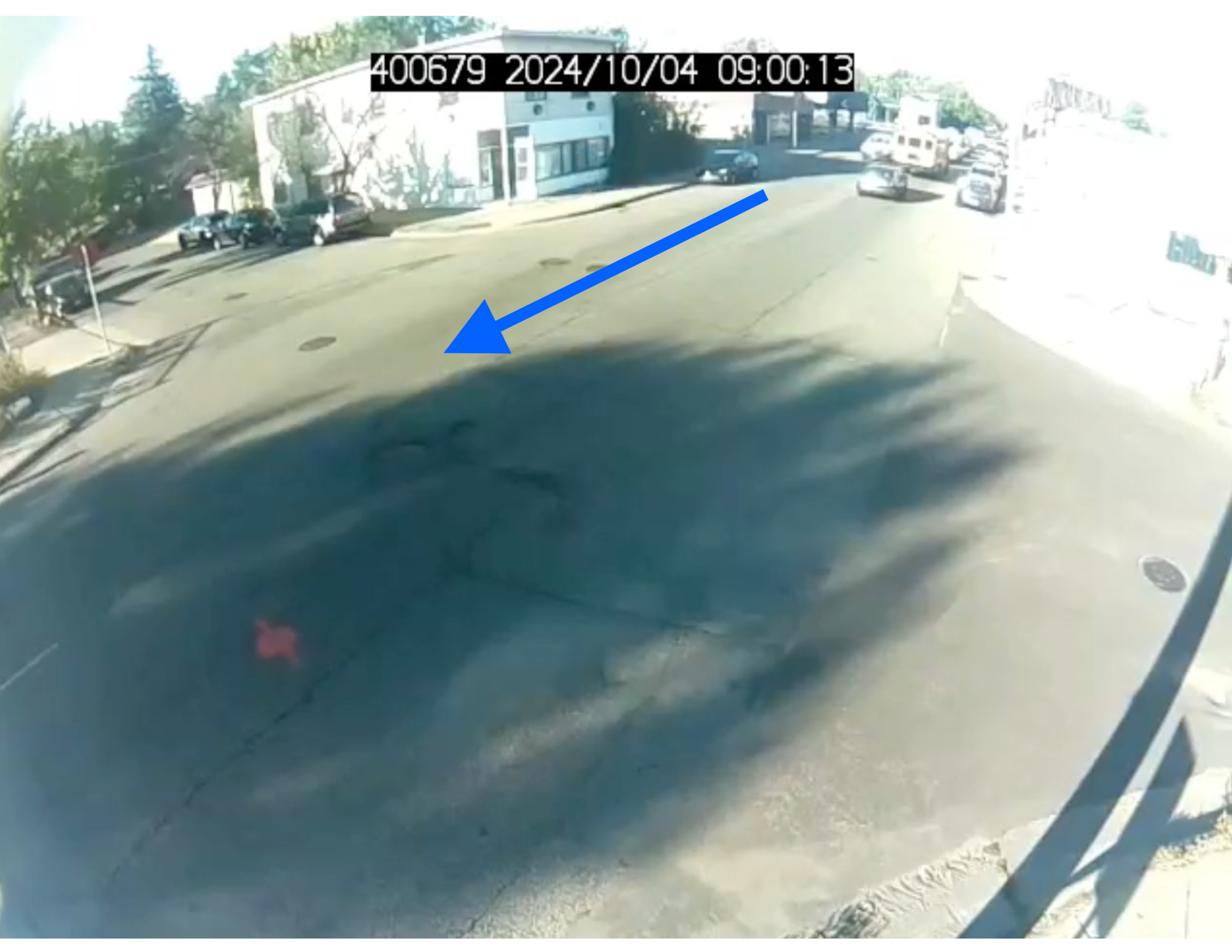} &
        \includegraphics[width=\imgW, height=\imgH, trim=50 50 50 50, clip]{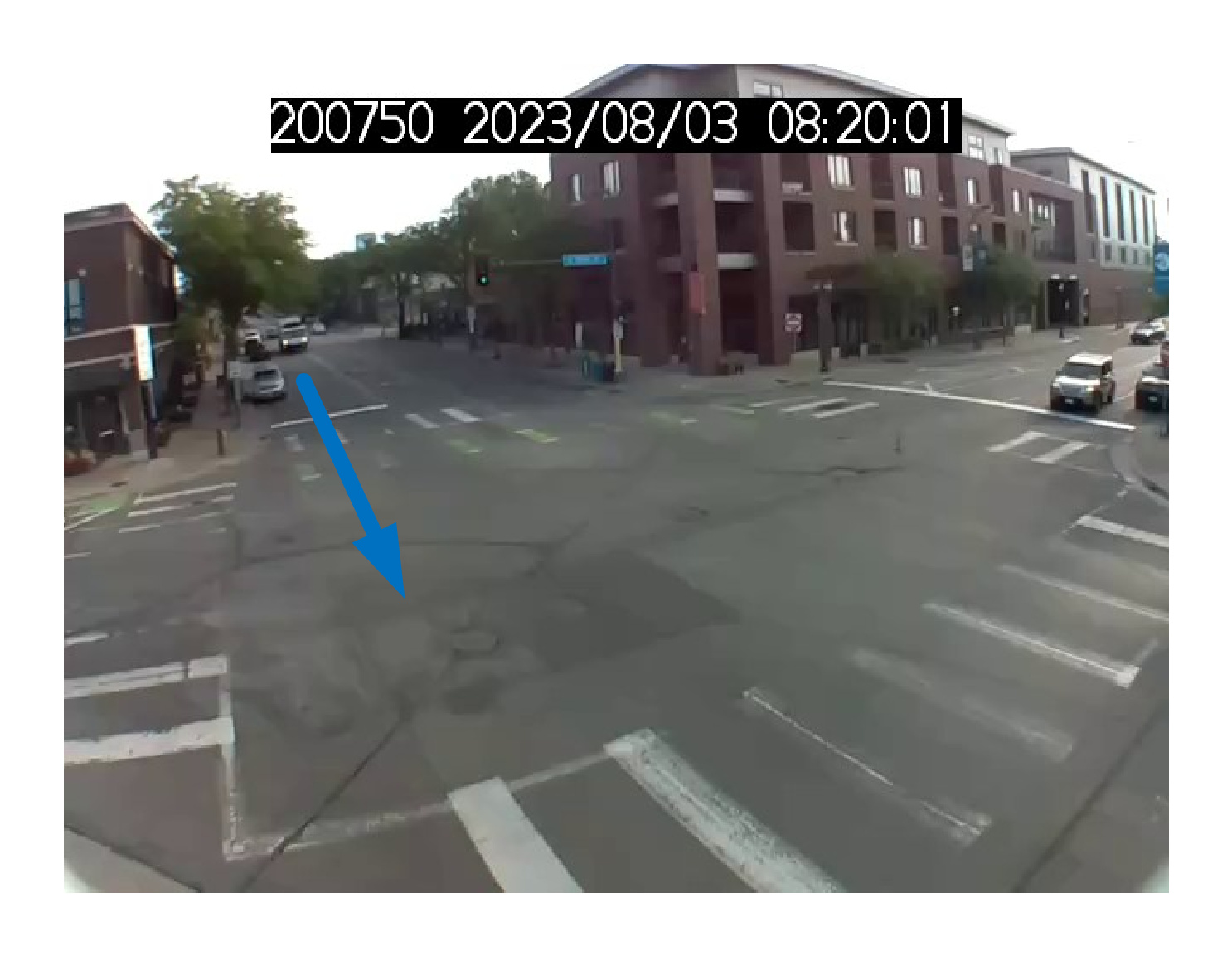} &
        \includegraphics[width=\imgW, height=\imgH, trim=50 50 50 50, clip]{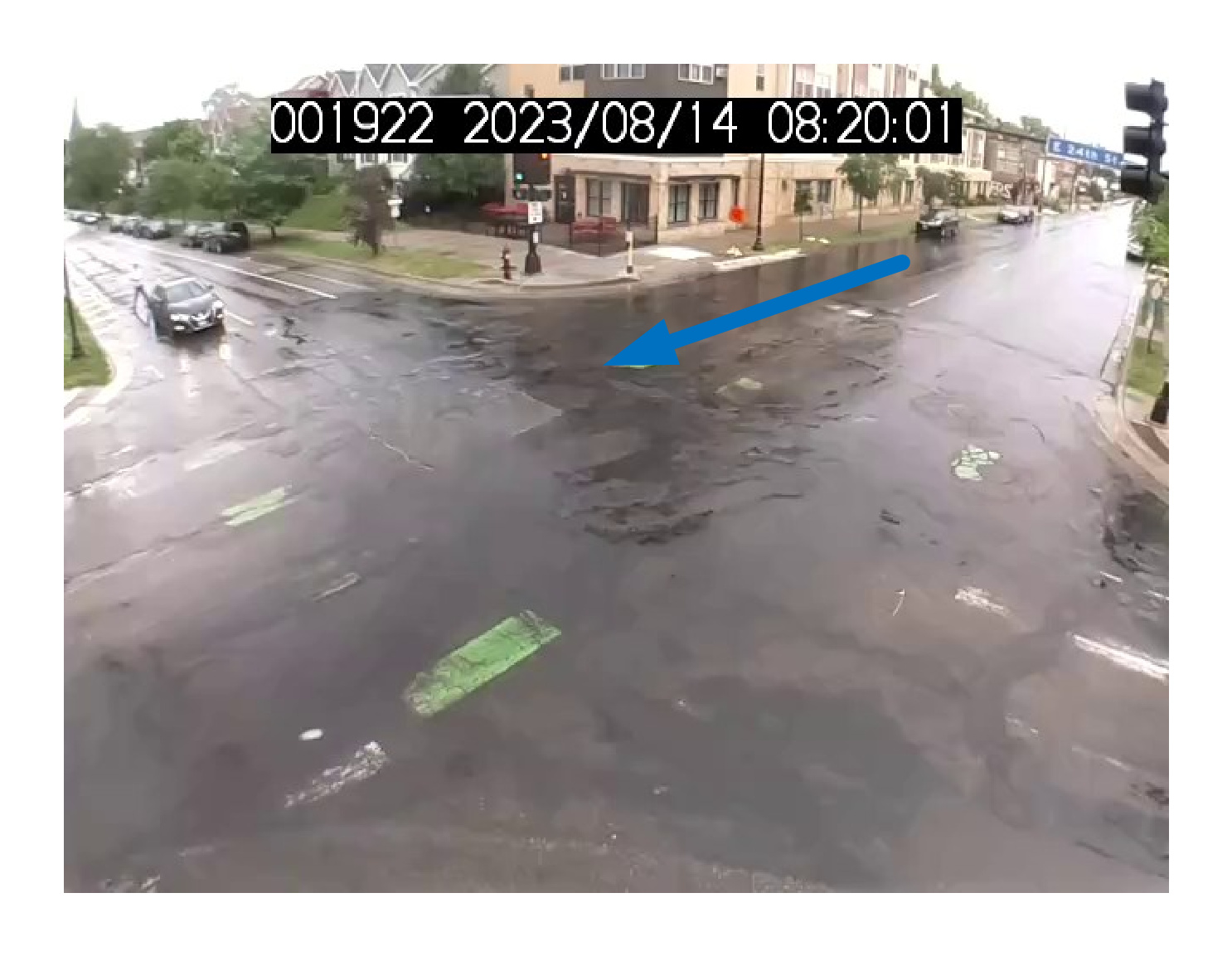} \\
        \rotatebox{90}{\hspace{7mm}\textbf{After}} &
        \includegraphics[width=\imgW, height=\imgH, trim=50 50 50 50, clip]{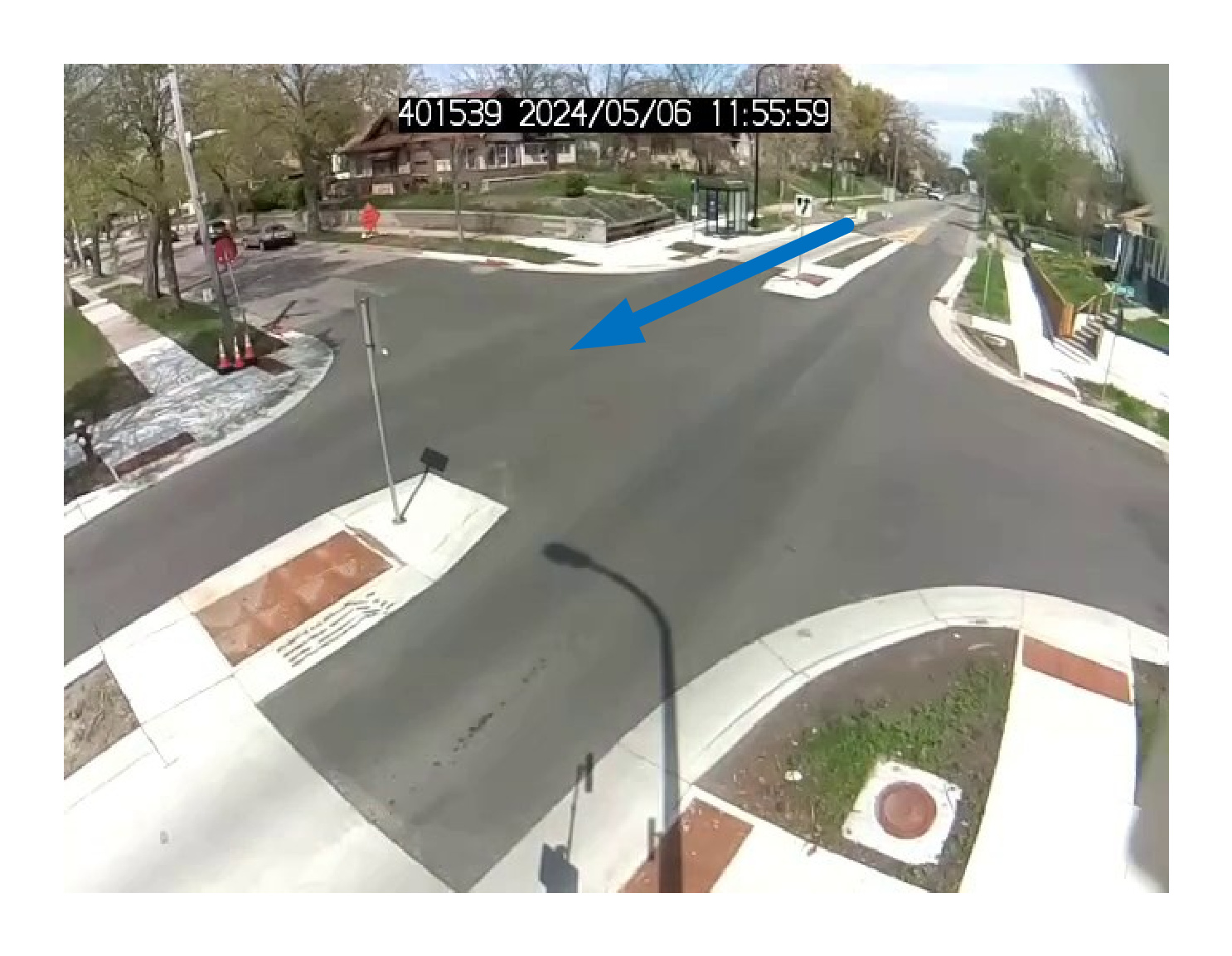} &
        \includegraphics[width=\imgW, height=\imgH, trim=50 50 50 50, clip]{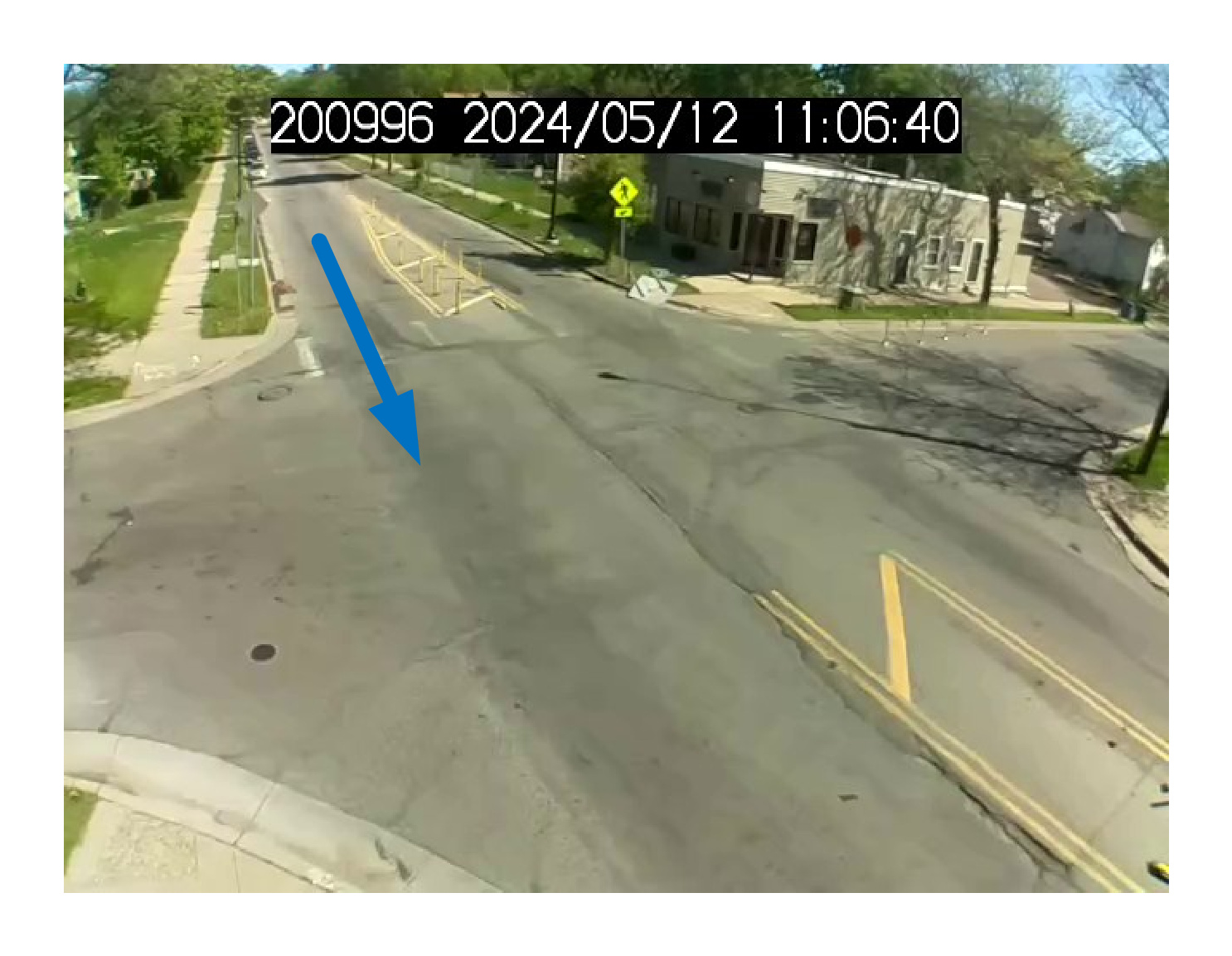} &
        \includegraphics[width=\imgW, height=\imgH, trim=50 50 50 50, clip]{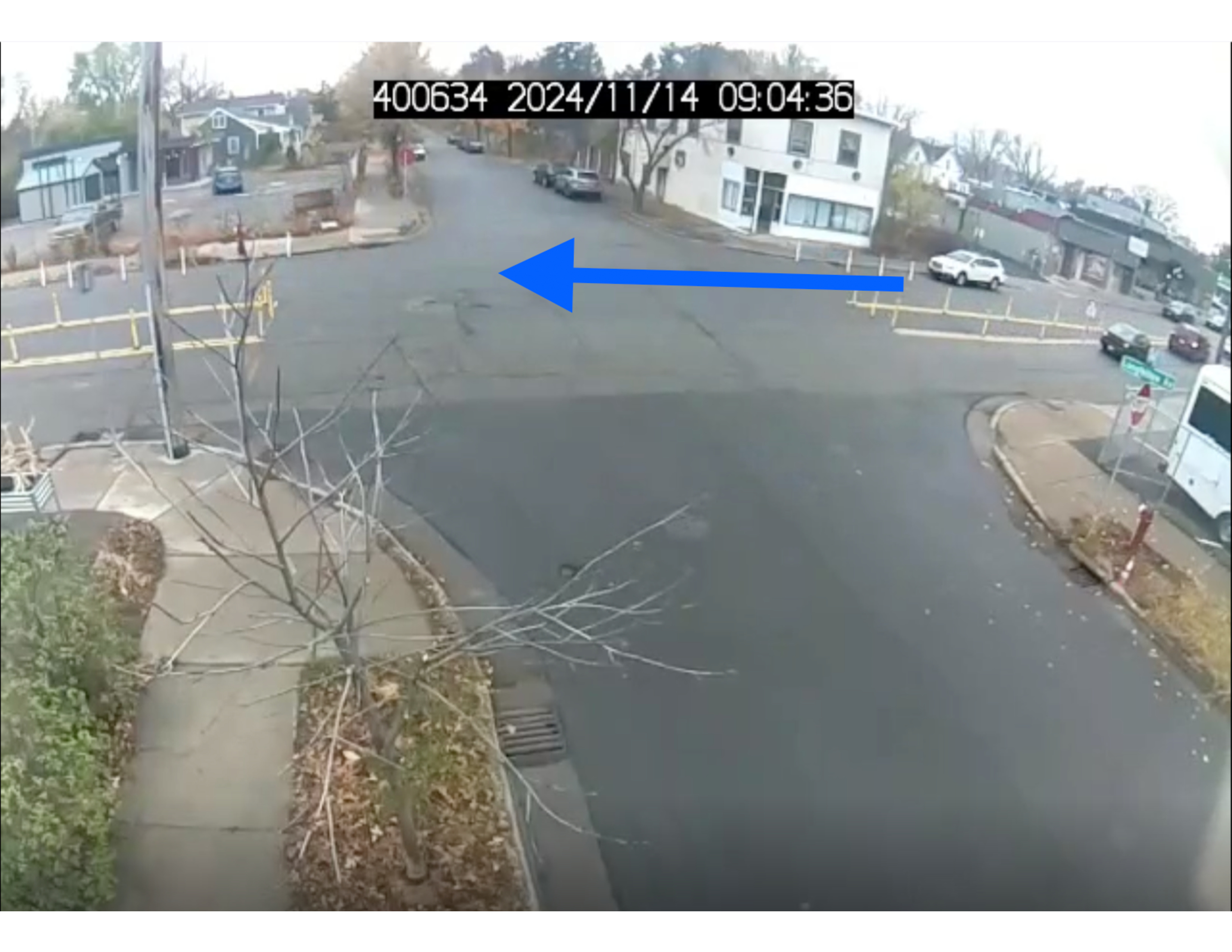} &
        \includegraphics[width=\imgW, height=\imgH, trim=50 50 50 50, clip]{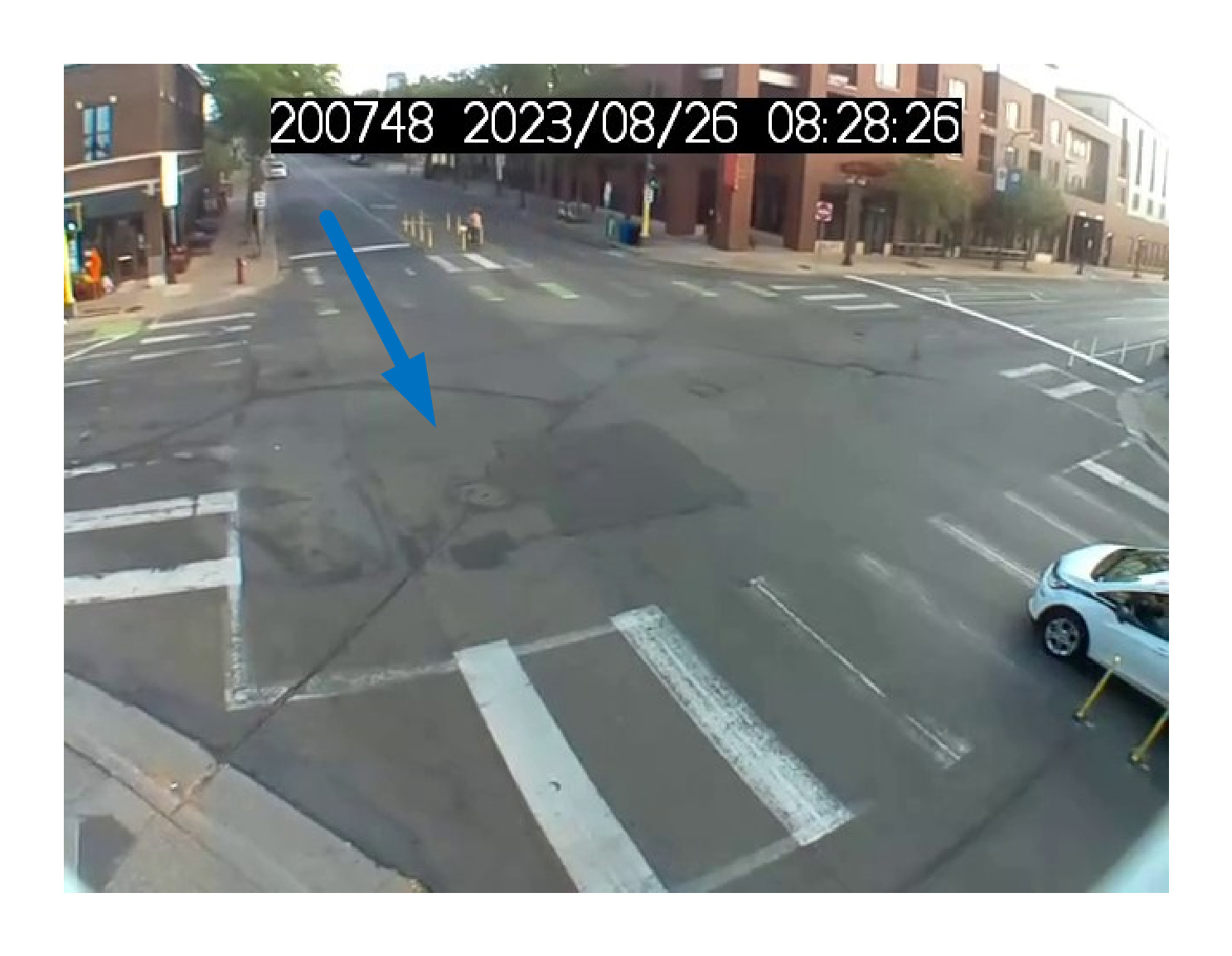} &
        \includegraphics[width=\imgW, height=\imgH, trim=50 50 50 50, clip]{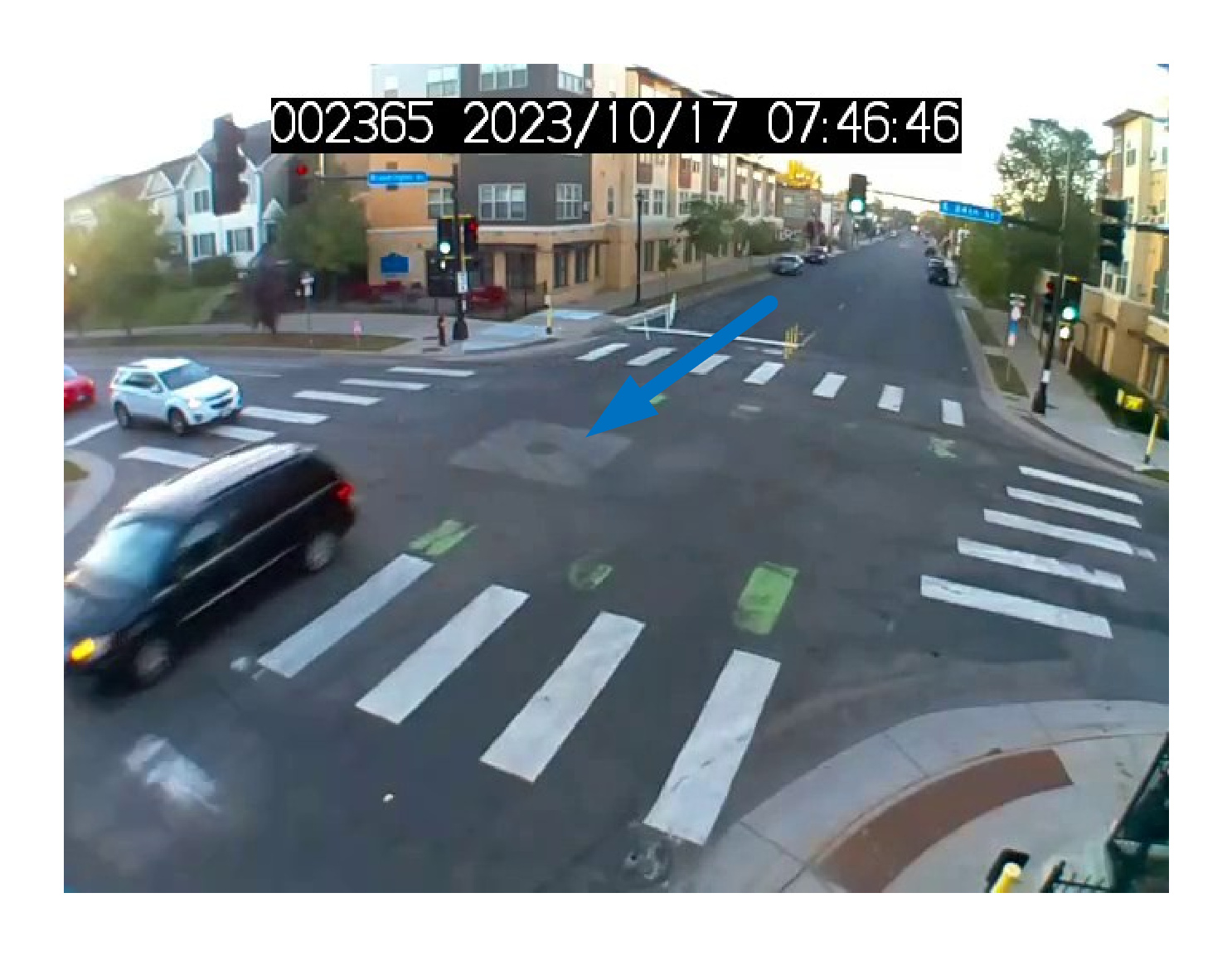} \\
        
        \multicolumn{6}{c}{} \\ \noalign{\vspace{0.5cm}} 
        
        % Block 2: Locations 6-9 (Aligned left with Block 1)
        & \textbf{Location 6} & \textbf{Location 7} & \textbf{Location 8} & \textbf{Location 9} & \\
        \rotatebox{90}{\hspace{7mm}\textbf{Before}} &
        \includegraphics[width=\imgW, height=\imgH, trim=50 50 50 50, clip]{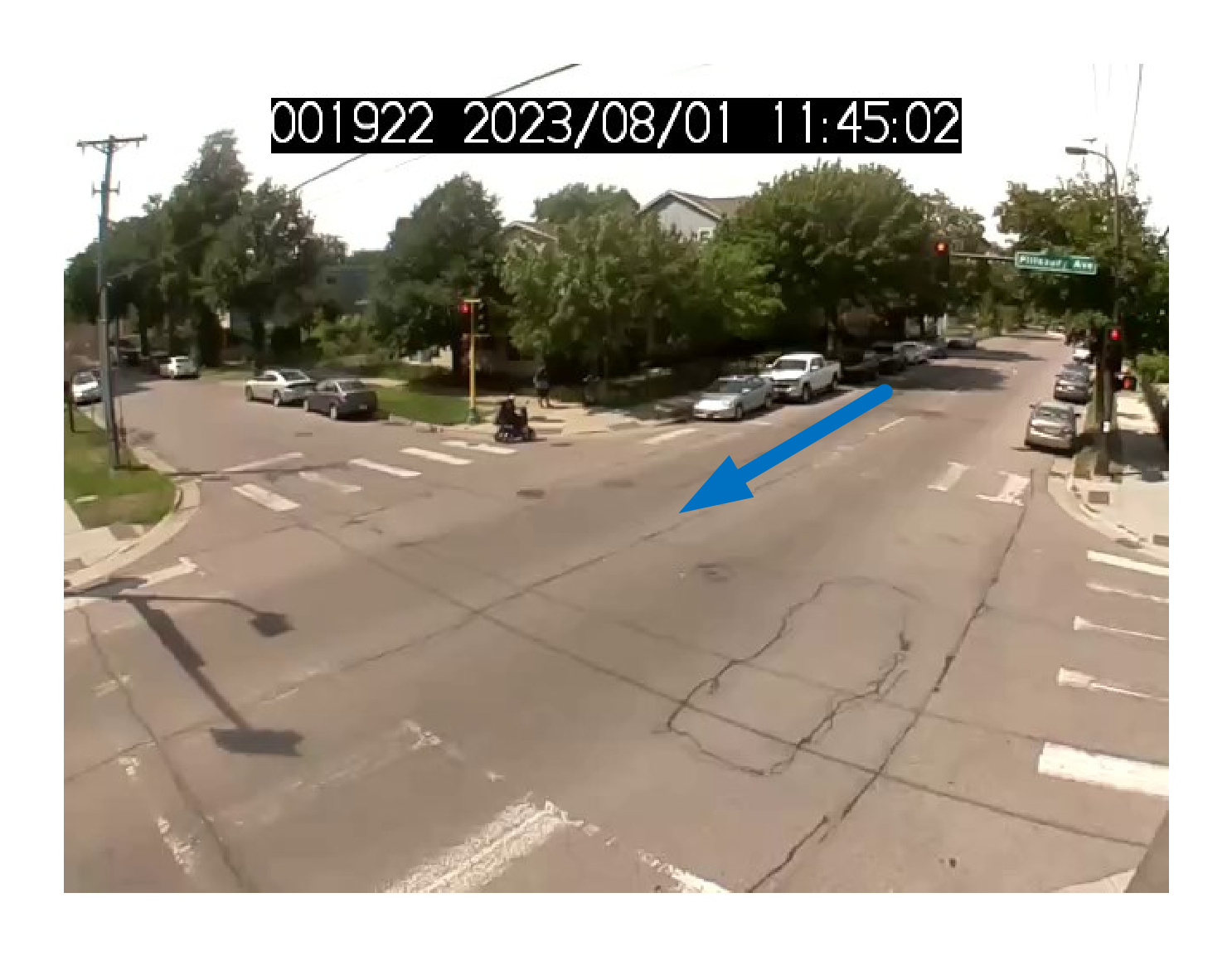} &
        \includegraphics[width=\imgW, height=\imgH, trim=50 50 50 50, clip]{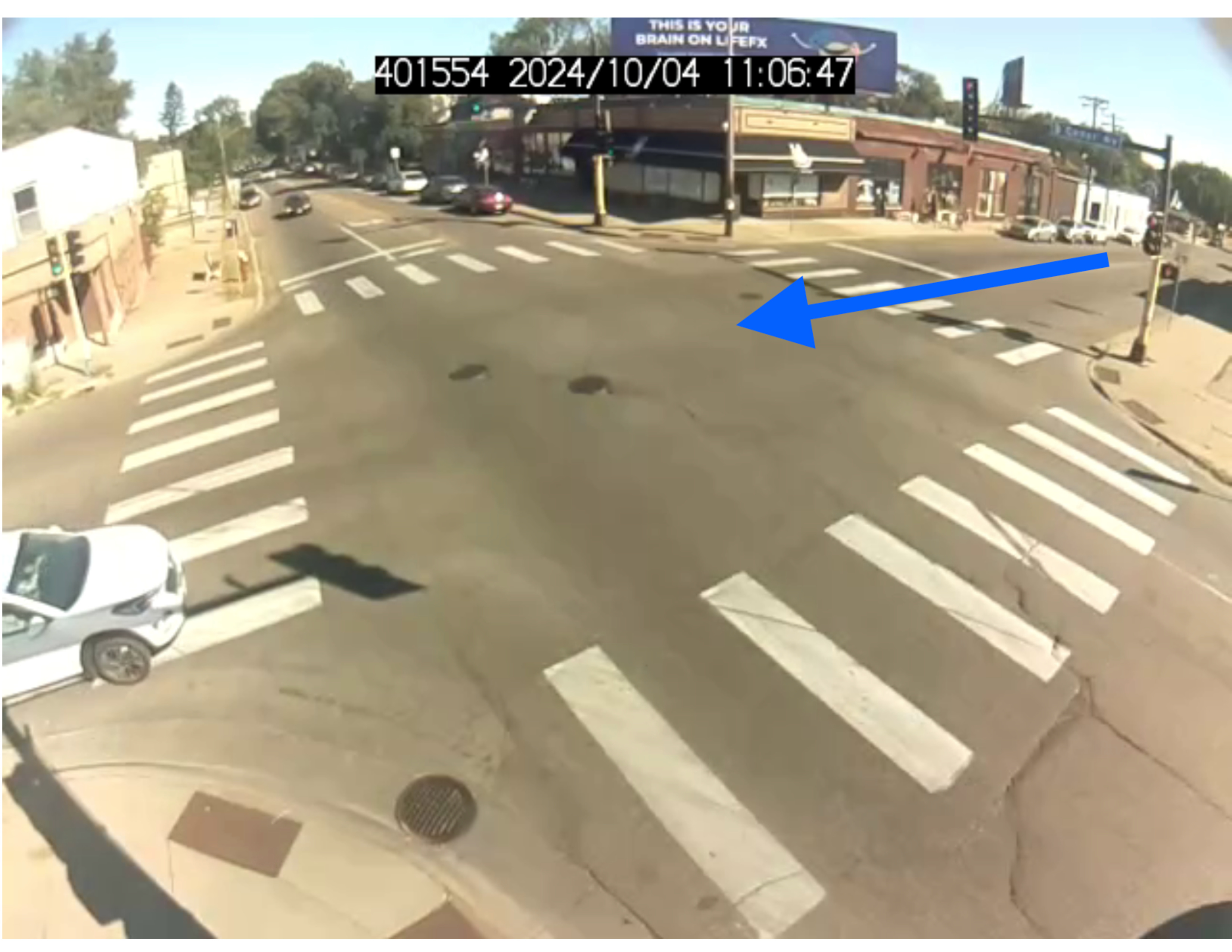} &
        \includegraphics[width=\imgW, height=\imgH, trim=50 50 50 50, clip]{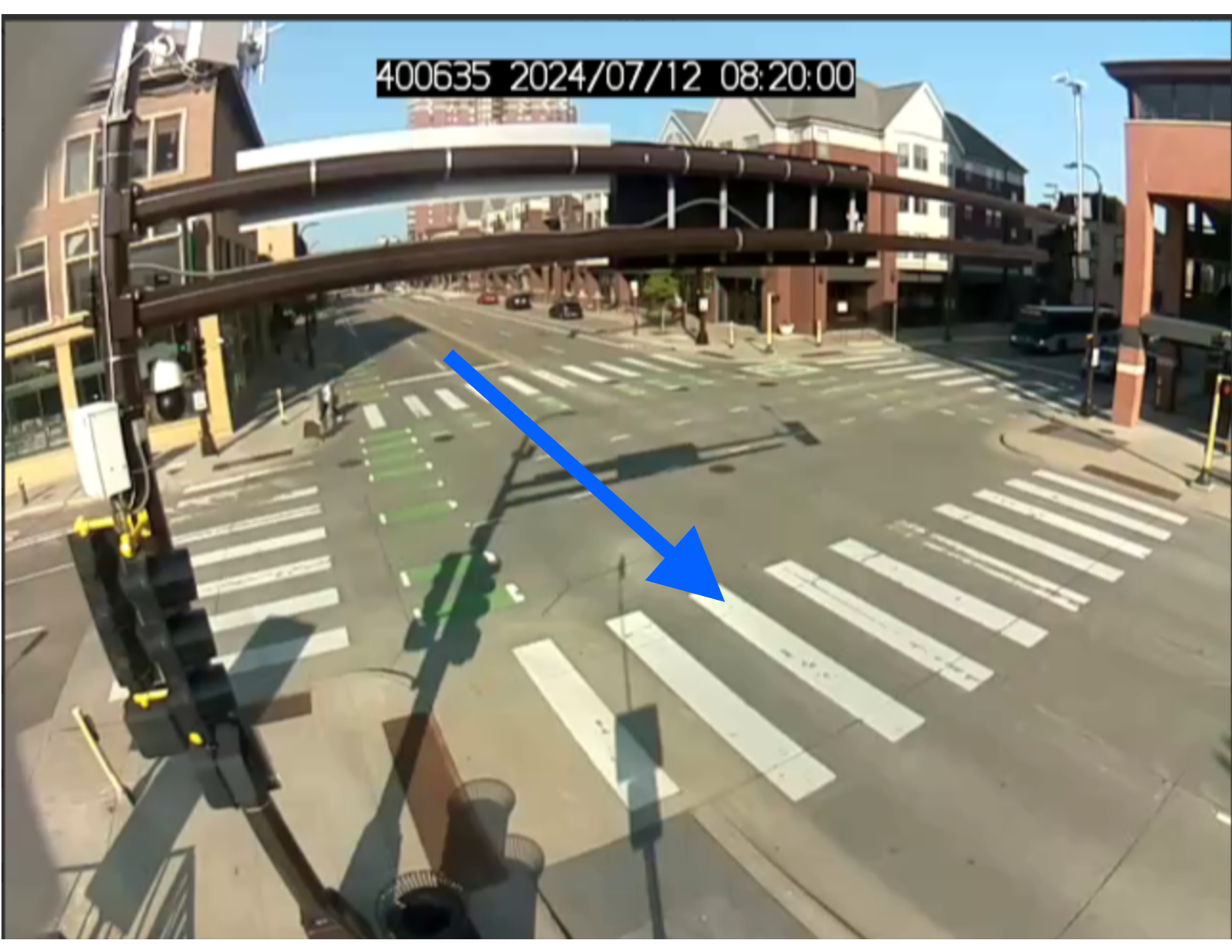} &
        \includegraphics[width=\imgW, height=\imgH, trim=50 50 50 50, clip]{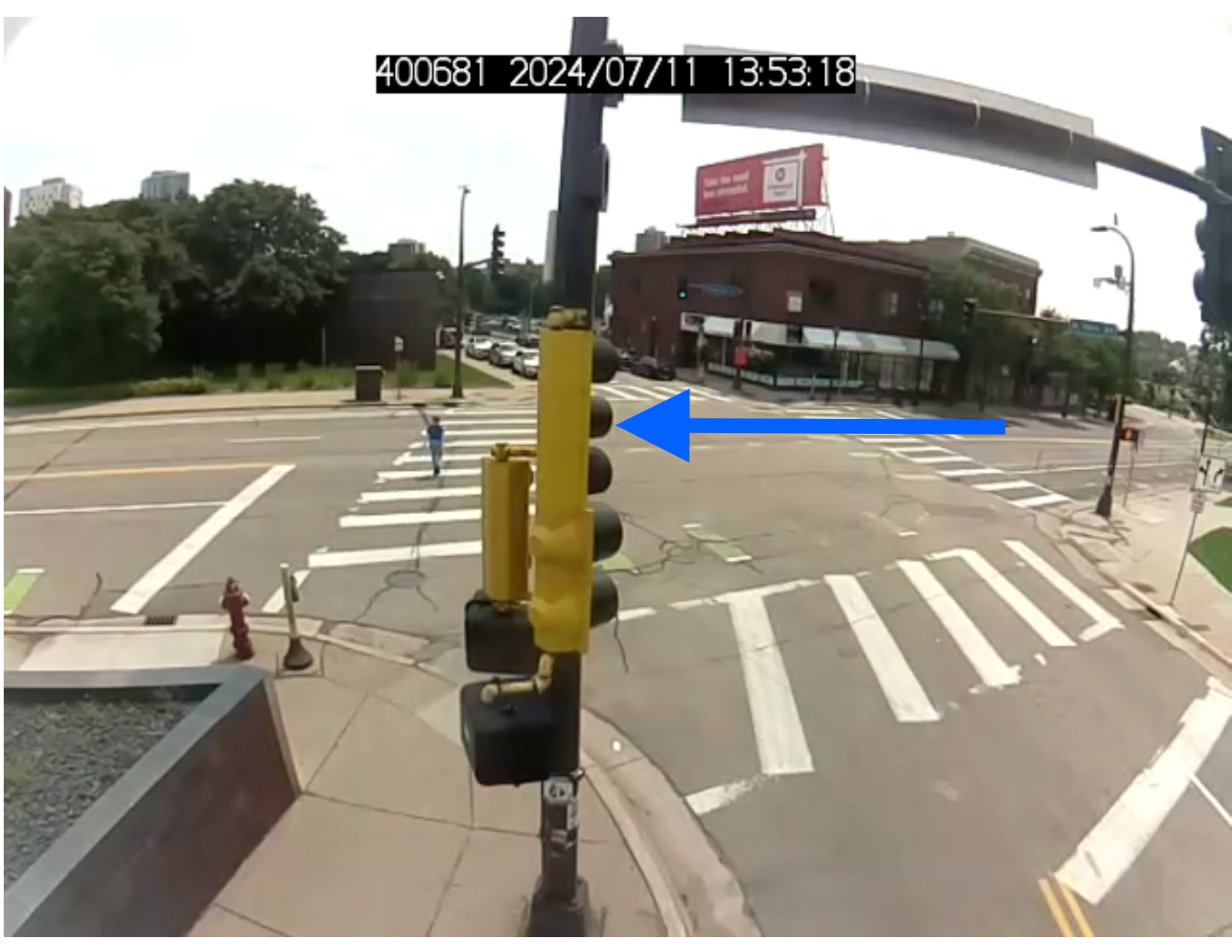} & \\
        \rotatebox{90}{\hspace{7mm}\textbf{After}} &
        \includegraphics[width=\imgW, height=\imgH, trim=50 50 50 50, clip]{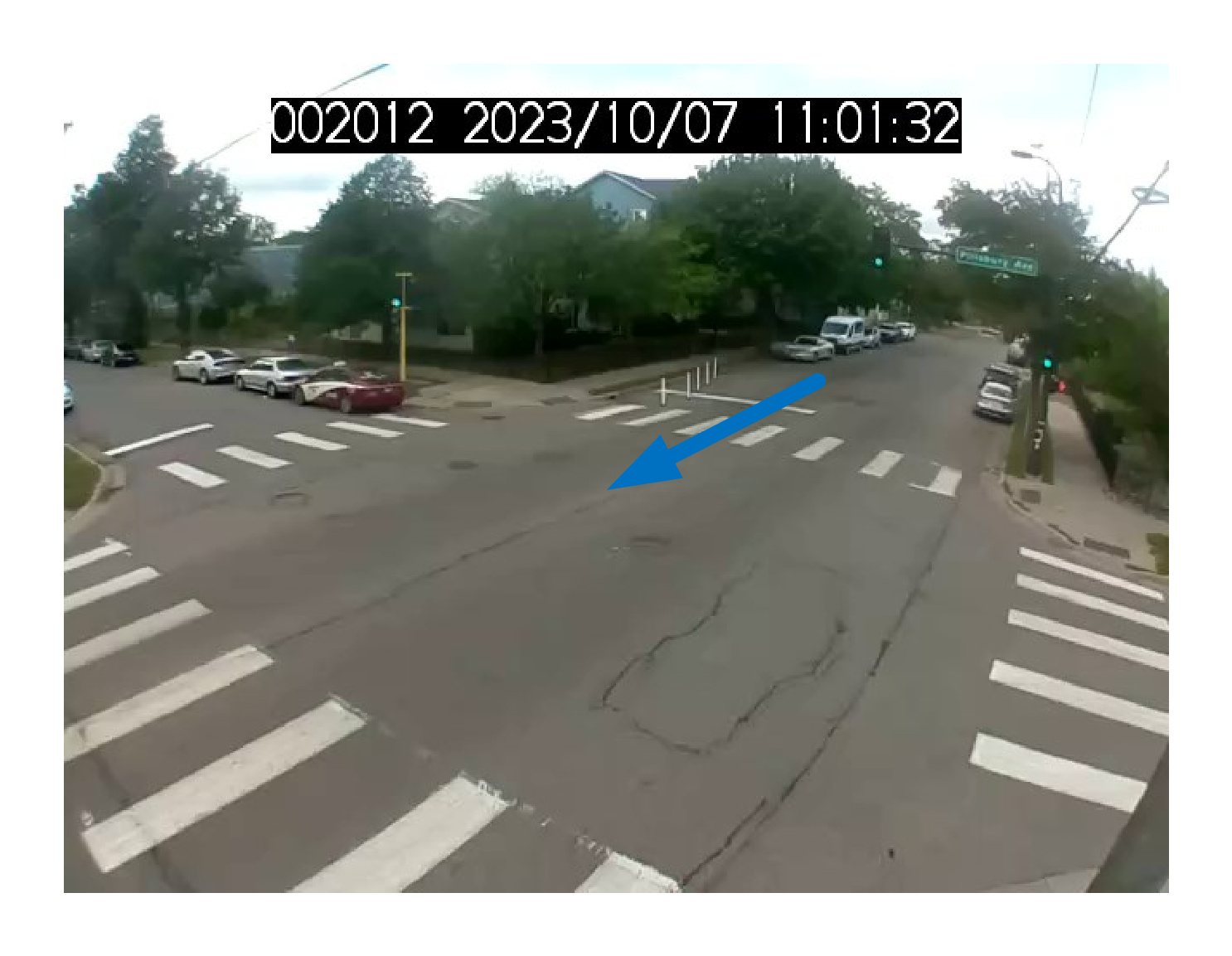} &
        \includegraphics[width=\imgW, height=\imgH, trim=50 50 50 50, clip]{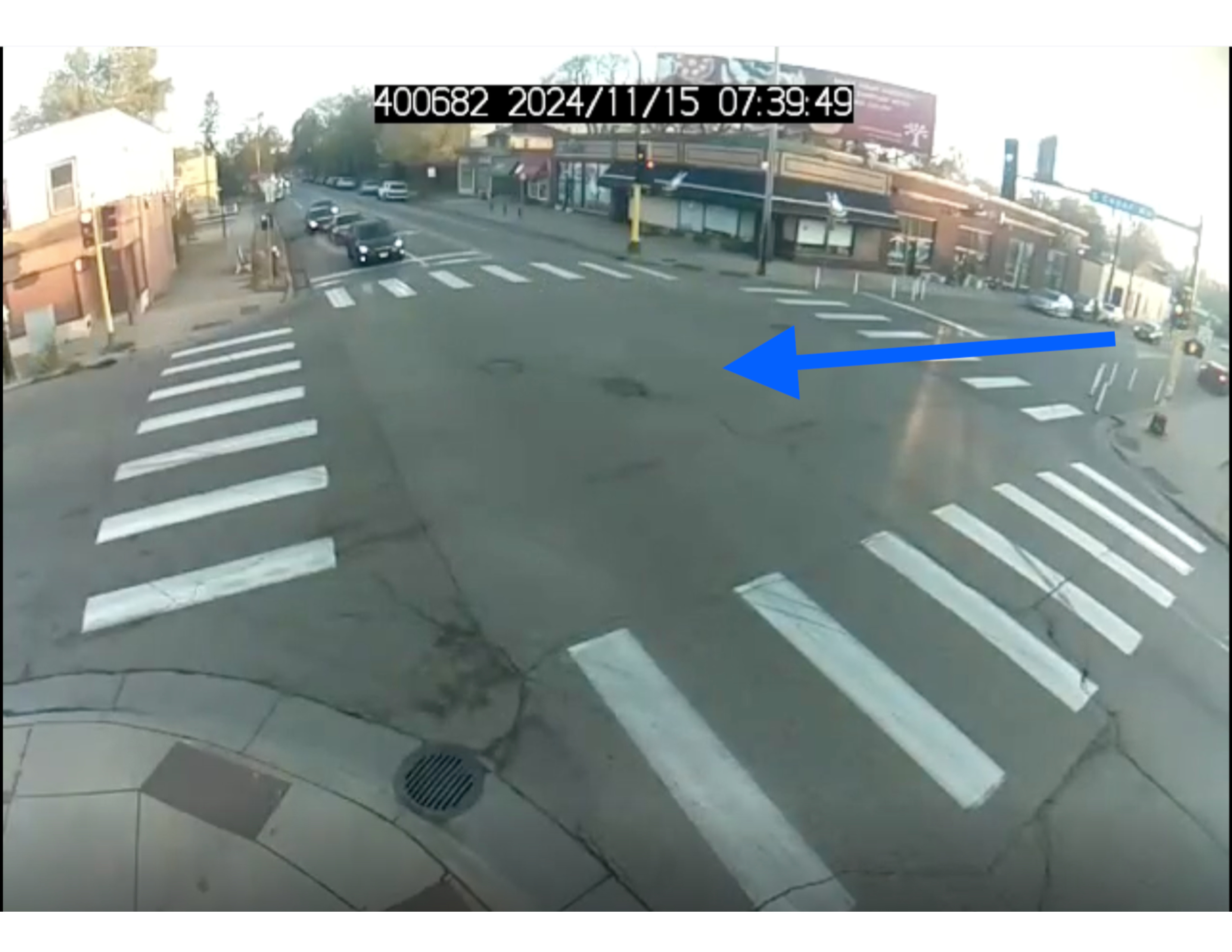} &
        \includegraphics[width=\imgW, height=\imgH, trim=50 50 50 50, clip]{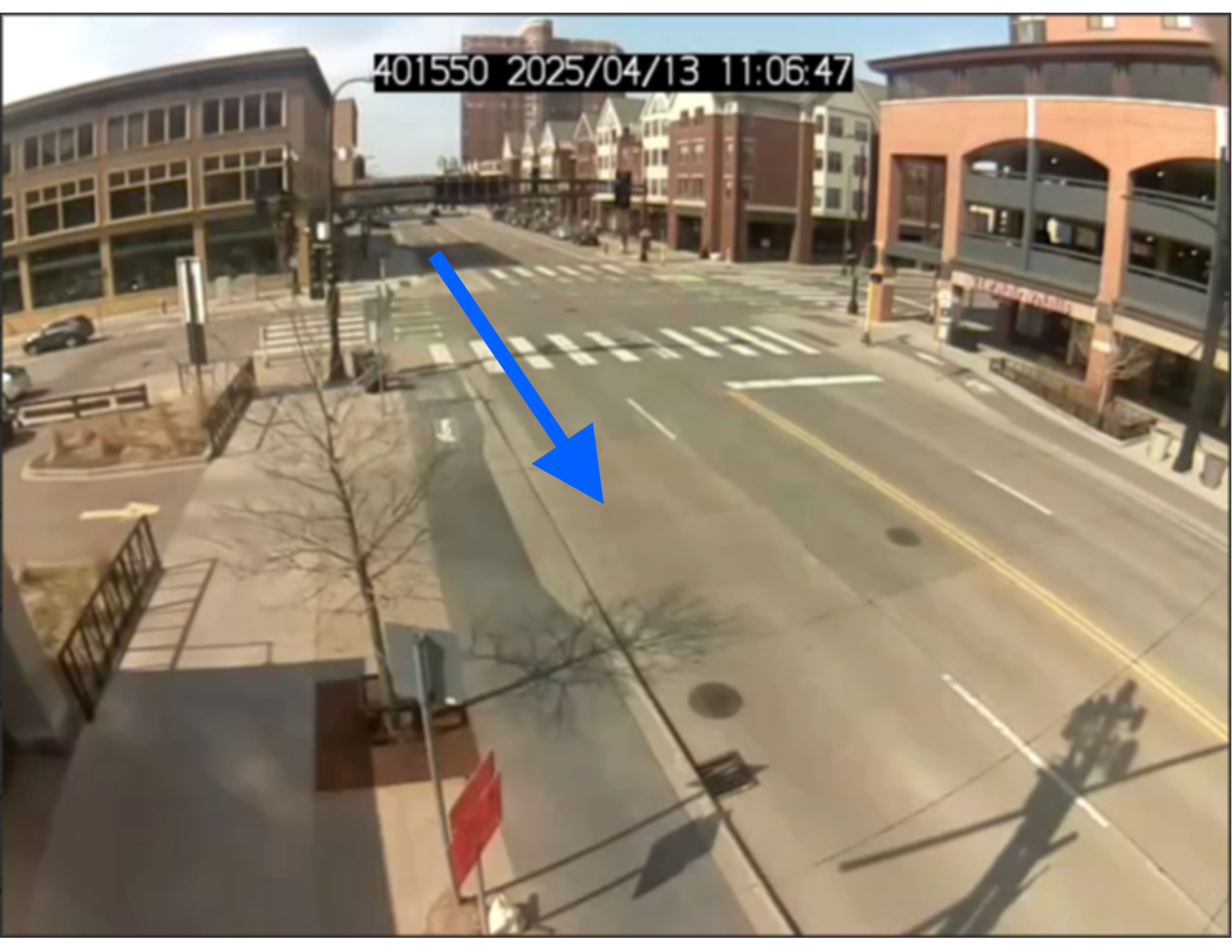} &
        \includegraphics[width=\imgW, height=\imgH, trim=50 50 50 50, clip]{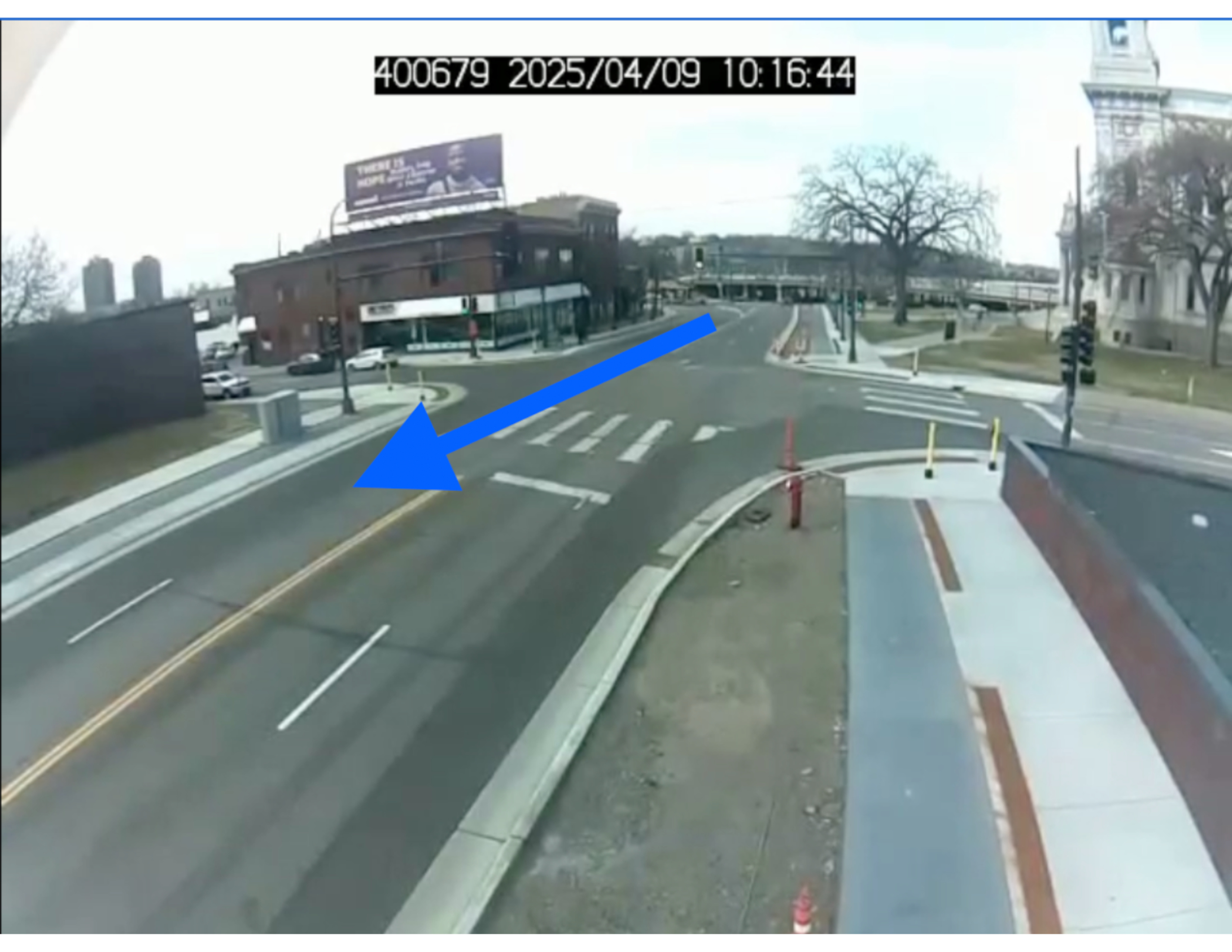} & \\
    \end{tabular}
    
    \caption{Before-and-after comparison of infrastructure changes for all nine locations. The top block shows Locations~1--5, and the bottom block shows Locations~6--9. For each site, the upper row displays pre-installation conditions and the lower row shows the post-installation interventions.}
    \label{fig:before_after_comparison}
\end{figure*}

We employed a set of infrastructure modifications at various locations, as summarized in Table~\ref{tab:location_list}. These sites were selected to capture a diverse range of traffic conditions, intersection controls, and pedestrian activity. Although the primary focus is on soft (quick-build) treatments, we include a small number of permanent/hardened installations as baseline references to reflect current practice. Because the number of permanent sites is limited and installations are not randomized, we do not claim causal ‘hard vs soft’ effects; comparisons are descriptive and site-specific. Specifically, Location~1, 8, and~9 feature hard infrastructure changes in the form of a Permanent Pedestrian Refuge and Permanent Curb Extension, while the rest of the Locations implements soft modifications, such as Temporary Curb Extension and Temporary Pedestrian Refuge, that use traffic bollards that are inexpensive and easier to install. To evaluate the impact of these changes, Fig.~\ref{fig:before_after_comparison} presents grouped comparisons before-and-after installation at all nine locations.

\begin{table*}[ht]
\small
\caption{Table with locations and their properties}
\label{tab:location_list}
\centering
\begin{adjustbox}{width=\textwidth}
\begin{tabular}{
    >{\centering\arraybackslash}p{3.5cm} |
    >{\centering\arraybackslash}p{1.2cm} |
    >{\centering\arraybackslash}p{2.5cm} |
    >{\centering\arraybackslash}p{5cm} |
    >{\centering\arraybackslash}p{2.5cm}
}
\hline
\textbf{Location} & 
\textbf{Loc. ID} & 
\textbf{Intersection Type} & 
\textbf{Installation Type} & 
\textbf{Direction} \\
    \hline
    N Lyndale Ave and 34th St & 1 & Unsignalized & Permanent Pedestrian Refuge & Southbound \\
    \hline
    N Lyndale Ave and 33rd St & 2 & Unsignalized & Temporary Curb Extension & Southbound \\
    \hline
    Longfellow Ave and 38th St & 3 & Unsignalized & Temporary Pedestrian Refuge + Temp Curb Extension & Eastbound \\
    \hline
    Nicollet Ave and 26th St & 4 & Signalized & Temporary Pedestrian Refuge & Southbound \\
    \hline
    Bloomington Ave and 24th St & 5 & Signalized & Temporary Curb Extension + Hardend Centerline & Northbound \\
    \hline
    Pillsbury Ave and 31st St & 6 & Signalized & Temporary Curb Extension & Eastbound \\
    \hline
    Cedar Ave and 38th St & 7 & Signalized & Temporary Curb Extension & Westbound \\
    \hline
    Hennepin Ave and 12th St & 8 & Signalized & Permanent Curb Extension & Eastbound \\
    \hline
    Maple Ave and Hennepin Ave & 9 & Signalized & Permanent Curb Extension & Eastbound \\
    \hline
\end{tabular}
\end{adjustbox}
\end{table*}

\subsection{Data Collection}
The sites discussed in Table \ref{tab:location_list} were selected to capture a diverse range of traffic conditions, intersection controls, and weather conditions. Video data were collected from manually installed CCTV cameras at each location, recording from 7:00 AM to 6:30 PM, with frame rates varying between 10 and 25 fps. In addition, pre-installation footage was collected prior to the deployment of the interventions at each site. 

For Locations 1, 2, 4, 5, and 6, footage was recorded in August 2023, whereas for Locations 3, 7, 8, and 9 was obtained in 2024; July for Locations 8 and 9, and October for Locations 3 and 7. Post-installation footage was then acquired at staggered times: in May 2024 for Locations 1 and 2, in August 2023 for Location 4, in October 2024 for Locations 5 and 6, in November 2024 for Locations 3 and 7, and in April 2025 for Locations 8 and 9. To mitigate the influence of seasonal variations on driving behavior, particularly for the November video collection, analysis was strictly limited to days with no snow/ice on the roads, ensuring that road surface friction was comparable to the pre-installation summer baseline. 

However, the usability of some of the footage was limited by camera angles that missed required intersections, adverse weather conditions such as rain, and blurry video quality, reducing the amount suitable for analysis. We planned to analyze 10 locations but included 9 due to challenges such as camera alignment, occlusion, and other site constraints. This can be observed in Tables \ref{tab:Unsignalized_samples} and \ref{tab:signalized_samples}, which show a lower number of usable video hours for Locations 2 and 4 resulting in low number of vehicles for evaluation. This highlights the real-world challenges, but enables a detailed examination of vehicle speeds, stop-and-go behavior, and pedestrian interactions under varying infrastructural and operational conditions. By incorporating multiple sites with different control mechanisms, layouts and environmental conditions, we aim to provide a assessment of how AI-based analytics can support pedestrian safety evaluations in real-world urban settings.

\begin{table}[ht]
\centering
\caption{ Summary of traffic video data collected at three unsignalized intersections across Pre-Installation and Post-Installation phases. The table lists the total vehicle sample size ($Smp$) and recording duration ($Hr$), with the Post-Installation data further categorized by Week 1 and Week 2 to analyze short-term temporal variances in traffic flow.}
\label{tab:Unsignalized_samples}
\footnotesize
\begin{tabular}{|c|c|c|c|c|c|c|}
\hline
\multicolumn{1}{|c|}{\textbf{Loc.}} & \multicolumn{2}{c|}{\textbf{Pre Installation}} & \multicolumn{4}{c|}{\textbf{Post Installation}} \\ \cline{2-7}
\textbf{ID} & \textbf{Smp} & \textbf{Hr} & \textbf{Smp} & \textbf{Hr} & \textbf{Smp} & \textbf{Hr} \\ \hline
1 & 13031 & 81.5 & 11659 & 81 & 11539 & 73.5 \\ \hline
2 & 11802 & 60 & 11395 & 60 & 14026 & 73.25 \\ \hline
3 & 21450 & 92 & 13784 & 73.5 & 2512 & 55.7 \\ \hline
\end{tabular}
\end{table}

\begin{table}[ht]
\centering
\caption{Summary of traffic video data collected at six signalized intersections across Pre-Installation and Post-Installation phases. The table lists the total vehicle sample size (Smp) and recording duration (Hr), with the Post-Installation data further categorized by Week 1 and Week 2 to analyze short-term temporal variances in traffic flow.}
\label{tab:signalized_samples}
\footnotesize
\begin{tabular}{|c|c|c|c|c|c|c|}
\hline
\multicolumn{1}{|c|}{\textbf{Loc.}} & \multicolumn{2}{c|}{\textbf{Pre Installation}} & \multicolumn{4}{c|}{\textbf{Post Installation}} \\ \cline{2-7}
\textbf{ID} & \textbf{Smp} & \textbf{Hr} & \textbf{Smp} & \textbf{Hr} & \textbf{Smp} & \textbf{Hr} \\ \hline
4 & 1673 & 17 & 6604 & 72 & 3820 & 42 \\ \hline
5 & 11267 & 79 & 3966 & 42.5 & 1646 & 23 \\ \hline
6 & 12020 & 89 & 7935 & 72 & 10648 & 72 \\ \hline
7 & 21477 & 92 & 9173 &  71 & 6918 &  52.5 \\ \hline
8 & 15585 & 68.25 & 12285 &  44.6 & 7371 &  31.1 \\ \hline
9 & 22469 & 89.25 & 15863 &  37.5 & 8915 &  34 \\ \hline
\end{tabular}
\end{table}

\subsection{AI Pipeline}
The AI pipeline used for speed-based analytics approach, builds upon the AI-based analytics framework introduced in \cite{katariya2024vegaedge}, with modifications tailored to our specific application. We adopt a Single-frame Processing approach that integrates speed estimation, as detailed in the following sections. To maintain a focused analysis, the pipeline processes detections exclusively within a defined area of interest, filtering out irrelevant data outside this region. 

We fine-tuned the YOLOv8l model \cite{Jocher_YOLO_by_Ultralytics_2023} using the BDD100K vehicle dataset \cite{yu2020bdd100k} to enhance object detection accuracy. Next, we integrated the ByteTrack algorithm, known for its strong performance in handling object association across frames. ByteTrack effectively maintains object consistency even in challenging conditions such as occlusions and dense traffic by leveraging deep association techniques. As demonstrated in prior studies \cite{zhang2022bytetrack, katariya2024vegaedge}, ByteTrack performs exceptionally well on datasets like BDD100K and MOT20, offering both accuracy and efficiency. Its ability to deliver high-speed tracking without sacrificing precision makes it an ideal choice for our AI-based analytics framework. Because our focus is downstream behavioral metrics (speed and yielding) rather than proposing a new detector/tracker, we adopt an established YOLOv8 + ByteTrack pipeline from prior work \cite{katariya2024vegaedge} and use conservative filtering to mitigate occasional detection/tracking noise.

Once tracking is complete, the extracted bounding boxes and vehicle IDs are sent to the speed estimation module, where we calculate the vehicle's average speed based on the distance traveled within the defined area of interest. Various filters are applied to refine the dataset and ensure accurate analysis as follows:

\begin{enumerate}
    \item Vehicle Type Selection: We focus only on cars, buses, and trucks, eliminating potential noise from motorcycles, bicycles, and pedestrians.
    \item Stationary Vehicle Removal: Vehicles that remain stationary within the area of interest are excluded, as they do not contribute to speed analysis.
    \item Following Vehicle Filtering: Vehicles traveling too closely (within 40 pixels) behind another vehicle are removed, as their speed and behavior may be influenced by the leading vehicle, introducing bias in the data.
    \item Direction-Based Filtering: Only vehicles moving in predefined directions (as shown in Table \ref{tab:location_list} and Fig.~\ref{fig:before_after_comparison}) are considered. Vehicles traveling in other directions are excluded to ensure consistency in speed analysis.
\end{enumerate}

\subsection{Speed Estimation}

\begin{figure}[t!]
    \centering
    \includegraphics[width=0.99\columnwidth]{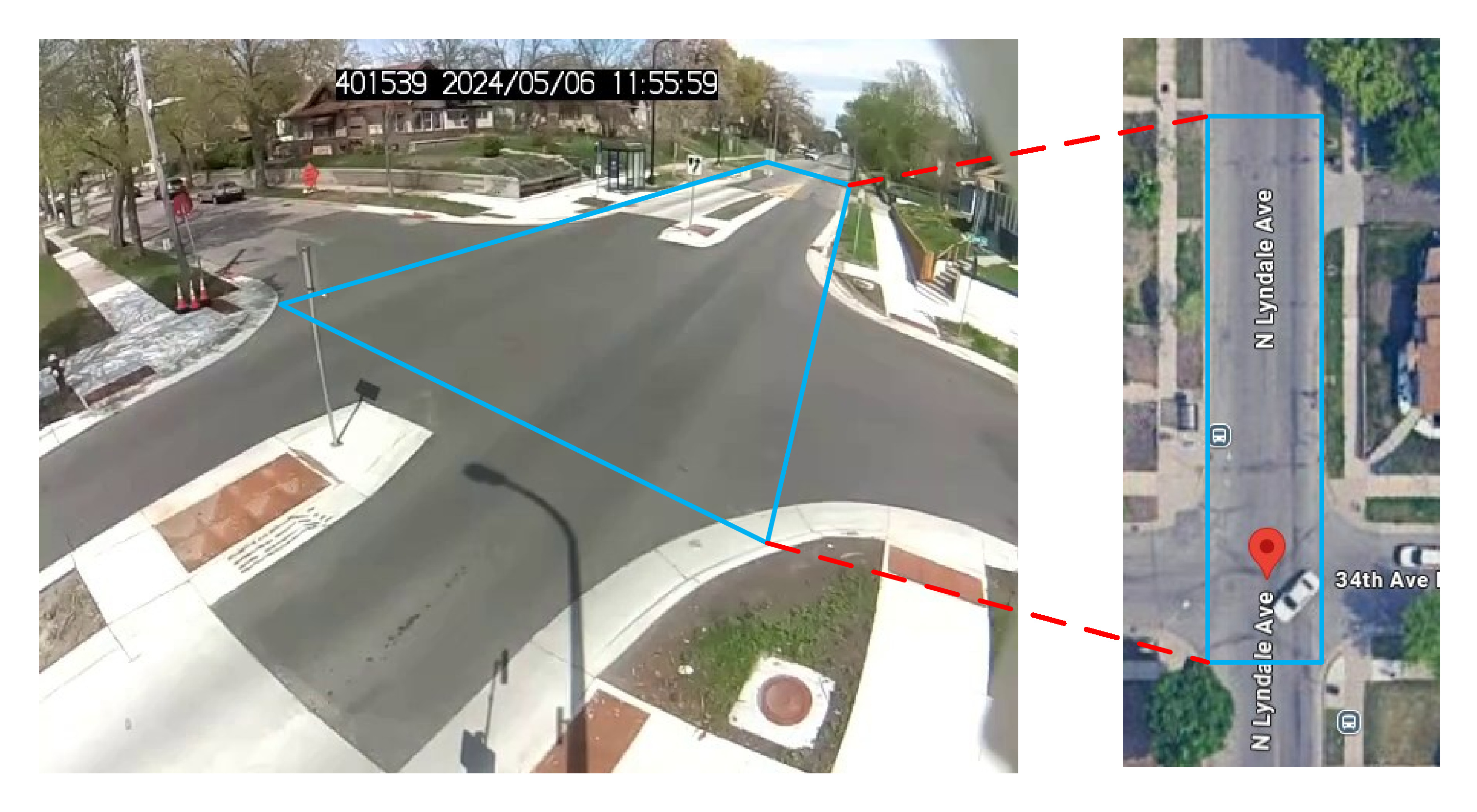}
    \caption{Mapping of image coordinates to real-world coordinates using perspective transformation for Site 1. 
    (Left) CCTV camera footage showing the selected reference points (black trapezium). 
    (Right) Corresponding real-world coordinates from satellite imagery, mapped to a rectangular region. 
    The red dashed lines indicate the homography transformation between the two views.}
    \label{fig:perspective_mapping}
\end{figure}

To accurately map pixel coordinates from an image to real-world dimensions, we employ a perspective transformation, similar to the approaches used by Mejia et al.\cite{mejia2021vehicle}, and Chen et al \cite{chen2021application}. This transformation ensures that detected vehicle coordinates in an image are converted to their corresponding real-world positions, allowing precise motion tracking and speed estimation. We used CCTV camera footage along with known real-world coordinates to establish a mapping between the image plane and the actual road environment, as shown in Fig. \ref{fig:perspective_mapping}. By selecting four or more reference points visible in both the image and the real-world coordinate system, we compute a homography matrix that enables the accurate localization of detected objects in real-world space. 

\subsubsection{Perspective Transformation}

The transformation between image coordinates and real-world coordinates is achieved using a homography matrix $\mathbf{H}$. Given an image point $\mathbf{p} = (x, y, 1)^T$ and its corresponding real-world point $\mathbf{P} = (X, Y, 1)^T$ in homogeneous coordinates, the transformation is defined as:

\begin{equation}
\label{eq:perspective_transform}
    \lambda 
    \begin{bmatrix}
        x' \\ y' \\ 1
    \end{bmatrix} = 
    \mathbf{H}
    \begin{bmatrix}
        X \\ Y \\ 1
    \end{bmatrix}
\end{equation}

where $\lambda$ is a scaling factor, and the homography matrix $\mathbf{H}$ is a $3 \times 3$ matrix:

\begin{equation}
\label{eq:homography}
    \mathbf{H} =
    \begin{bmatrix}
        h_{11} & h_{12} & h_{13} \\
        h_{21} & h_{22} & h_{23} \\
        h_{31} & h_{32} & h_{33}
    \end{bmatrix}
\end{equation}

This transformation is computed using four or more point correspondences, solving the Direct Linear Transformation (DLT) problem to derive $\mathbf{H}$. 

\subsubsection{Speed Computation}

Once pixel coordinates are mapped to the real-world coordinate system, speed estimation is performed by tracking each vehicle’s displacement over multiple frames.

Vehicle speed was derived from the Euclidean displacement of the tracked centroid over a sliding window ($W \approx 1s$). Specifically, we compute displacement using the oldest and newest tracked positions within the last W frames, where W = min (tracked length (>0.5s), fps) (i.e., up to a ~1-second window:

\begin{equation}
\label{eq:distance}
    d = \sqrt{(X_t - X_0)^2 + (Y_t - Y_0)^2}
\end{equation}

where $(X_0, Y_0)$ denotes the vehicle position at the start of the sliding window (oldest retained point), $(X_t, Y_t)$ denotes the most recent position in that window.

The speed $v$ of the vehicle is then given by:

\begin{equation}
\label{eq:velocity}
    v = \frac{d}{\Delta t}
\end{equation}

where $\Delta t = \frac{t}{\text{fps}}$ represents the elapsed between the two positions used in the sliding window. We begin reporting speed only when the tracked trajectory is at least \( \frac{\mathrm{fps}}{2} \) frames long (\(\approx 0.5\) s).

The proposed speed estimation methodology ensures accurate and reliable velocity computations by incorporating a structured computational pipeline. The transformation of detected vehicle positions from pixel coordinates to real-world coordinates is achieved through a homography-based perspective transformation as in Eq. \ref{eq:perspective_transform}, where a computed transformation matrix $\mathbf{H}$ shown in Eq. \ref{eq:homography} maps the bottom center of the bounding box to a reference plane. Speed estimation using perspective transform approaches is a well-established methodology in traffic monitoring \cite{chen2021application}. Previous validation studies have demonstrated that when properly calibrated, this method achieves high precision with error margins as low as 0.56 to 1.43 mph \cite{rodriguez2022analysis}. In this study, we strictly adhered to these established calibration protocols, using static reference points, to ensure our speed estimates fall within these accepted accuracy bounds. By comparing the AI-estimated speeds with established benchmarks, we confirmed that the automated system accurately captures the direction and magnitude of speed changes.

To maintain temporal consistency in vehicle trajectory analysis, an object tracking mechanism is employed, wherein detected objects are assigned unique identifiers, and their positional history is recorded over multiple frames. This historical tracking ensures continuity in displacement calculations (Eq. \ref{eq:distance}), mitigating the effects of short-term occlusions and detection inaccuracies. The Euclidean distance between a vehicle’s initial and current positions within a designated time window is used to compute its real-world displacement, providing a basis for speed estimation.

Given that the accuracy of velocity computations is contingent on precise time measurement, the elapsed time $\Delta t$ is derived from the video frame rate (fps) and the number of frames elapsed between consecutive position updates. Finally, the speed is calculated using Eq. \ref{eq:velocity}. Additionally, the AI-based analytics framework dynamically updates the coordinate history of tracked vehicles, ensuring that only sufficiently long trajectories (at least 0.5 seconds long) are considered for speed estimation, thereby mitigating potential inaccuracies due to transient detection artifacts. The modularity of this approach allows for seamless integration into larger AI-based analytics pipelines. The combination of perspective correction, robust object tracking, and calibrated speed estimation ensures that the methodology remains adaptable to a wide range of urban surveillance environments.

For both signalized and unsignalized intersections, we examine whether these interventions lead to reductions in mean and 85th percentile vehicle speeds. Additionally, for unsignalized intersections, we analyzed vehicle behavior near the intersection to capture pre-and-post-intervention trends, to assess if vehicles slow down after infrastructure modifications improving pedestrian safety or if there is no change. We also investigate whether these effects are sustained over time or if speeds rebound after an initial adjustment period.

\section{Results}
In this section, we present the results of our before-and-after evaluation of vehicle speeds and driver behavior at both signalized and unsignalized intersections. We begin with a high-level summary of speed trends across all sites, followed by detailed analyses for each intersection type. Our evaluation focuses on the immediate impact of the infrastructure changes (Week 1) and the observed behavioral adjustments (Week 2).

Sites include predominantly soft (quick-build) treatments, with a small number of permanent/hardened baselines used for contextual comparison; results are reported by site and are not intended to estimate causal differences between treatment types. Notably, the locations with permanent infrastructure modifications (Locations 1, 8, and 9) also show measurable speed reductions similar to the trends observed at sites with soft quick-build treatments.

\subsection{Analysis of Unsignalized Intersections}
\begin{figure*}[t!]
    \centering
    \scriptsize
    \begin{tabular}{c c c}
        \textbf{Pre-installation} & \textbf{Post-installation Week 1} & \textbf{Post-installation Week 2} \\
        \includegraphics[width=0.32\textwidth]{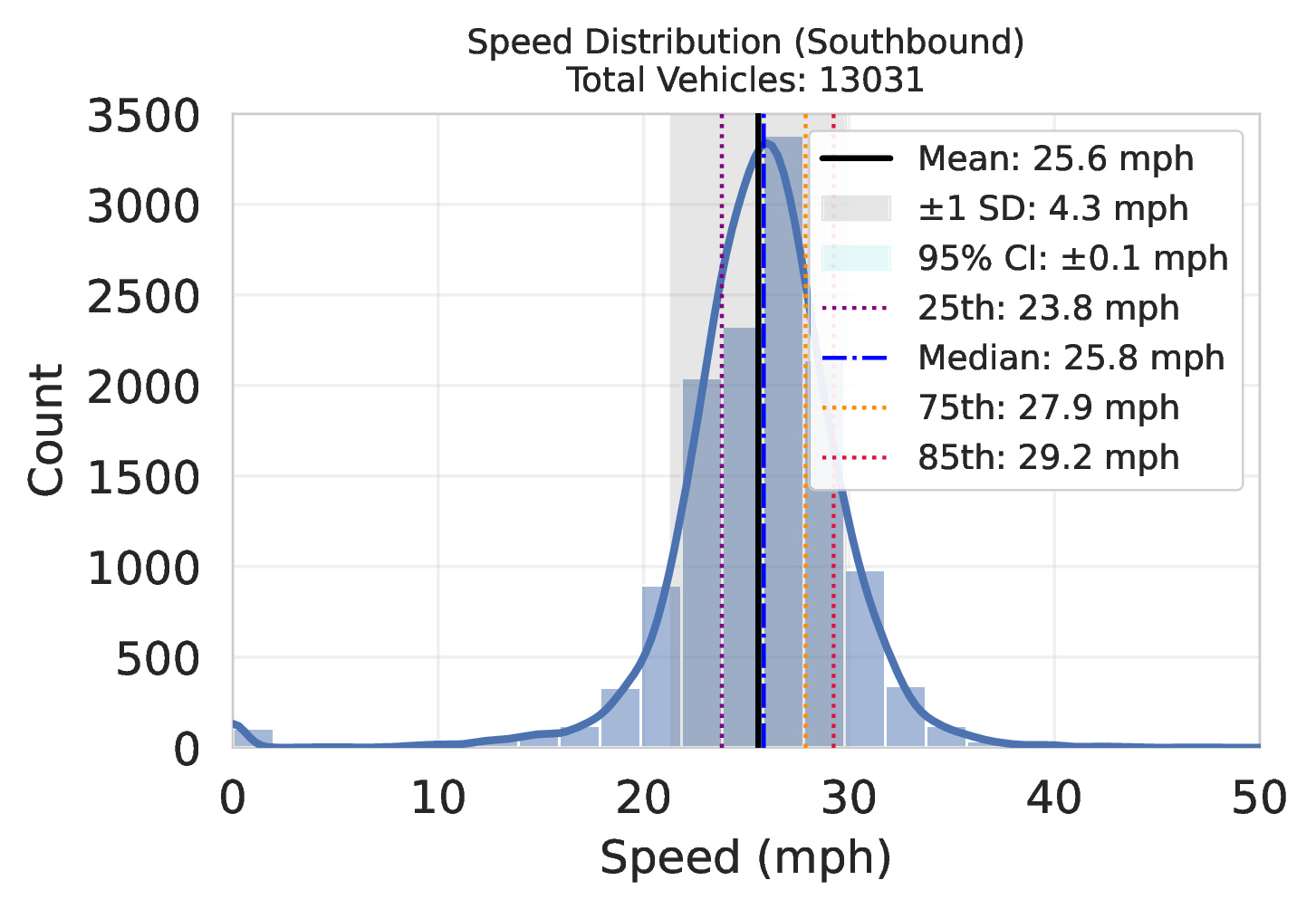} &
        \includegraphics[width=0.32\textwidth]{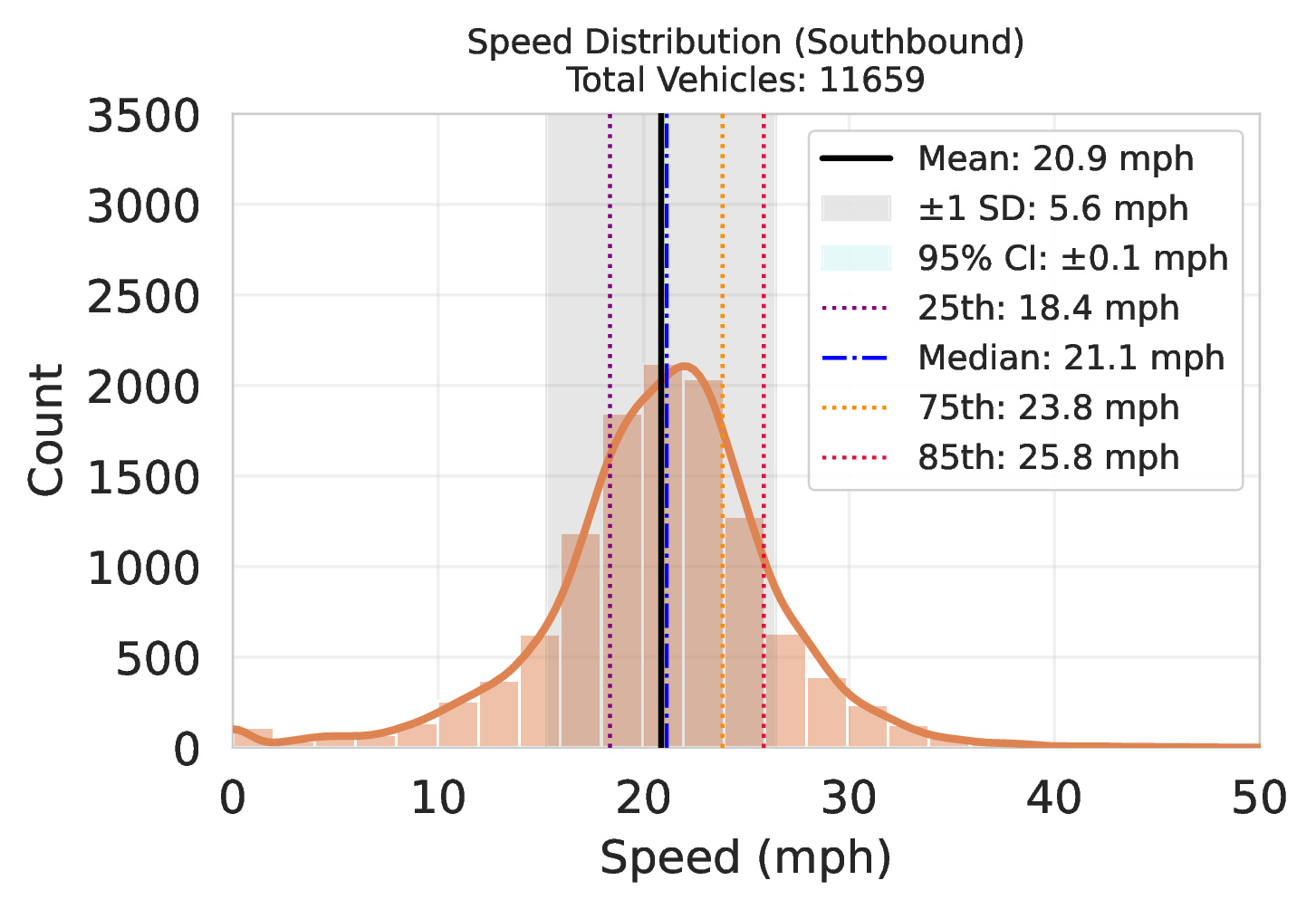} &
        \includegraphics[width=0.32\textwidth]{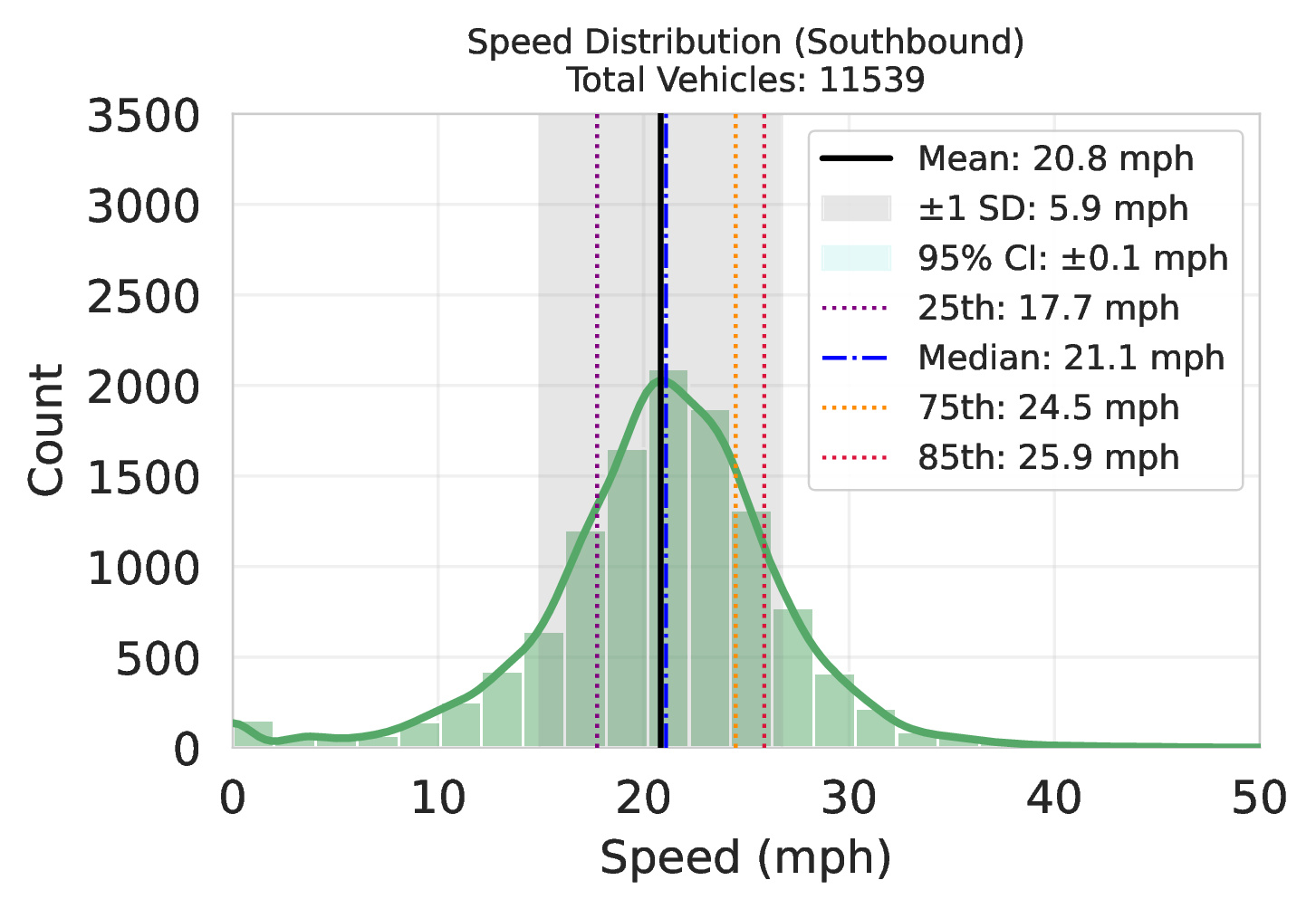} \\
        \multicolumn{3}{c}{Location 1} \\
        \includegraphics[width=0.32\textwidth]{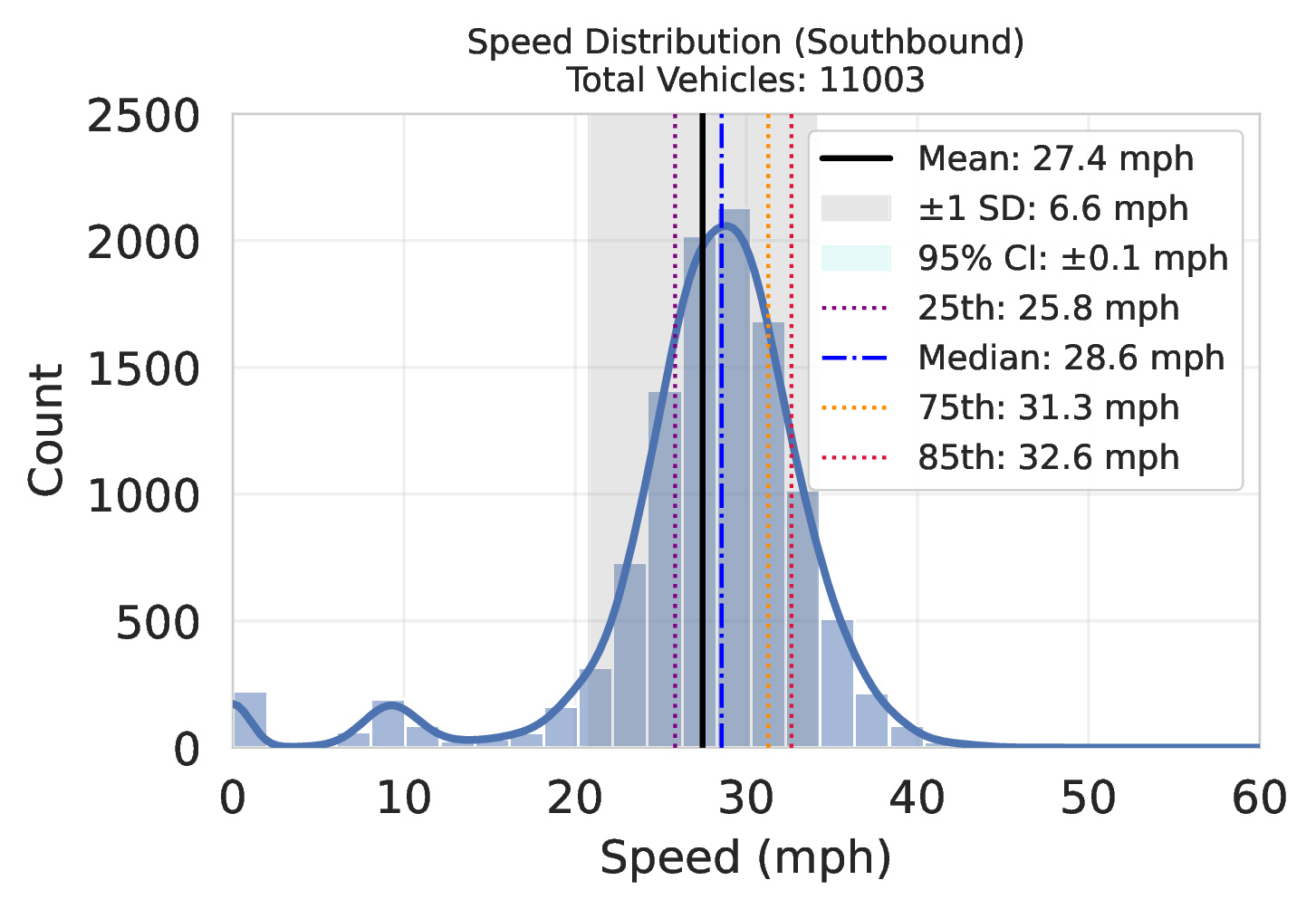} &
        \includegraphics[width=0.32\textwidth]{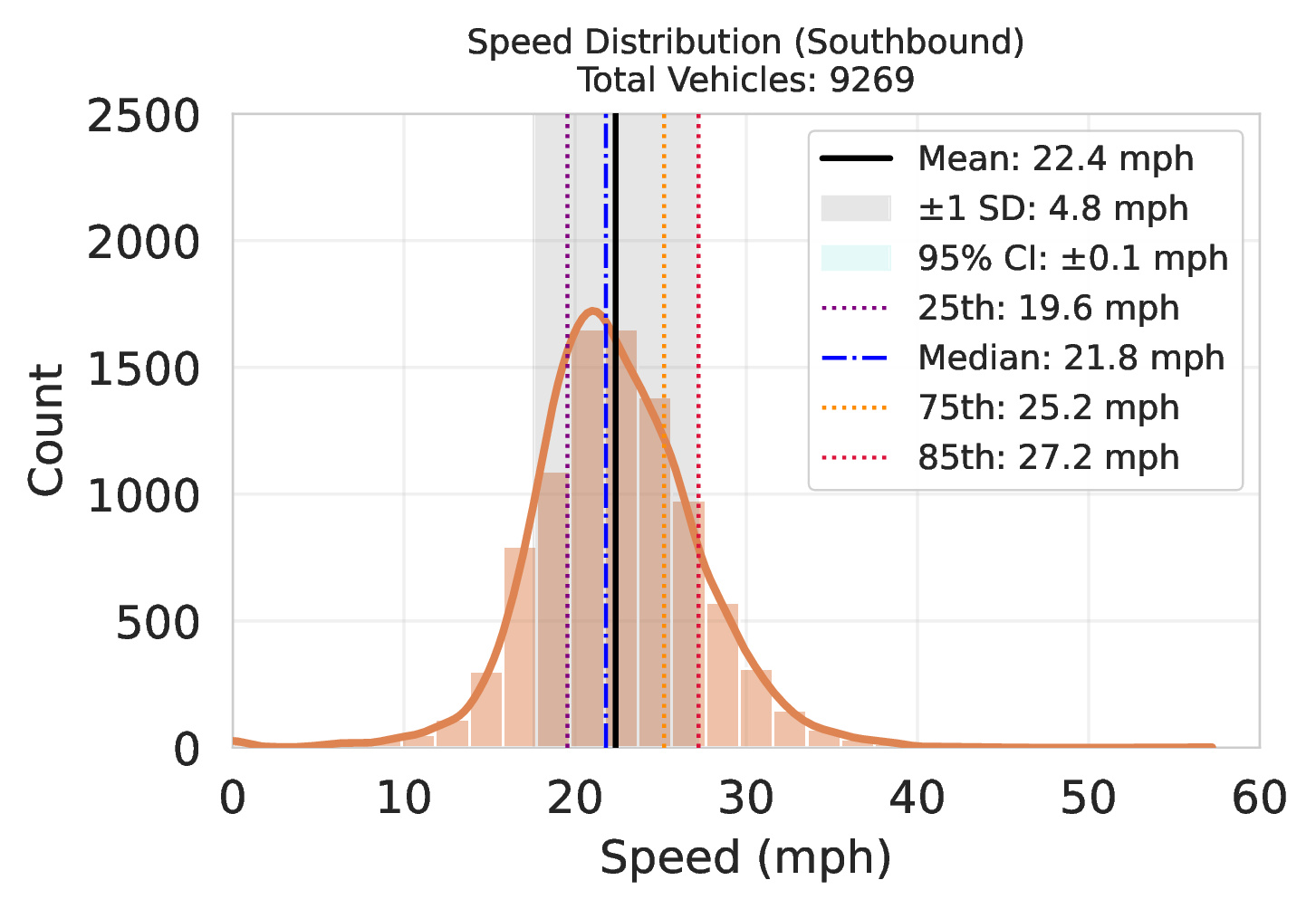} &
        \includegraphics[width=0.32\textwidth]{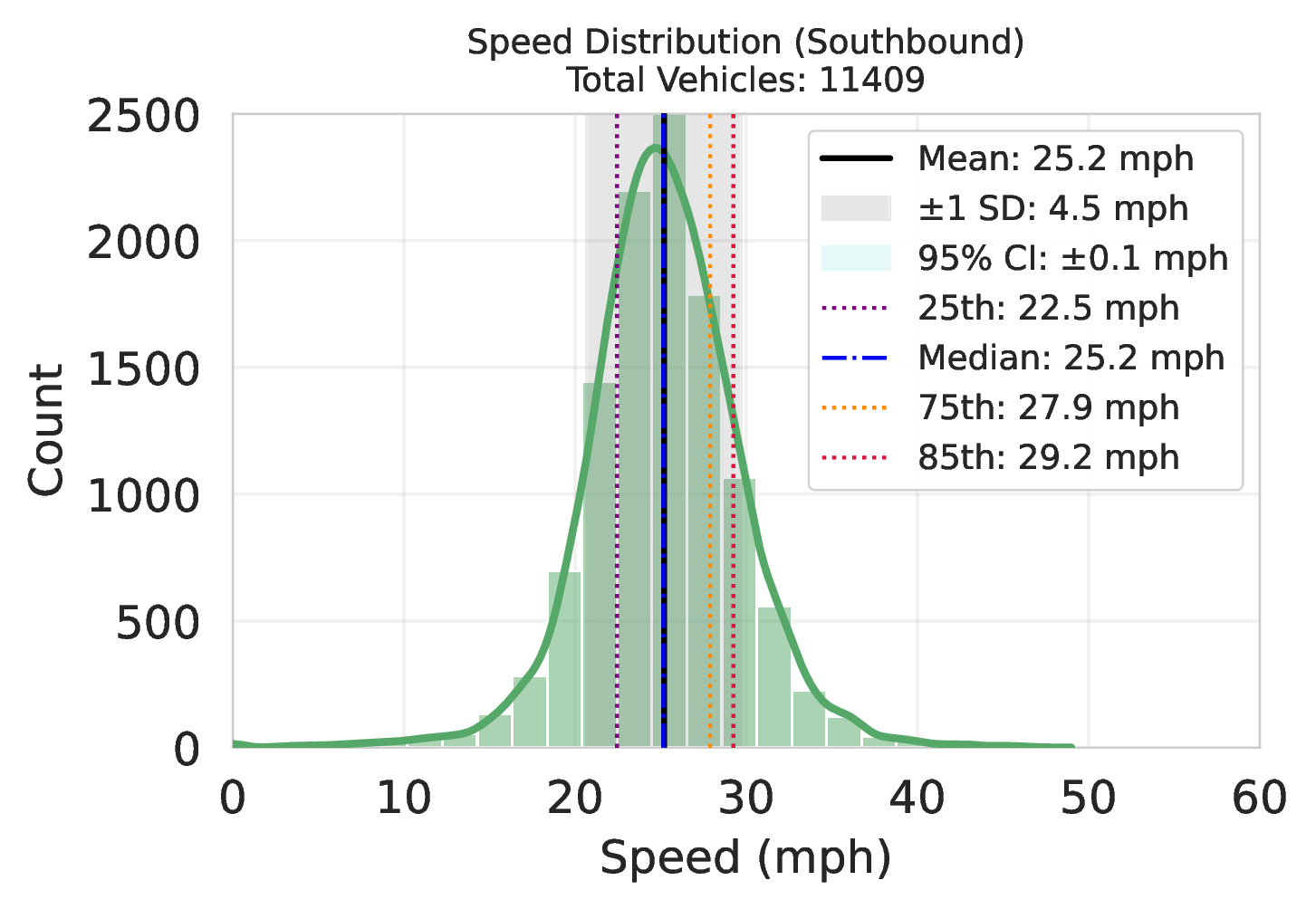} \\
        \multicolumn{3}{c}{Location 2} \\
        \includegraphics[width=0.32\textwidth]{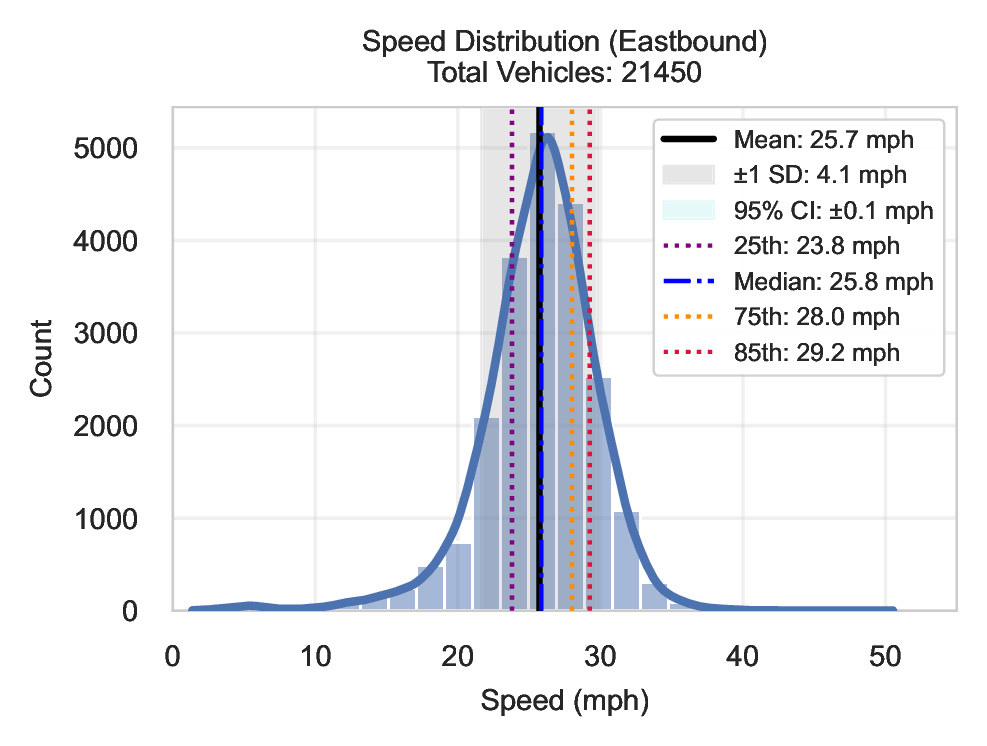} &
        \includegraphics[width=0.32\textwidth]{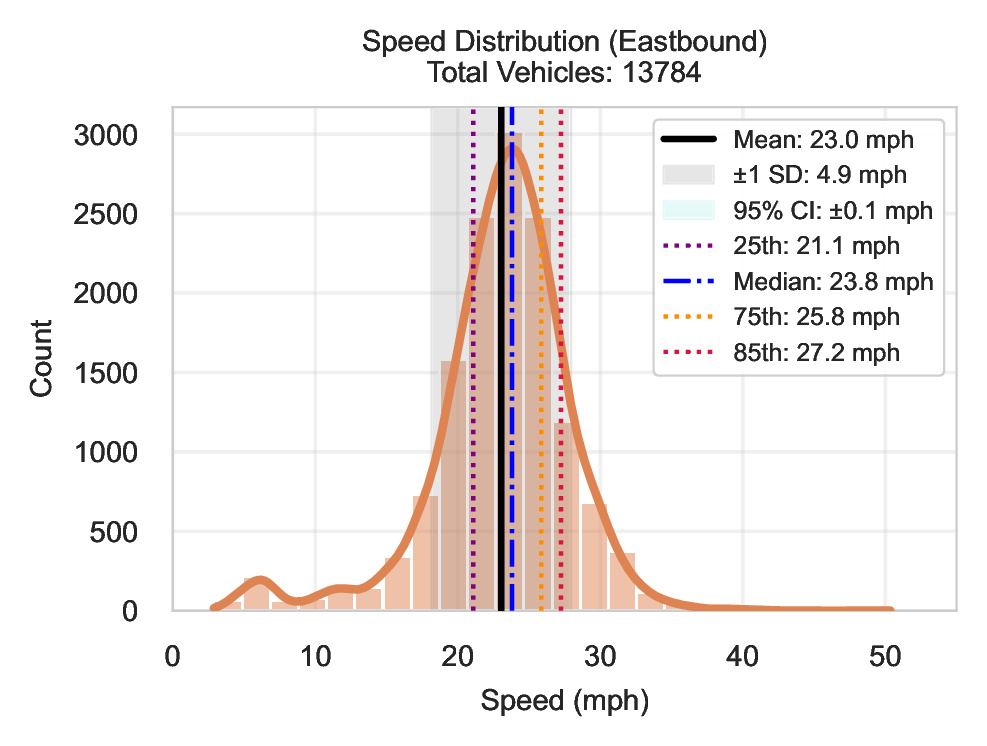} &
        \includegraphics[width=0.32\textwidth]{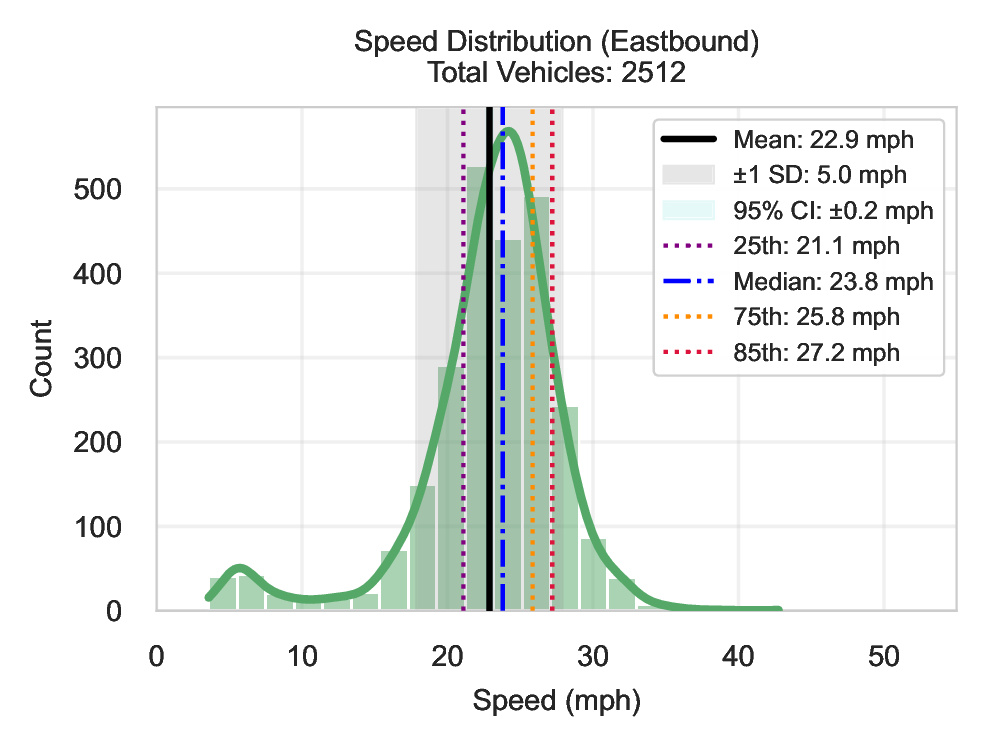} \\
        \multicolumn{3}{c}{Location 3} \\
    \end{tabular}
    \caption{Comparison of speed distributions at unsignalized intersections (Locations 1--3) before and after installation. The graphs illustrate the number of vehicles traveling at each speed, with pre-installation in blue, post-installation Week 1 in orange, and post-installation Week 2 in green, highlighting changes in speed patterns.}
    \label{fig:unsignalized_comparison}
\end{figure*}

\begin{figure*}[t!]
    \centering
    \scriptsize
    \begin{tabular}{>{\centering\arraybackslash}p{0.31\textwidth} >{\centering\arraybackslash}p{0.31\textwidth} >{\centering\arraybackslash}p{0.31\textwidth}}
        \textbf{Pre-installation} & \textbf{Post-installation Week 1} & \textbf{Post-installation Week 2} \\
        \noalign{\vspace{2mm}}
        \multicolumn{3}{c}{\hspace*{10mm}\includegraphics[width=0.95\textwidth]{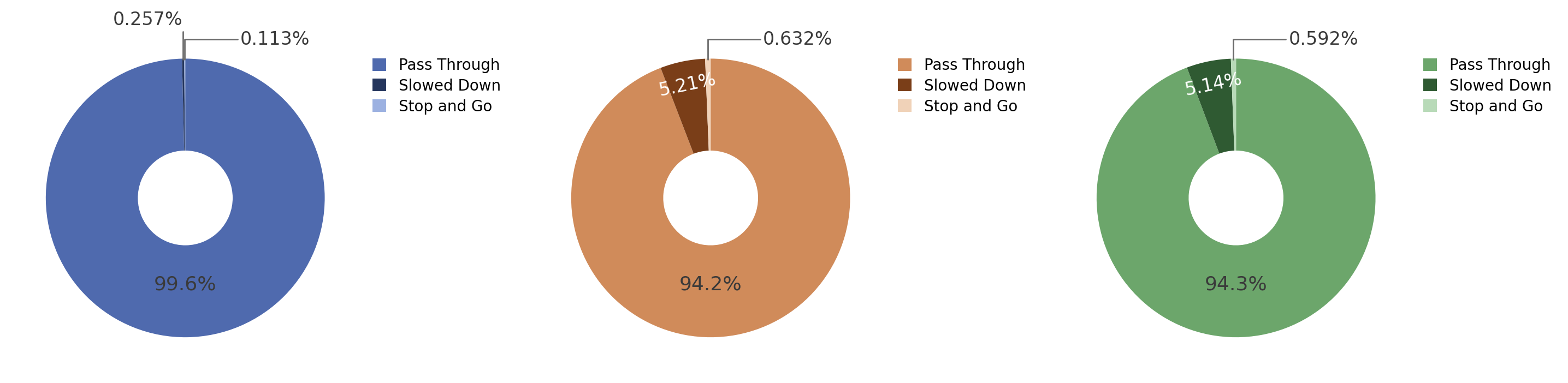}} \\
        \multicolumn{3}{c}{\textbf{Location 1}} \\
        \noalign{\vspace{4mm}}
        \multicolumn{3}{c}{\hspace*{10mm}\includegraphics[width=0.95\textwidth]{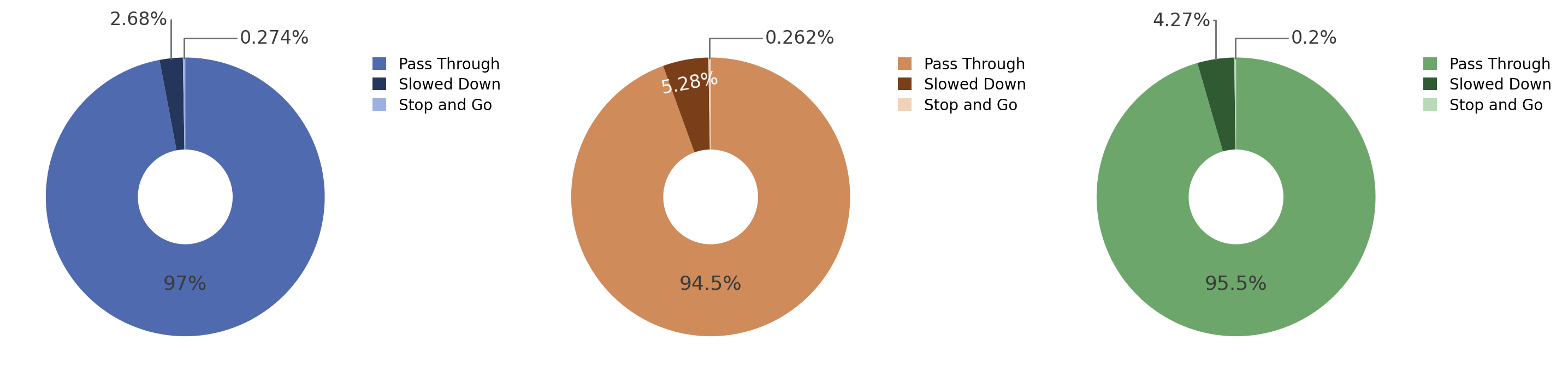}} \\
        \multicolumn{3}{c}{\textbf{Location 2}} \\
        \noalign{\vspace{4mm}}
        \multicolumn{3}{c}{\hspace*{10mm}\includegraphics[width=0.95\textwidth]{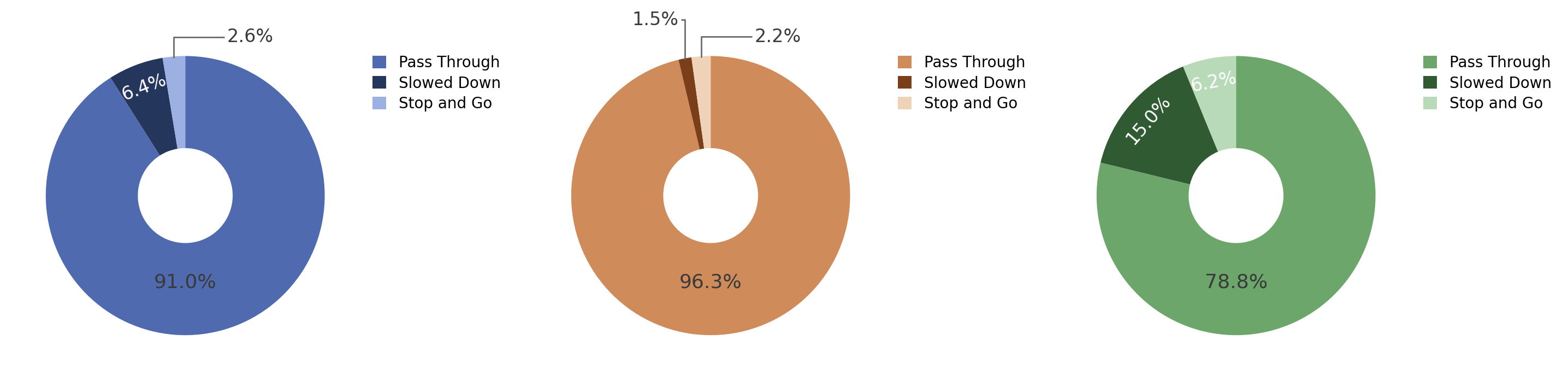}} \\
        \multicolumn{3}{c}{\textbf{Location 3}} \\
    \end{tabular}
    \caption{Comparison of maneuver distributions at unsignalized intersections (Locations 1--3) before and after soft infrastructure changes. Each row presents the behavioral shifts across three phases: Pre-Installation, Post-Installation Week 1, and Post-Installation Week 2. The charts illustrate the percentage of vehicles performing different maneuvers: Pass Through, Slow Down, and Stop and Go.}
    \label{fig:pie_comparison_tabular}
\end{figure*}

The results, presented in Figure ~\ref{fig:unsignalized_comparison} and Table~\ref{tab:mean_speed_unsig}, reveal clear reductions in vehicle speeds following the installation, highlighting the role of hard and soft infrastructure changes in influencing driver behavior. Additionally, vehicle behavior was categorized into three groups: pass-through, slow down, and stop-and-go, as shown in Figure~\ref{fig:pie_comparison_tabular}. These trends not only underscore the effectiveness of the physical changes at the intersections but also highlights the role of AI-driven video analytics in accessing the changes.

\subsubsection{Speed Trends}
Post-installation, Locations 1, 2, and 3 exhibited consistent speed reductions. As shown in Table \ref{tab:mean_speed_unsig}, mean speed decreases ranged from 2.2~mph to 5.0~mph across both observation weeks. While a partial rebound relative to Week 1 was visible at Location 2, mean speeds at all three sites remained below their pre-installation baselines through Week 2.

The 85th-percentile speeds followed a similar downward trend. As detailed in Table \ref{tab:85th_speed_unsig}, these values declined by up to 5.4~mph in Week 1 and remained lower than their baselines in Week 2, reflecting a sustained traffic-calming effect across all unsignalized intersections.

\begin{table}[h]
\centering
\caption{Comparison of Mean Speeds (in mph) Before and After Installation for Unsignalized Intersections}
\label{tab:mean_speed_unsig}
\footnotesize
\begin{tabular}{|c|c|c|c|c|c|}
\hline
\textbf{Loc. ID} & \textbf{Pre} & \textbf{Post W1} & $\Delta_{W1}^{Mean}$ & \textbf{Post W2} & $\Delta_{W2}^{Mean}$ \\ \hline
1 & 25.6 & 20.9 & -4.7 & 20.8 & -4.8 \\ \hline
2 & 27.4 & 22.4 & -5.0 & 25.2 & -2.2 \\ \hline
3 & 25.7 & 23.0 & -2.7 & 22.9 & -2.8 \\ \hline
\end{tabular}
\end{table}

\begin{table}[h]
\centering
\caption{Comparison of 85th Percentile Speeds (in mph) Before and After Installation for Unsignalized Intersections}
\label{tab:85th_speed_unsig}
\footnotesize
\begin{tabular}{|c|c|c|c|c|c|}
\hline
\textbf{Loc. ID} & \textbf{Pre} & \textbf{Post W1} & $\Delta_{W1}^{85^{th}}$ & \textbf{Post W2} & $\Delta_{W2}^{85^{th}}$ \\ \hline
1 & 29.2 & 25.8 & -3.4 & 25.9 & -3.3 \\ \hline
2 & 32.6 & 27.2 & -5.4 & 29.2 & -3.4 \\ \hline
3 & 29.2 & 27.2 & -2.0 & 27.2 & -2.0 \\ \hline
\end{tabular}
\end{table}

Using the results from Figure \ref{fig:unsignalized_comparison}, the comparisons of speed reductions before and after infrastructure modifications reveal a consistent downward trend across all treated sites.

\subsubsection{Vehicle Behavior at intersection}
Beyond speed reduction, the interventions also led to measurable shifts in vehicle behavior, as illustrated in Figure~\ref{fig:pie_comparison_tabular}. 
To quantify these shifts, behaviors were categorized based on the vehicle's moving mean speed ($v_{mean}$) upon approaching the crossing. 
Stop-and-go behavior was identified where $v_{mean} < 5$~mph; this threshold was selected to capture both complete halts and ``rolling stops'' common in yielding scenarios, while accounting for minor sensor variance. 
Slow-down behavior was defined as speeds dropping between 5 and 10~mph, representing a ``yield-ready'' deceleration without a full stop. 
Pass-through behavior characterized vehicles maintaining speeds $\ge 10$~mph, indicating no significant reaction to the crossing context.

Before installation, the vast majority of vehicles (91\%--99.6\%) at these sites exhibited pass-through behavior, reflecting the inherent risk at intersections where few drivers naturally slowed for the crossings.

Post-installation, a behavioral shifts is observed at most sites. In Week 1, the share of pass-through vehicles decreased across Locations 1 and 2, while Location 3 showed a more muted immediate response. Overall, the initial results indicate that the interventions successfully prompted a greater proportion of drivers to adjust their speed when approaching the crosswalks.

By Week 2, behavioral changes had stabilized, with all three sites showing a higher share of drivers slowing or stopping compared to pre-installation levels. Notably, Location 3 displayed a pronounced delayed shift, with a significant increase in yielding and stopping behaviors (exceeding 20\%) as illustrated in the maneuver distributions.

\subsection{Analysis of Signalized Intersections}
Unlike the unsignalized intersections, where vehicle slowing down or stopping could be attributed to driver behavior, signalized intersections inherently regulate vehicle movement through programmed signal timing. Consequently, vehicle behavior metrics such as stop-and-go movements or slowing down are not effective indicators of the intervention's impact and are therefore not analyzed in this section. Instead, this analysis focuses on vehicle speed distributions before and after the intervention, as shown in Table~\ref{tab:mean_speed_sig} and Figure~\ref{fig:signalized_speed_Comparision}. 

\begin{figure*}[t!]
    \centering
    \scriptsize
    \begin{tabular}{c c c}
        \textbf{Pre-installation} & \textbf{Post-installation Week 1} & \textbf{Post-installation Week 2} \\
        \includegraphics[width=0.32\textwidth]{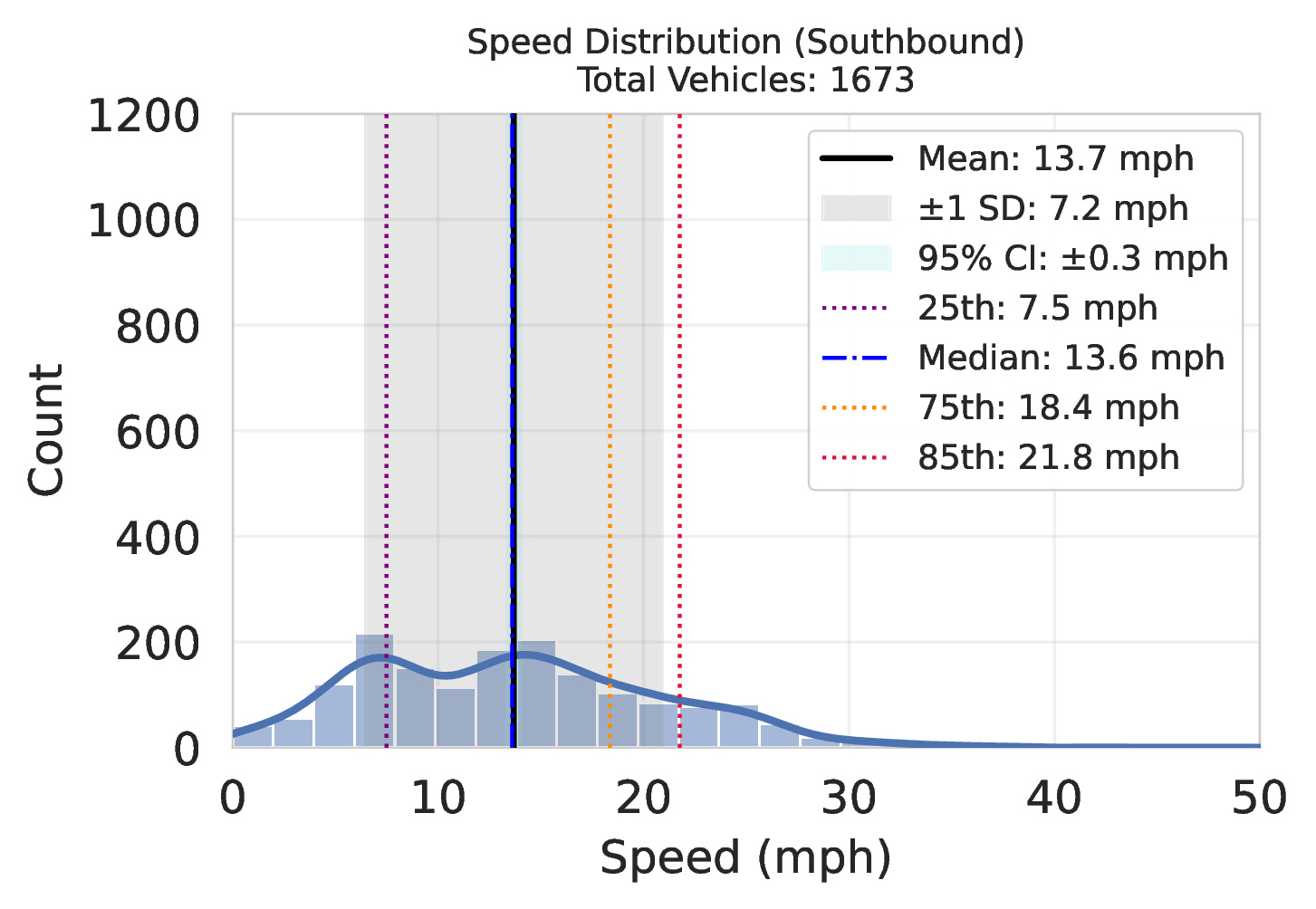} &
        \includegraphics[width=0.32\textwidth]{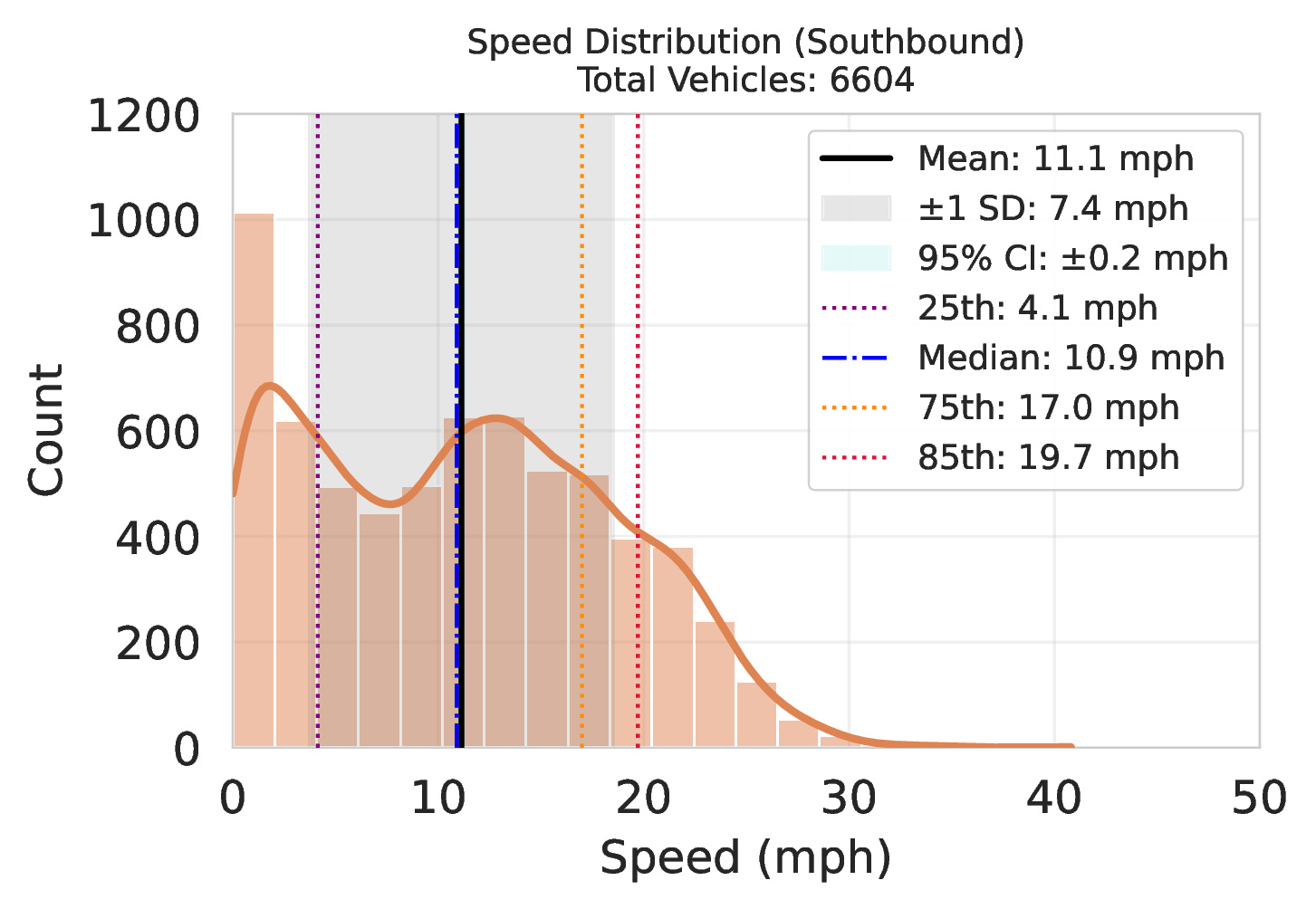} &
        \includegraphics[width=0.32\textwidth]{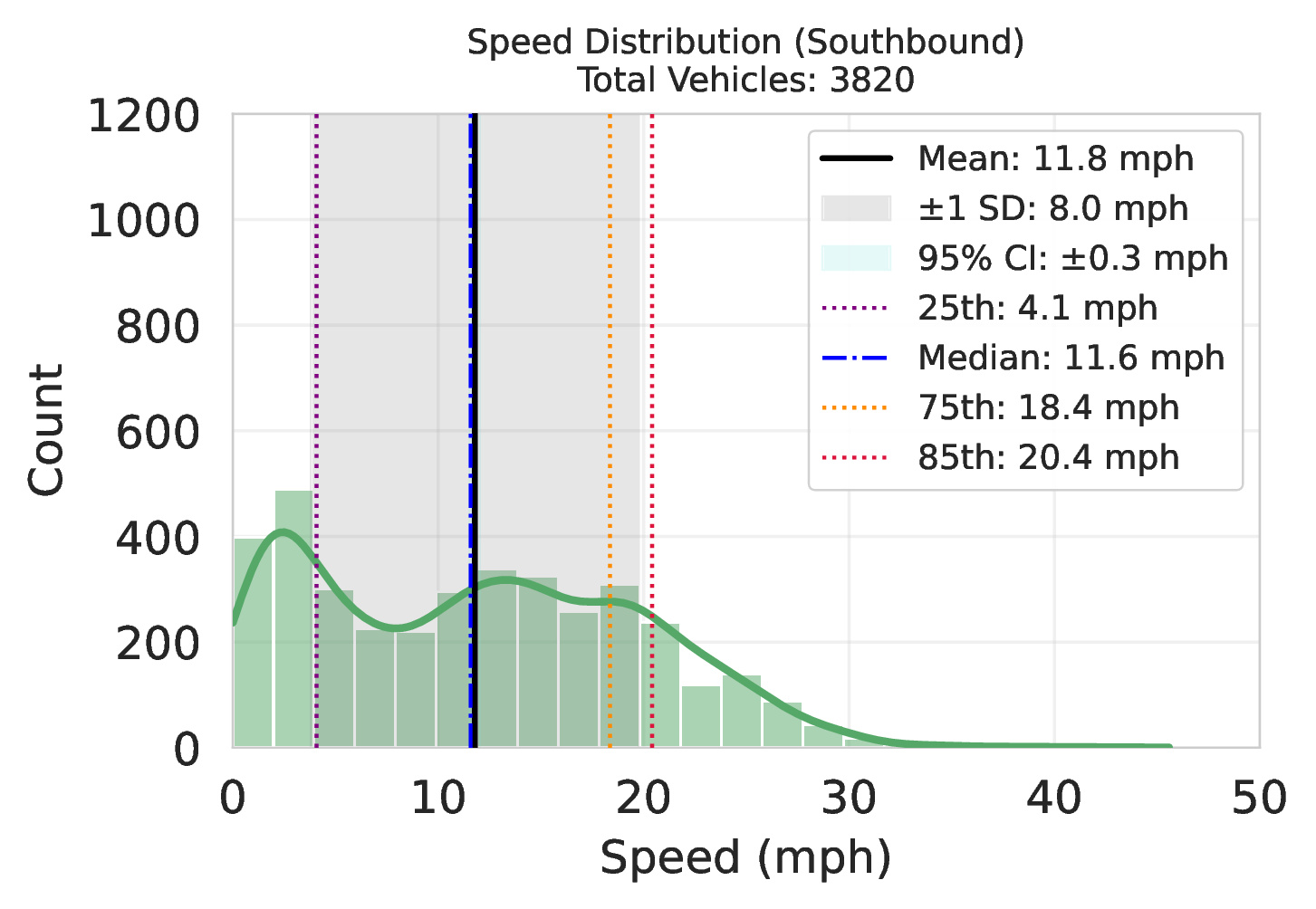} \\
        \multicolumn{3}{c}{\textbf{Location 4}} \\
        \includegraphics[width=0.32\textwidth]{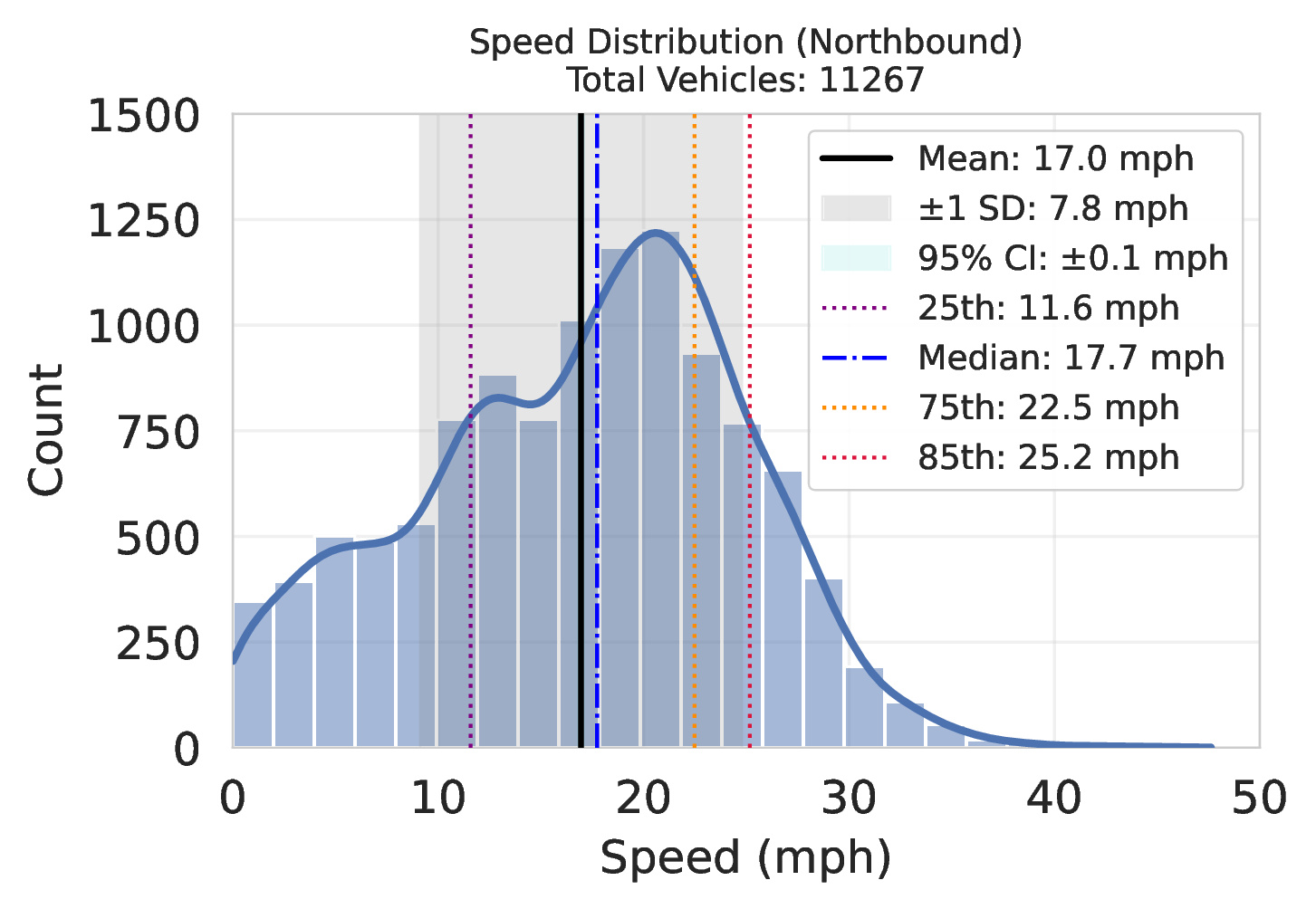} &
        \includegraphics[width=0.32\textwidth]{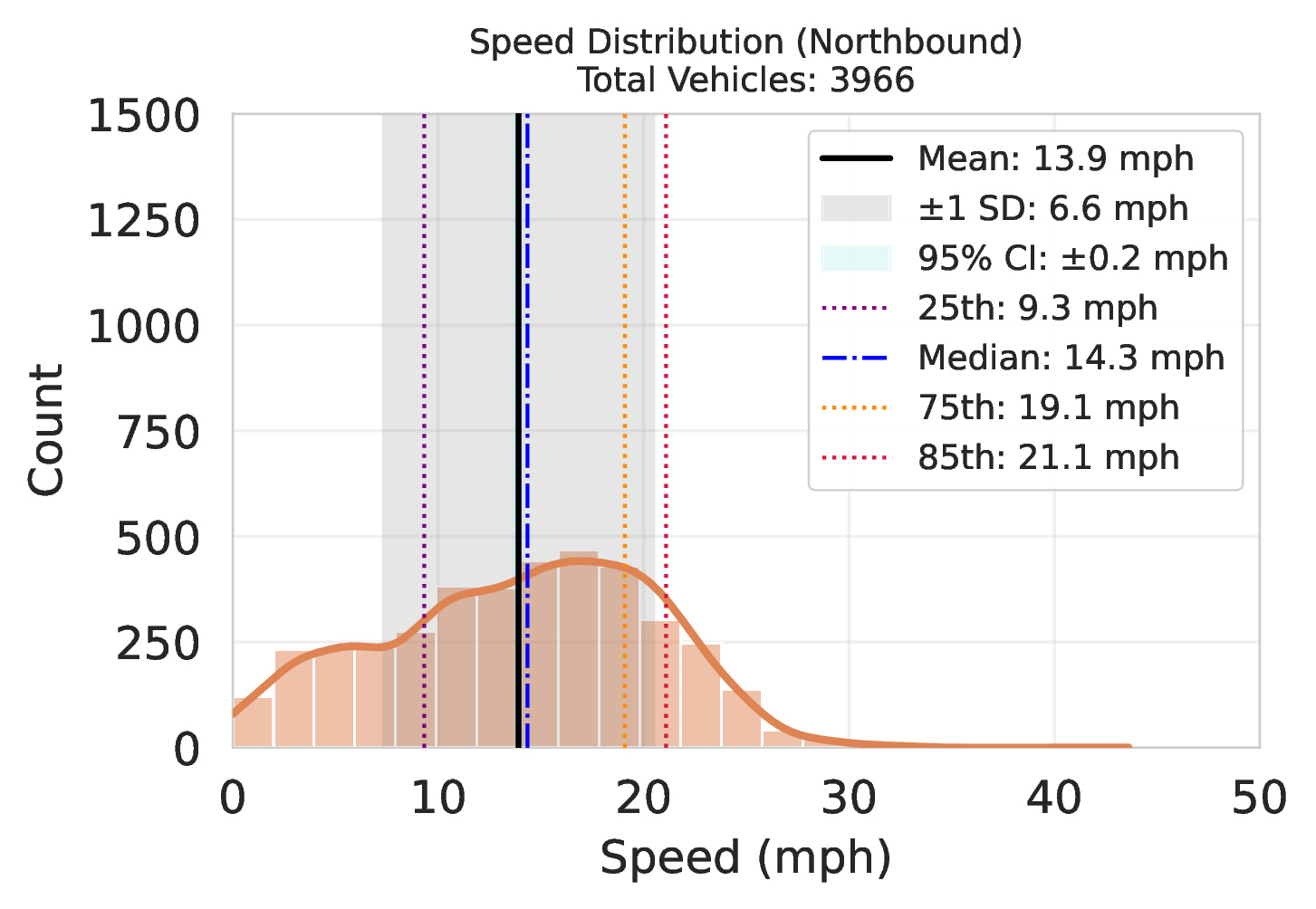} &
        \includegraphics[width=0.32\textwidth]{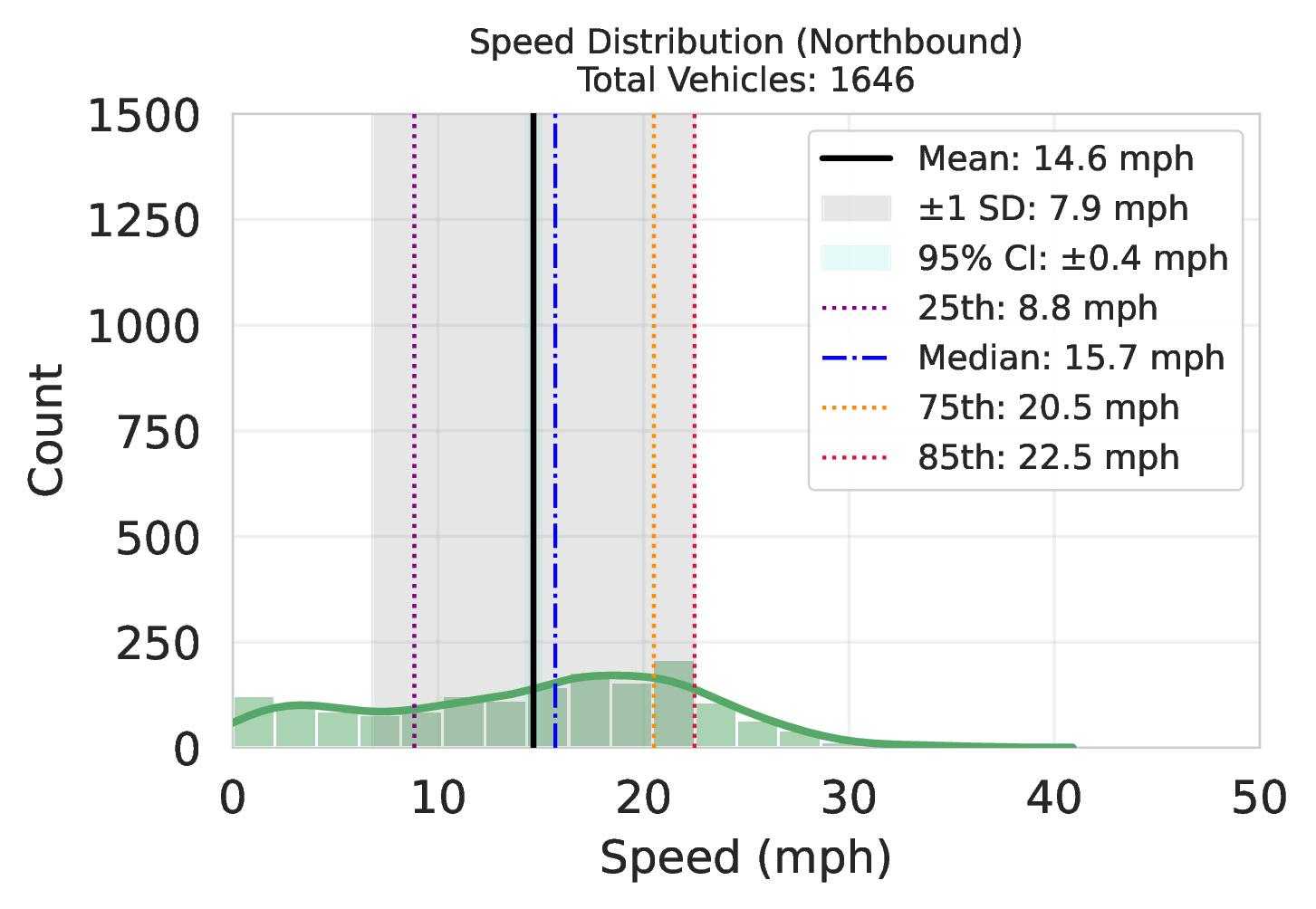} \\
        \multicolumn{3}{c}{\textbf{Location 5}} \\
        \includegraphics[width=0.32\textwidth]{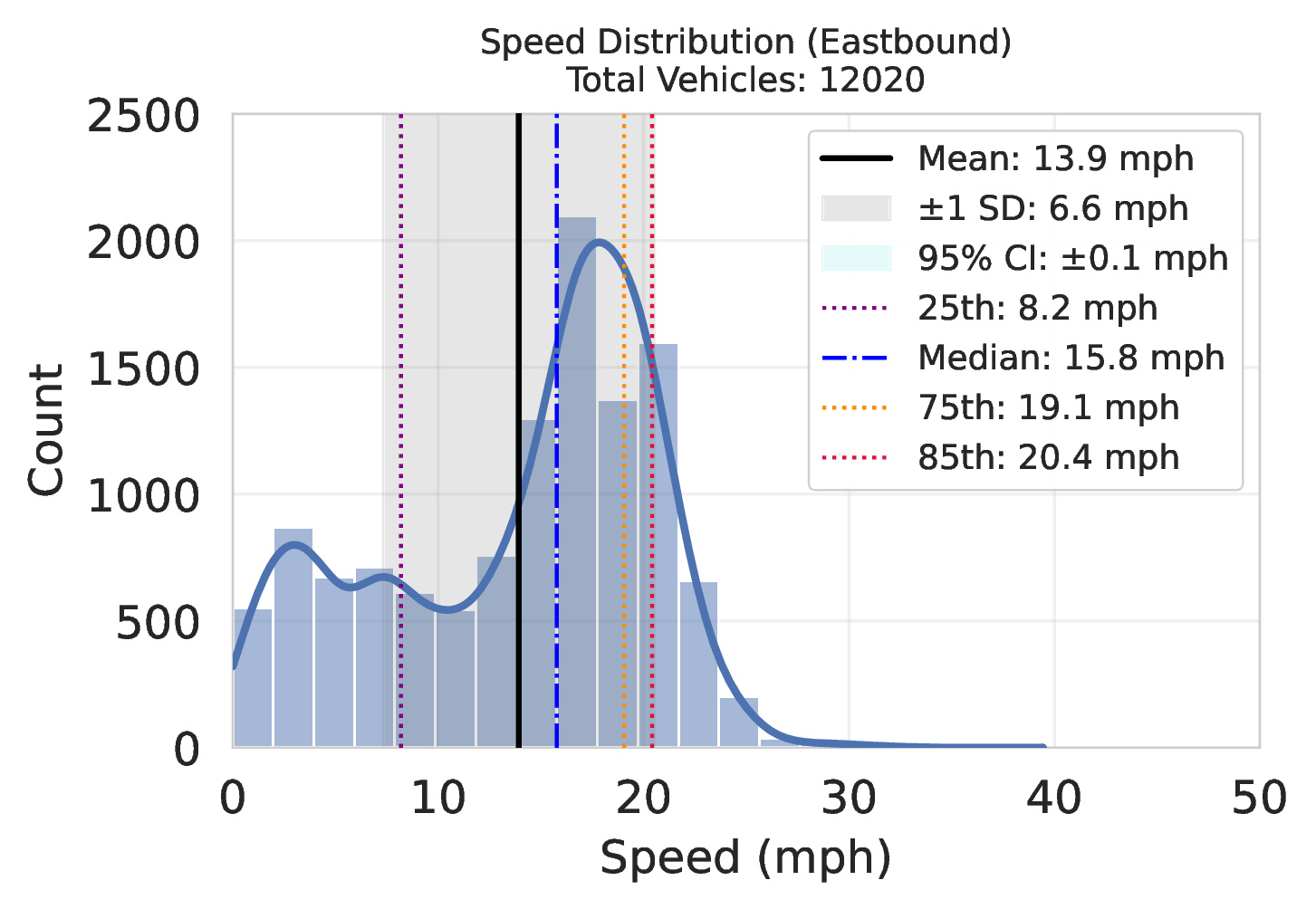} &
        \includegraphics[width=0.32\textwidth]{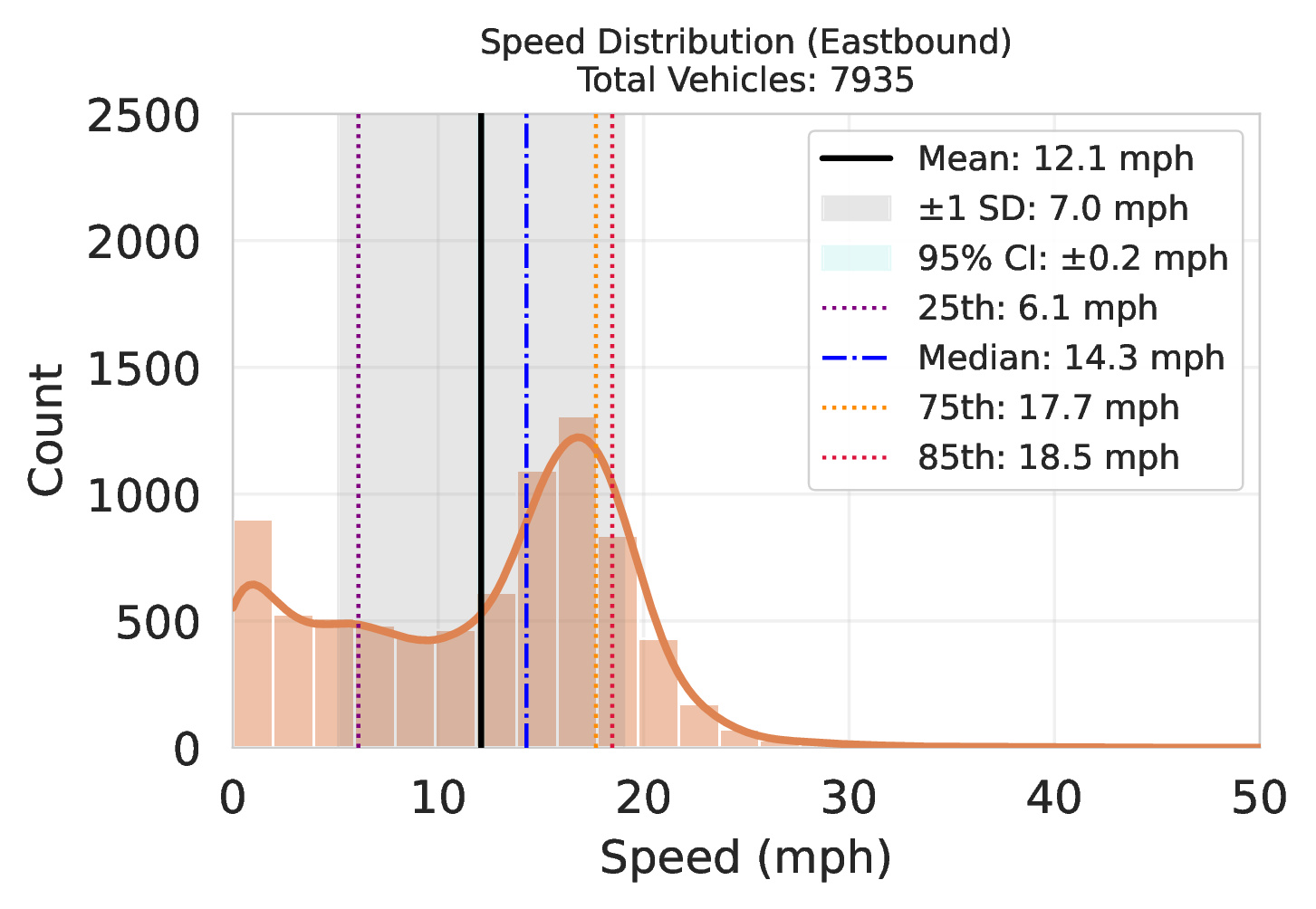} &
        \includegraphics[width=0.32\textwidth]{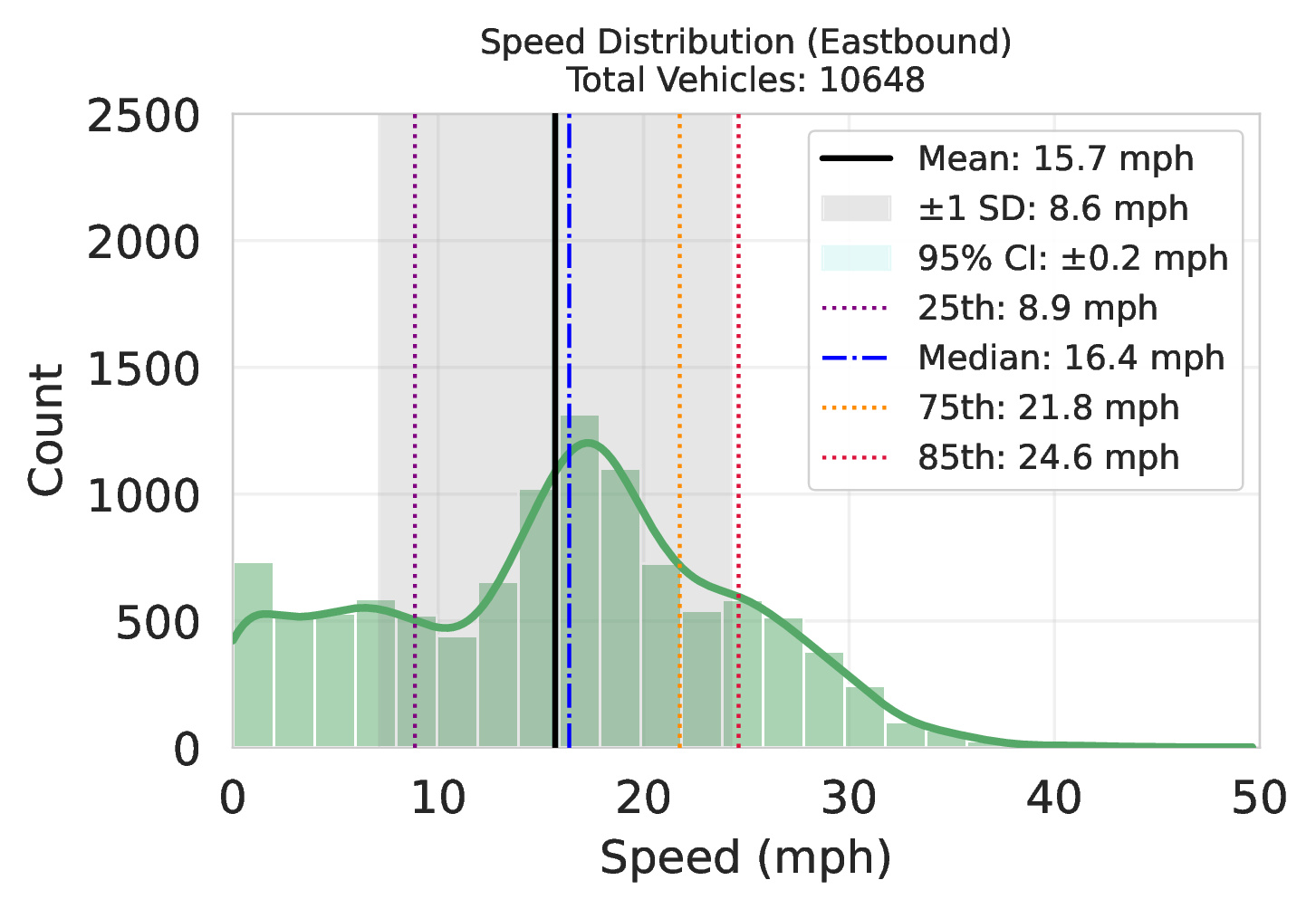} \\
        \multicolumn{3}{c}{\textbf{Location 6}} \\
    \end{tabular}
    \caption{Comparison of speed distributions across six signalized intersections (Locations 4--9) before and after installation. (Part 1 of 2)}
    \label{fig:signalized_speed_Comparision}
\end{figure*}

\begin{figure*}[t!]
    \ContinuedFloat
    \centering
    \scriptsize
    \begin{tabular}{c c c}
        \textbf{Pre-installation} & \textbf{Post-installation Week 1} & \textbf{Post-installation Week 2} \\
        \includegraphics[width=0.32\textwidth]{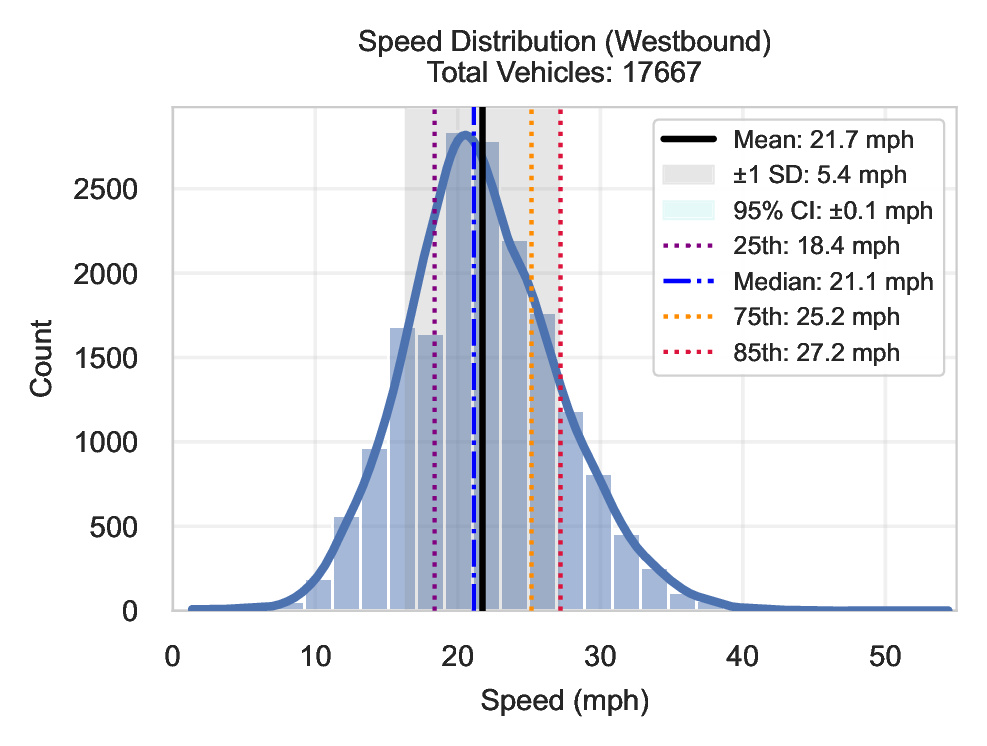} &
        \includegraphics[width=0.32\textwidth]{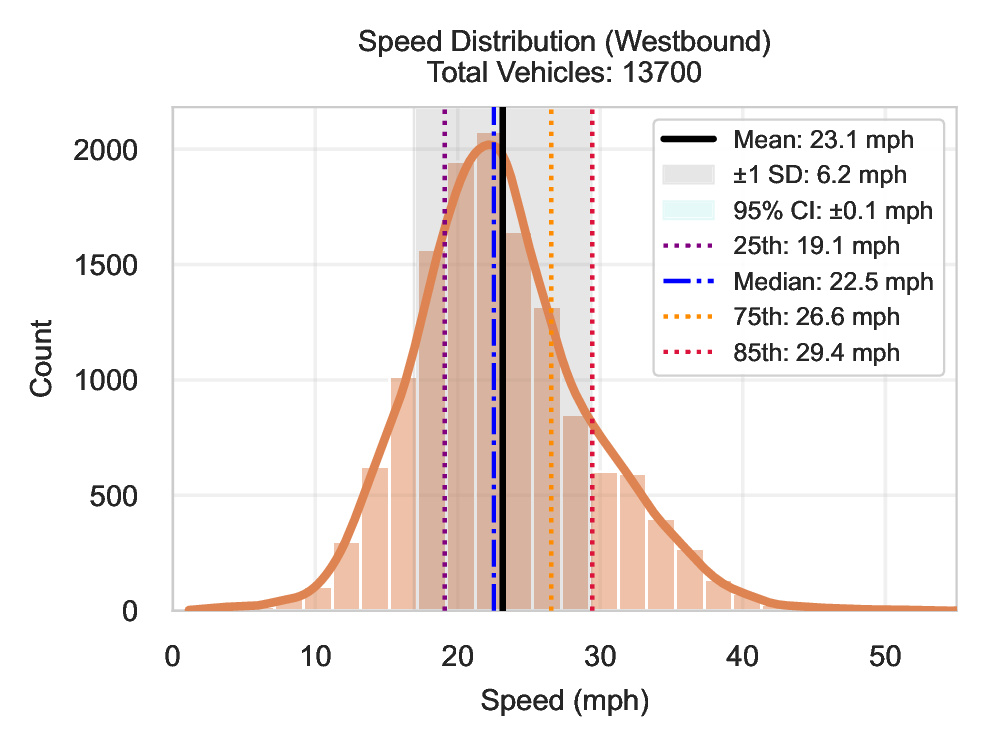} &
        \includegraphics[width=0.32\textwidth]{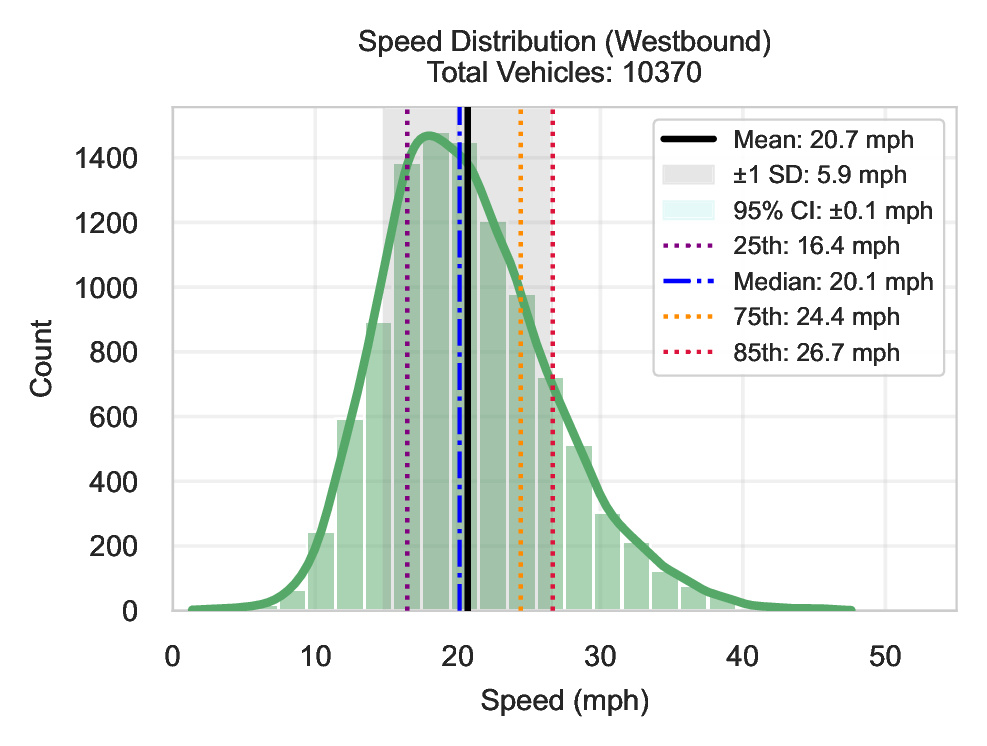} \\
        \multicolumn{3}{c}{\textbf{Location 7}} \\
        \includegraphics[width=0.32\textwidth]{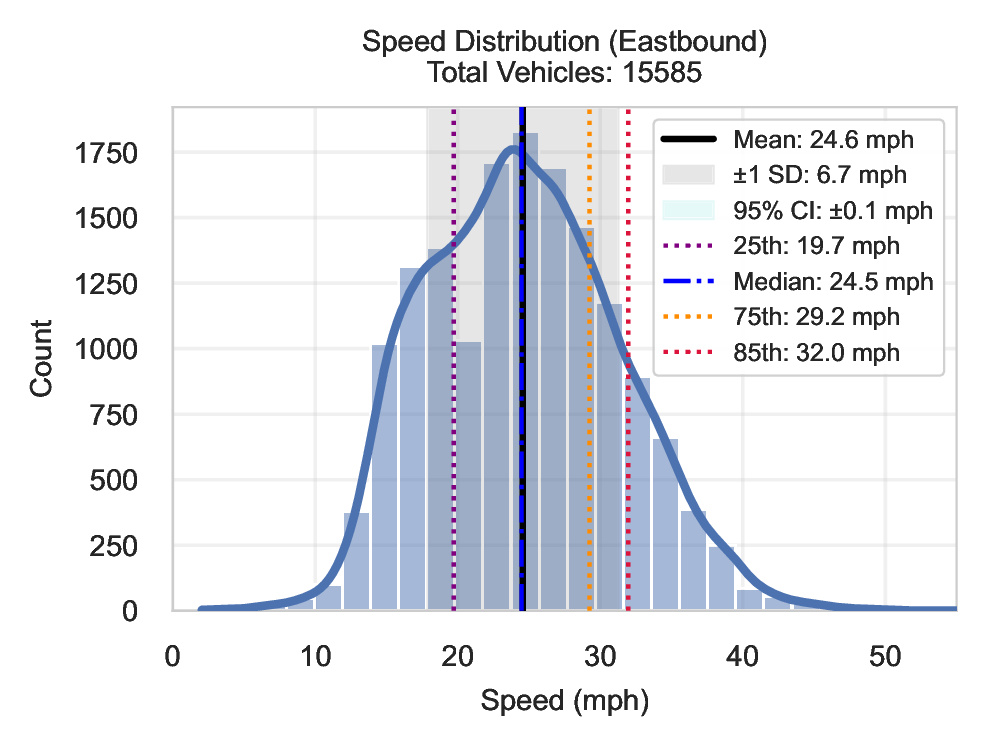} &
        \includegraphics[width=0.32\textwidth]{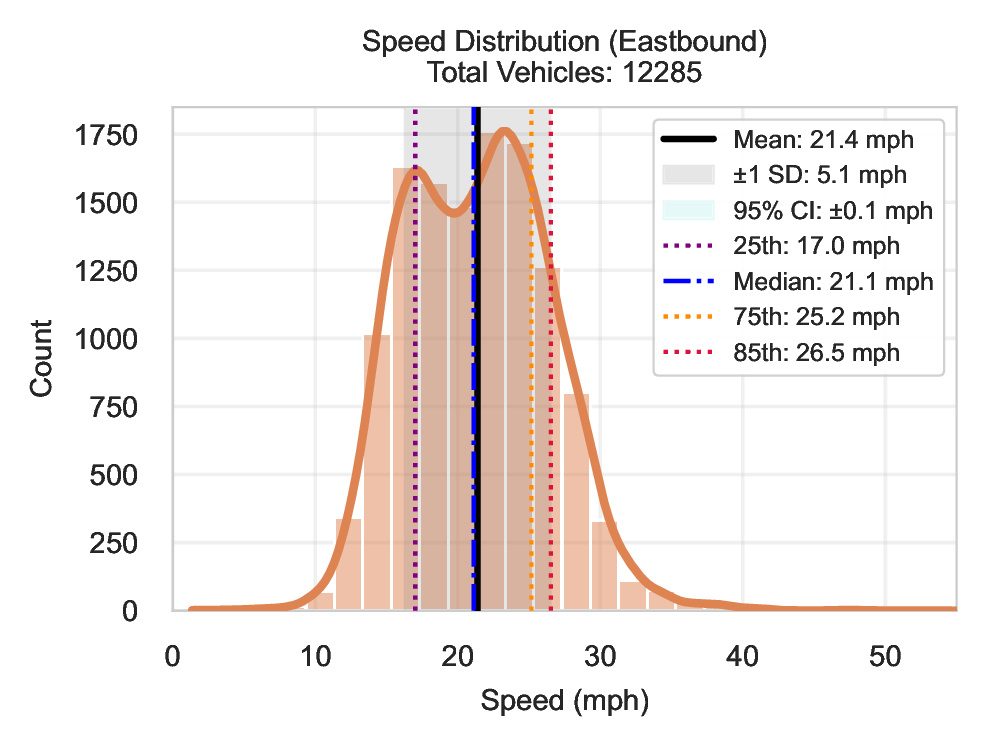} &
        \includegraphics[width=0.32\textwidth]{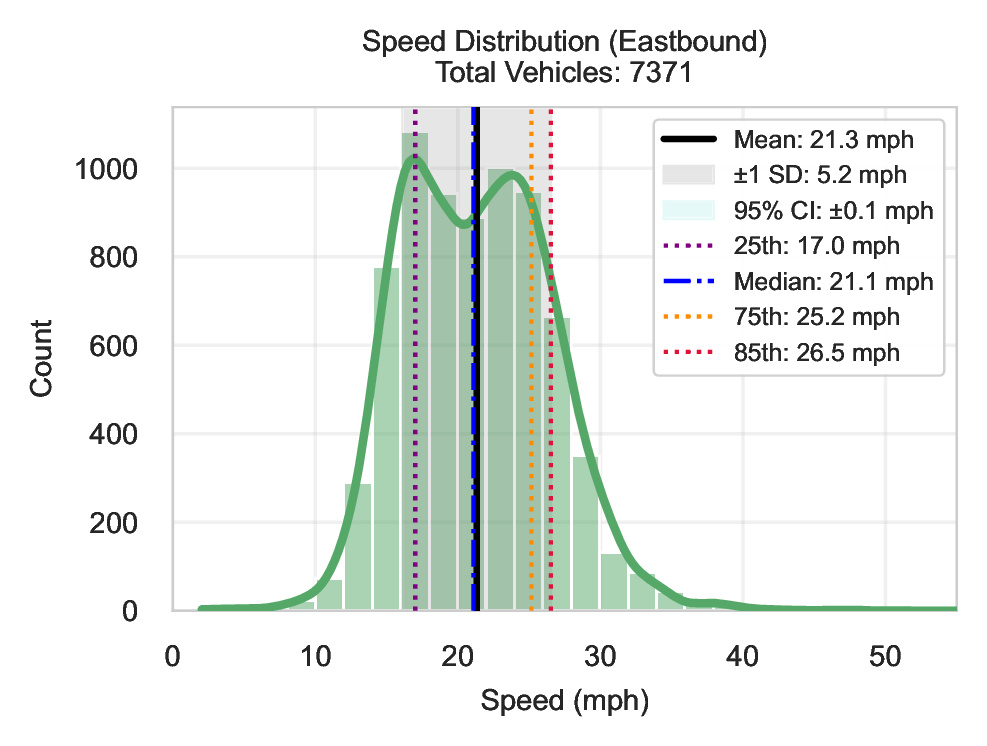} \\
        \multicolumn{3}{c}{\textbf{Location 8}} \\
        \includegraphics[width=0.32\textwidth]{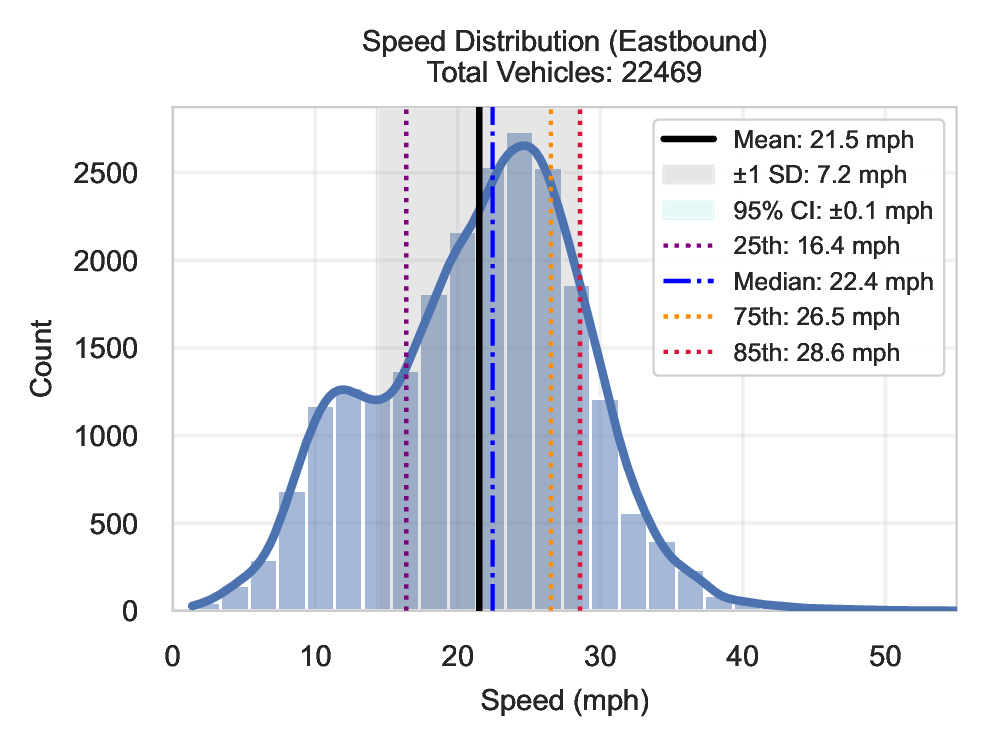} &
        \includegraphics[width=0.32\textwidth]{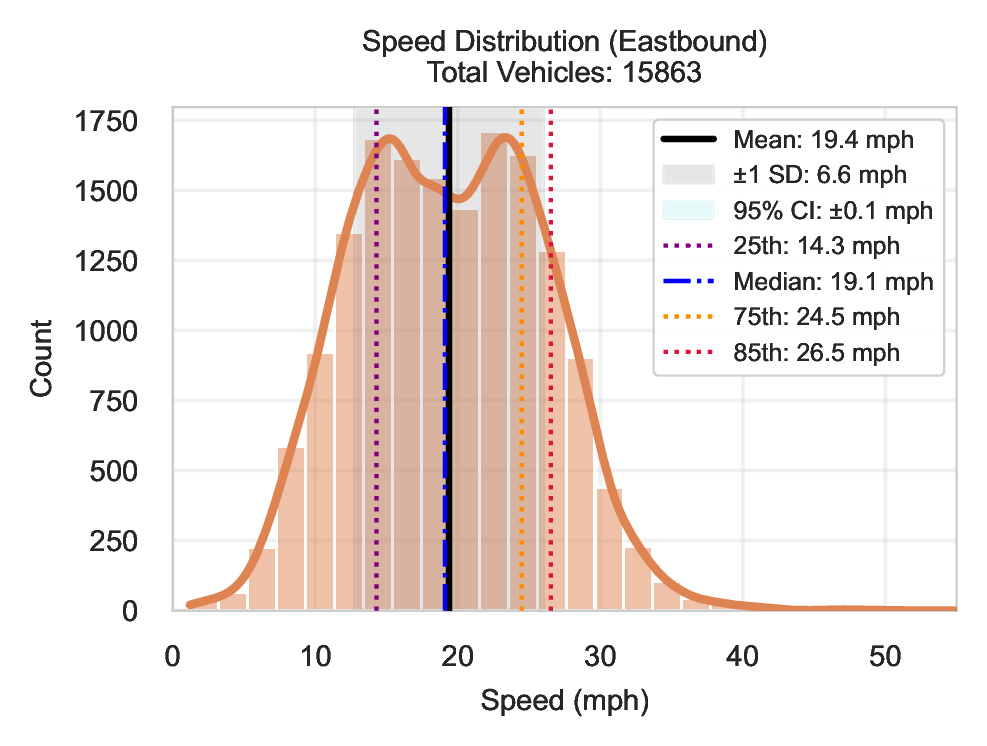} &
        \includegraphics[width=0.32\textwidth]{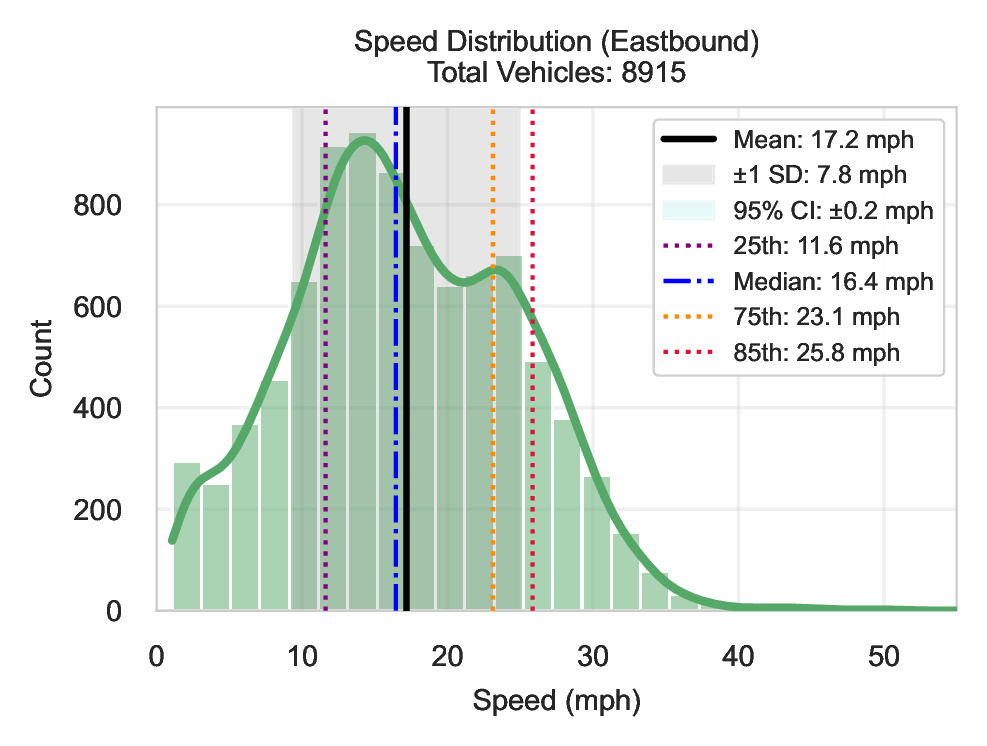} \\
        \multicolumn{3}{c}{\textbf{Location 9}} \\
    \end{tabular}
    \caption{Comparison of speed distributions across six signalized intersections (Locations 4--9) before and after installation. (Part 2 of 2)}
    \label{fig:signalized_speed_Comparision_cont}
\end{figure*}

\subsubsection{Speed Trends}
As illustrated in Figure \ref{fig:signalized_speed_Comparision}, pre-installation mean speeds ranged from 13.7~mph to 24.6~mph, with 85th-percentile speeds reaching up to 32.0~mph across the six locations.

Following the installation of curb extensions and pedestrian refuges, a clear reduction in vehicle speeds was observed. In Week 1, mean speeds dropped at five of the six locations (Locations 4, 5, 6, 8, and 9) as detailed in Table \ref{tab:mean_speed_sig}, while 85th-percentile speeds declined across these same sites (Table \ref{tab:85th_speed_sig}). Location 7 remained the only exception, showing a negligible change in both metrics during this initial phase.

By Week 2, some speed rebound was noted at Locations 4, 5, and 6, though most sites maintained speeds below their pre-installation baselines. Notably, Location 6 exhibited a more significant increase in Week 2, with mean and 85th-percentile speeds rising above pre-installation levels, an exception explored further in the discussion section. Overall, five of the six signalized intersections showed a net traffic-calming effect despite these localized rebounds. 

\begin{table}[ht]
\centering
\caption{Comparison of Mean Speeds (in mph) Before and After Installation for Signalized Intersections}
\label{tab:mean_speed_sig}
\footnotesize
\begin{tabular}{|c|c|c|c|c|c|}
\hline
\textbf{Loc. ID} & \textbf{Pre} & \textbf{Post W1} & $\Delta_{W1}^{Mean}$ & \textbf{Post W2} & $\Delta_{W2}^{Mean}$ \\ \hline
    4 & 13.7 & 11.1 & -2.6 & 11.8 & -1.9 \\ \hline
    5 & 17.0 & 13.9 & -3.1 & 14.6 & -2.4 \\ \hline
    6 & 13.9 & 12.1 & -1.8 & 15.7 & +3.6 \\ \hline
    7 & 14.8 & 15.5 & +0.7 & 13.7 & -1.1 \\ \hline
    8 & 24.6 & 21.4 & -3.2 & 21.3 & -3.3 \\ \hline
    9 & 21.5 & 19.4 & -2.1 & 17.2 & -4.3 \\ \hline
\end{tabular}
\end{table}

\begin{table}[ht]
\centering
\caption{Comparison of 85th Percentile Speeds (in mph) Before and After Installation for Signalized Intersections}
\label{tab:85th_speed_sig}
\footnotesize
\begin{tabular}{|c|c|c|c|c|c|}
\hline
\textbf{Loc. ID} & \textbf{Pre} & \textbf{Post W1} & $\Delta_{W1}^{85^{th}}$ & \textbf{Post W2} & $\Delta_{W2}^{85^{th}}$ \\ \hline
    4 & 21.8 & 19.7 & -2.1 & 20.4 & -1.4 \\ \hline
    5 & 25.2 & 21.1 & -4.1 & 22.5 & -2.7 \\ \hline
    6 & 20.4 & 18.5 & -1.7 & 24.6 & +4.2 \\ \hline
    7 & 19.8 & 19.5 & -0.3 & 18.0 & -1.8 \\ \hline
    8 & 32.0 & 26.5 & -5.5 & 26.5 & -5.5 \\ \hline
    9 & 28.6 & 26.5 & -2.1 & 25.8 & -2.8 \\ \hline
\end{tabular}
\end{table}

\section{Discussion}
Our AI-based video analysis pipeline played a pivotal role in efficiently processing and interpreting the traffic footage. The system achieved a throughput of approximately 23.33frames per second (fps) on the available hardware (AMD Ryzen Threadripper PRO 3975WX CPU with 32-Cores and Nvidia A6000 GPU), enabling us to analyze roughly 10.5 hours of video data (7:30AM to 6:00PM) captured at 10fps in only about 4.5 hours of computation. This level of efficiency underscores how AI can handle large volumes of traffic video data in a time frame far shorter than real-time, which is crucial for scalable and timely traffic monitoring. This efficiency is a stark contrast to manual evaluation, which could take days or even weeks to process the same volume of data, often requiring human observers to review hours of footage. Moreover, the pipeline’s speed estimation proved to be highly accurate. Perspective-transform methods similar to ours have been reported to achieve errors on the order of only $0.5$–$1.5$~mph, reinforcing the reliability of the speed measurements \cite{rodriguez2022analysis}. This combination of efficiency and accuracy provided a robust foundation for evaluating the impact of the intersection interventions and can be scaled for other roadway safety and analysis applications.

Using this AI-based analytics framework, we observed clear evidence
 of vehicle speed reduction following soft Infrastructure changes. In the first week after the installation of the pedestrian refuge and curb extension, average speeds dropped noticeably at most treated locations compared to the pre-intervention baseline, with mean and 85th-percentile speeds typically declining by about 2–5~mph. By the second week, however, the pattern became more mixed. Several intersections maintained or slightly strengthened these initial reductions, whereas others showed a partial rebound in speeds. In a small number of cases, speeds in Week~2 even rose above their pre-installation levels.This indicates that while the infrastructure changes produced an immediate and meaningful calming effect on traffic at most sites, some of that effect diminished over time and, at certain locations, may have reversed.

Several factors may explain this modest rebound in speed observed between the first and second week at some intersections. One likely reason is behavioral adaptation: drivers who were initially surprised or cautious in response to the unfamiliar curb extensions and pedestrian refuges may become more comfortable over time and begin to revert toward their habitual driving speeds. As drivers grow familiar with the presence of these treatments, the initial novelty or uncertainty that prompted extra caution can wear off. In essence, the traffic-calming elements start to blend into the expected road environment, and the perceived risk or emphasis on slowing down diminishes. This increased driver familiarity and shifting perception of the road environment could thus lead to slightly higher speeds after the first few days, even though the physical changes to the roadway remain in place. Such behavioral dynamics suggest that the immediate impact of safety interventions, while strong, can taper as road users acclimate to the new conditions.

The maneuver classifications at the unsignalized intersections generally mirrored the speed reductions, with all three sites showing higher shares of slowing
 and stop-and-go movements by Week~2. At one location, this shift emerged more strongly in Week~2 than in Week~1, highlighting that behavioral responses can unfold on slightly different time scales even under a common treatment.

Interestingly, not all locations exhibited a net decrease in speed. One location stood out as an exception, where vehicle speeds increased rather than decreased post-intervention: Location 6 among the signalized intersections. The unexpected speed increase at Location 6 is attributed to the removal of on-street parking
 required for the intervention. As seen in Figure \ref{fig:before_after_comparison}, removing parked vehicles widened the effective visual corridor, reducing roadway friction and inadvertently encouraging higher speeds, counteracting the calming measures. 

Beyond these factors, it’s worth considering the role of confounding variables that could influence vehicle speeds. Weather conditions, for instance, played a part in this study, with rain reducing the usability of some footage. Wet roads might naturally slow drivers, skewing speed data in unpredictable ways across the pre- and post-installation periods. Traffic volume and time of day also likely varied, as video collection spanned multiple weeks
 and hours (7:00 AM to 6:30 PM). While the study design mitigated some of these effects by focusing on usable footage and applying filters (e.g., excluding stationary vehicles or those not moving in the predefined direction), complete control over external influences was difficult. The video collection spanned different seasons
, but speed reductions were observed consistently across both same-season comparisons (e.g., Location 5) and cross-season comparisons, suggesting that the infrastructure, rather than seasonality, was the primary driver of behavioral change. The AI pipeline’s robustness in handling frame rate variations and filtering irrelevant data helped maintain analytical integrity, but these confounders remind us that real-world traffic dynamics are shaped by more than just infrastructure.

In summary, this study demonstrates how an AI-driven traffic analysis pipeline can serve as an effective tool for rapidly assessing and understanding the impacts of soft infrastructure on driver behavior. By coupling this technology with proactive strategies like, real-time driver feedback and adaptive, and long-duration monitoring, cities and DoTs can not only evaluate interventions more rigorously but also respond and adjust in near-real-time to ensure these interventions achieve their intended long-term safety outcomes. The integration of AI analytics into traffic management thus holds significant promise for enhancing road safety, enabling a shift from retrospective analysis to a more interactive and continuous improvement process for intersection safety interventions.

\subsection{Future Direction}
Looking forward, we see several opportunities to enhance both the AI-based analytics framework and the traffic interventions
 based on these results. One promising improvement is the integration of real-time feedback mechanisms. For instance, the AI system could be connected to dynamic message signs or warning lights at the intersection that activate when vehicles are detected exceeding safe speeds. This immediate feedback loop would leverage the AI’s detection capabilities to actively encourage drivers to slow down in the moment, potentially reinforcing the calming effect of the physical infrastructure. Another avenue is to develop adaptive AI systems that continuously learn from new data. As more footage is analyzed over days and weeks, the AI could refine its algorithms (for example, improving detection/tracking under different lighting or weather, or re-calibrating speed estimates if any drift is observed) to maintain high accuracy and robustness. An adaptive system might also be able to distinguish between normal variation and true changes in behavior more effectively, providing smarter alerts or analyses over time. Furthermore, extending the monitoring period to evaluate long-term behavioral trends would be highly beneficial. While our current analysis covered two weeks post-implementation, monitoring over several months would reveal whether drivers continue to adjust their behavior (positively or negatively) as the changes become a permanent fixture. Long-term data could capture seasonal effects, longer-term adaptation, or the impact of any additional measures (such as enforcement or public awareness campaigns) that often accompany infrastructure changes. Such an extended evaluation would inform whether the initial speed reductions are sustained or if they erode further, thereby guiding whether additional interventions are needed to maintain safety gains.

\section{Conclusion}
This study used AI-driven video analytics to quantify the impact of Pedestrian Refuge, Curb Extension and Centerlines on vehicle speeds and driver behavior at unsignalized and signalized intersections. The results show that at unsignalized intersections, mean speed reductions of up to 18.75\% and 85th-percentile speed reductions of up to 16.56\%. Similarly, at signalized intersections, speed reductions reached up to 20.0\% for mean speeds and 17.19\% for 85th-percentile speeds except for one location. Specifically at unsignalized sites, we also found increases of as much as 12.2\% in the share of vehicles that slow down near the crossings, indicating greater caution of drivers in the vicinity of the interventions. At the same time, not all locations exhibited a net decrease in speed: a small number of sites showed partial rebounds or even higher speeds in the later post-installation period, pointing to the influence of behavioral adaptation and site-specific traffic conditions on long-term outcomes.

The AI-based analytics framework provided a rapid, scalable and cost-effective method for evaluating infrastructure changes, reducing the reliance on manual assessments. While the interventions were effective in lowering speeds, sustained enforcement mechanisms such as AI-driven monitoring, adaptive traffic control, and pedestrian-triggered interventions may be helpful in maintaining long-term compliance. Future work should focus on testing this approach across diverse urban environments to further validate its effectiveness.

\section*{Acknowledgments} This research is funded by the Minnesota Department of Transportation.

% \section*{Author Contributions}
% \noindent
% \textbf{Vinit Katariya}: Conceptualization, Methodology, Software, Formal analysis, Data curation, Writing – original draft, Writing – review \& editing, Visualization.\\
% \textbf{Seungjin Kim}: Formal analysis, Writing – results and video analytics, Visualization.\\
% \textbf{Curtis Craig}: Investigation, Resources, Data curation, Writing – review \& editing.\\
% \textbf{Nichole Morris}: Investigation, Resources, Data curation, Review \& editing.\\
% \textbf{Hamed Tabkhi}: Conceptualization, Supervision, Formal analysis, Writing – review \& editing, Project administration.

\section*{Data Availability Statement}
The data that support the findings of this study are available from the corresponding author, Vinit Katariya, upon reasonable request.

\section*{Conflict of Interest Declaration}
The authors declare no potential conflicts of interest with respect to the research, authorship, and/or publication of this article.

\bibliographystyle{elsarticle-num}
\bibliography{references}
\end{document}